\documentclass[a4paper,twoside]{article}

\usepackage{epsfig}
\usepackage{subcaption}
\usepackage{calc}
\usepackage{amssymb}
\usepackage{amstext}
\usepackage{amsmath}
\usepackage{amsthm}
\usepackage{multicol}
\usepackage{pslatex}
\usepackage{apalike}
\usepackage{algorithm2e}
\usepackage{hyperref}
\usepackage{mathrsfs} 
\usepackage{amsfonts}
\usepackage{mathtools}
\usepackage{amssymb}
\usepackage{dsfont}
\usepackage{caption}
\usepackage{multirow}
\usepackage{tracefnt}
\usepackage{booktabs}
\usepackage{siunitx} 
\usepackage{placeins}
\usepackage{tikz}
\usetikzlibrary{calendar,fpu,matrix, positioning,arrows,arrows.meta,calc,decorations.pathreplacing,decorations.text,spy}
\usepackage{adjustbox}
\usepackage{afterpage}
\usepackage[colorinlistoftodos,prependcaption,textsize=tiny]{todonotes}

\newcommand{\teaser}{%
\begin{figure}
  \includegraphics[width=\columnwidth]{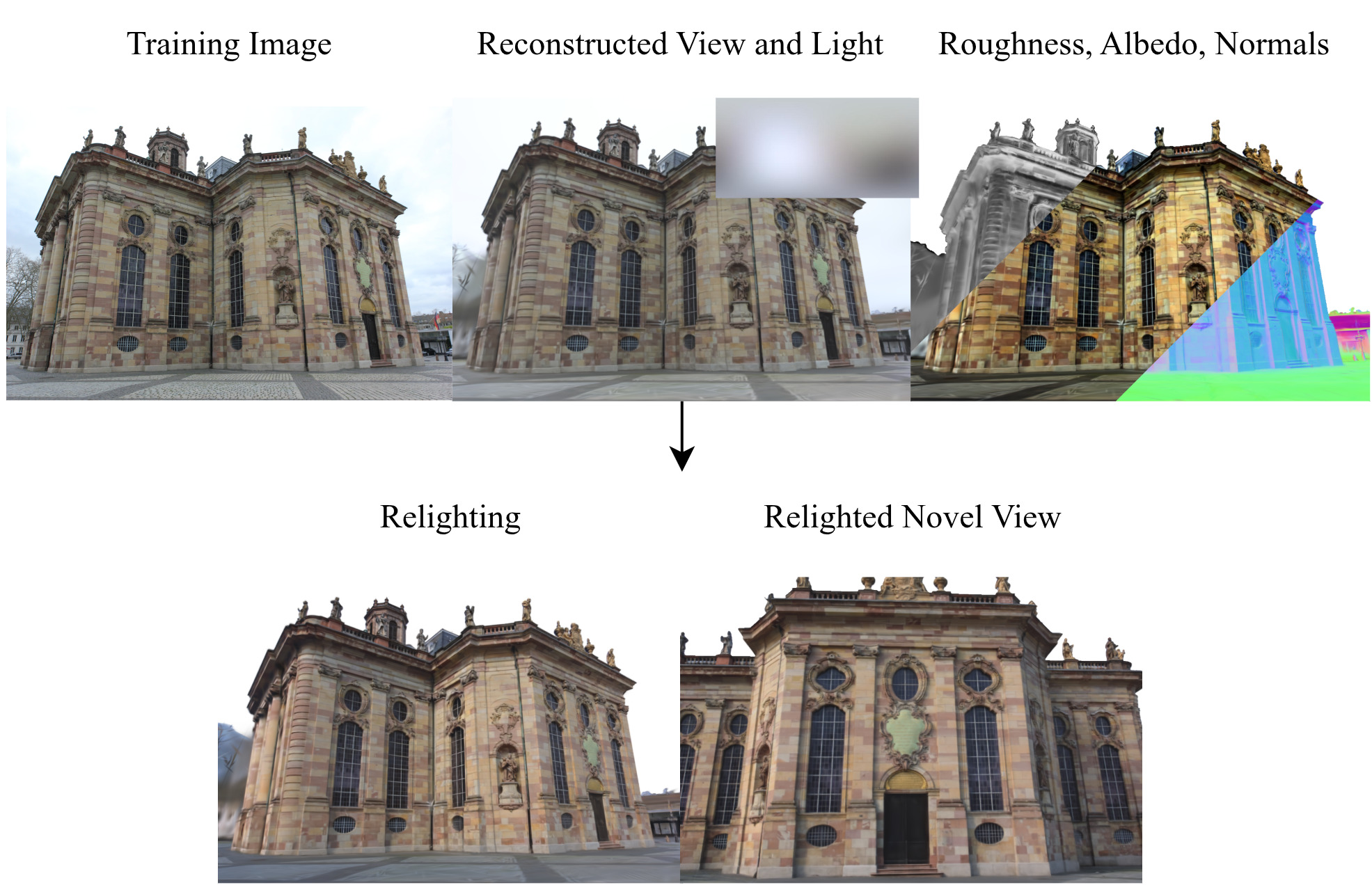}
  \captionof{figure}{R3GW reconstructs a relightable 3D Gaussian Splatting representation of an outdoor scene from a collection of unconstrained multiview images. Our method factorizes the scene’s foreground into geometry, material, and illumination, with the lighting represented as an environment map. This enables illumination editing along with novel view synthesis. The sky, as a non-reflective surface, is modeled with a dedicated set of Gaussians that remain independent of scene lighting and material.}
  \label{fig:teaser}
\end{figure}
}
\usepackage[bottom]{footmisc}
\usepackage{SCITEPRESS}     
\usepackage{orcidlink}

\begin{document}

\title{R3GW: Relightable 3D Gaussians for Outdoor Scenes in the Wild}

\author{\authorname{Margherita Lea Corona\sup{1}\orcidlink{0009-0009-6946-9353}, Wieland 
Morgenstern\sup{1}\orcidlink{0000-0001-5817-7464}, Peter Eisert\sup{1,2}\orcidlink{0000-0001-8378-4805} and Anna Hilsmann\sup{1}\orcidlink{0000-0002-2086-0951}}
\affiliation{\sup{1}Fraunhofer Heinrich Hertz Institute, Berlin, Germany}
\affiliation{\sup{2}Humboldt Universität zu Berlin, Berlin, Germany}
\email{\{margherita.lea.corona, wieland.morgenstern, peter.eisert, anna.hilsmann\}@hhi.fraunhofer.de}
}

\abstract{3D Gaussian Splatting (3DGS) has established itself as a leading technique for 3D reconstruction and novel view synthesis of static scenes, achieving outstanding rendering quality and fast training. However, the method does not explicitly model the scene illumination, making it unsuitable for relighting tasks. Furthermore, 3DGS struggles to reconstruct scenes captured in the wild by unconstrained photo collections featuring changing lighting conditions. In this paper, we present R3GW, a novel method that learns a relightable 3DGS representation of an outdoor scene captured in the wild. Our approach separates the scene into a relightable foreground and a non-reflective background (the sky), using two distinct sets of Gaussians. R3GW models view-dependent lighting effects in the foreground reflections by combining Physically Based Rendering with the 3DGS scene representation in a varying illumination setting. We evaluate our method quantitatively and qualitatively on the NeRF-OSR dataset, offering state-of-the-art performance and enhanced support for physically-based relighting of unconstrained scenes. Our method synthesizes photorealistic novel views under arbitrary illumination conditions. Additionally, our representation of the sky mitigates depth reconstruction artifacts, improving rendering quality at the sky-foreground boundary.}

\onecolumn \maketitle \normalsize \setcounter{footnote}{0}
\vfill

\section{\uppercase{Introduction}}
\label{sec:introduction}
\teaser
In recent years, 3D Gaussian Splatting (3DGS) \cite{kerbl3Dgaussians} has emerged as a revolutionary technique for 3D reconstruction and novel view synthesis of static scenes captured by a collection of multiview images. Unlike prior approaches based on NeRF \cite{mildenhall2021nerf}, which represent the scene volume implicitly with a multi-layer perceptron (MLP), 3DGS uses a discrete set of 3D Gaussian primitives. Each Gaussian is parameterized by learnable attributes that encode the geometry and appearance. To render the scene, 3DGS utilizes an efficient and differentiable rasterization algorithm. In contrast to NeRF, the explicit representation and rendering strategy of 3DGS allow for fast training and real-time rendering while maintaining high visual fidelity. Despite its success, the method presents a major limitation. 3DGS does not disentangle the lighting from the material and geometry of the scene, preventing relighting under novel illumination conditions and thus restricting the approach to the task of novel view synthesis. Given the growing demand for controllable, photorealistic scene representations, e.g.\ for immersive AR/VR experiences, extending 3DGS to enable relighting has recently become a key challenge in the research community. 

Recent works \cite{R3DG2023,liang2023gs,jiang2023gaussianshader} introduced a relightable 3DGS scene representation by integrating Physically Based Rendering (PBR) into the rendering 
pipeline of the method, enabling the decomposition of the scene into lighting, material, and geometry. However, these approaches are restricted to the representation of static environments with fixed illumination across the images. As a result, they do not generalize to scenes captured \textit{in the wild} by unconstrained photo collections, which are characterized by transient objects and, particularly, by changing illumination.

In this paper, we present R3GW, a method that extends 3DGS to support relighting of outdoor scenes captured in the wild. An outdoor environment photographed in an unconstrained setting, in addition to its varying illumination, poses a unique challenge. The sky, unlike the foreground objects, is not a reflective surface, and its appearance is not governed by light-material interaction. Our approach addresses this with a decoupled representation, separately modeling the appearance of the foreground and the sky with two distinct sets of Gaussians with different attributes. For the foreground, R3GW learns the varying illumination of the scene per image, represented by an environment map and parameterized by Spherical Harmonics (SH). For reflection, a Cook-Torrance BRDF parameterized following the Disney material model \cite{cook1982reflectance,burley2012physically} is used. The sky appearance, on the other hand, is modeled independently of the environment light by parameterizing the sky color with low-degree SH, which are learned separately per image.

Our contributions are summarized as follows:
\begin{enumerate}
    \item We present a new 3D Gaussian Splatting framework that handles outdoor scenes captured in the wild and supports relighting under arbitrary environment maps.
    \item We integrate PBR with a Cook-Torrance BRDF into the 3DGS scene representation in a varying illumination setting. To achieve this, we leverage the split-sum approximation technique \cite{SplitSumApprox} and the  SH representation of the environmental light.
    \item We propose a decoupled sky-foreground representation for the 3DGS framework. The sky is represented as a set of Gaussians constrained to the background region of the upper hemisphere enclosing the scene. This separation yields a physically plausible scene representation, where the sky appearance remains independent of surface material. Moreover, the design of the sky region enhances the scene depth estimation and consequently improves the rendering of elements located at the sky-foreground boundary.
\end{enumerate}
\section{\uppercase{Related Work}}
\label{sec:related_work}

\subsection{Relightable 3D Gaussian Splatting for Static Scenes}

Recently, \cite{R3DG2023,liang2023gs,chen2024gi,wu20253d} have adapted 3DGS to support both novel view synthesis and relighting. These methods enable relighting by explicitly modeling reflections using PBR, either by integrating it directly into the color attribute of the Gaussians \cite{R3DG2023} or by applying deferred shading after rasterization \cite{liang2023gs,chen2024gi,wu20253d}. The incoming illumination of the scene is modeled as a global learnable environment map and the 3DGS attributes are augmented with surface normals and BRDF material features. Gaussian Shader \cite{jiang2023gaussianshader} proposes a similar approach by adopting a simplified shading model, and, unlike \cite{R3DG2023,liang2023gs,chen2024gi,wu20253d}, it does not model indirect lighting to account for soft shadows and interreflection effects. Despite enabling high-quality novel view synthesis under arbitrary lighting conditions, these methods rely on the assumption of static lighting. This makes them unsuitable for editing the illumination of scenes captured in an unconstrained environment. 
Our approach addresses this limitation by repurposing PBR combined with 3DGS in an in-the-wild setting.
\subsection{3D Gaussian Splatting for Outdoor Scenes in the Wild}

A number of methods have tailored 3DGS for unconstrained outdoor environments, addressing the appearance variations using per-image and per-Gaussian embeddings \cite{kulhanek2024wildgaussians,zhang2024gaussian,xu2024splatfacto}. Nevertheless, these approaches focus on removing transient objects and do not enable relighting. Similar to our work, WildGaussians \cite{kulhanek2024wildgaussians} augments the initial set of Gaussians with sky Gaussians sampled on the sphere around the scene. However, unlike our method, these Gaussians are neither tracked nor guided to effectively represent the sky.
\subsection{Relightable 3D Representation of Outdoor Scenes in the Wild}
\label{sec:related_work_rel_3drep_w}

Several approaches have extended the neural scene representation (NeRF) of \cite{mildenhall2021nerf} to enable relighting of outdoor scenes in the wild. These methods typically assume a diffuse reflectance model, where the color of a point of the scene surface is independent of the viewing direction and determined by the interaction of incident light, albedo, surface normal, and shadows.

NeRF-OSR \cite{rudnev2022nerf} learns the incident illumination per image, represented as an environment light parameterized by low-degree SH, and models albedo, surface normals, and shadows with additional MLPs. SOL-NeRF \cite{sun2023sol} utilizes a different lighting representation, modeling the scene illumination as a combination of a skylight, parameterized by SH, and sunlight, parameterized by Spherical Gaussians. The method additionally incorporates a prior on the sunlight color. Similarly, NeuSky \cite{gardner2024sky} regularizes the learned environment light of the scene with a neural illumination prior. Compared to earlier deep-learning approaches for outdoor scene relighting \cite{philip2019multi,yu2020self}, these methods improve rendering quality and enable editing both the camera viewpoint and illumination. However, due to the implicit scene representation of NeRF, they suffer from slow training and rendering. Moreover, these approaches only model diffuse reflections, neglecting view-dependent lighting effects. 

In contrast, FEGR \cite{wang2023fegr} uses a hybrid deferred renderer that combines the volumetric rendering of the neural field with mesh extraction to reproduce view-dependent specular highlights and shadows. Nonetheless, the approach remains costly to train. On the other hand, SR-TensoRF \cite{chang2024fast}, building upon \cite{chen2022tensorf}, significantly reduces the training time, but still relies on the assumption of low-frequency lighting.

LumiGauss \cite{kaleta2025lumigauss} introduces a relightable 2D Gaussian Splatting \cite{huang20242d} representation for outdoor scenes captured in the wild. The method assumes a diffuse reflectance model and enables relighting by learning per-image environment light parameterized by second-order SH. The color of the Gaussians is computed as a function of the environment light, albedo, surface normal, and per-Gaussian shadows modeled with SH. LumiGauss achieves high-quality reconstruction and relighting results while offering faster training and rendering than the related NeRF-based techniques. Nonetheless, the method models diffuse reflections only and uses a single collection of Gaussians to represent both the sky and the foreground of the scene. Our method, while not explicitly modeling shadows, adopts a view-dependent reflectance model for the foreground of the scene, which captures more complex lighting effects in addition to diffuse reflections. Moreover, with our decoupled representation of the sky and foreground, we model their appearance independently, ensuring that the Gaussians representing the sky are not assigned with material properties and surface normals.

Concurrently with our work, GaRe \cite{Haiyang2025GaRe} tailors 3DGS for outdoor scene relighting in the wild by enhancing the Gaussians with shading latent codes. These latent vectors are decoded from per-image global illumination features that capture sunlight, skylight, and indirect lighting effects. Unlike our method, GaRe does not enable relighting under arbitrary environment maps due to its implicit illumination model. Furthermore, the approach still relies on the assumption of Lambertian surfaces.
\section{\uppercase{Preliminaries on 3D Gaussian Splatting}}
\label{sec:preliminaries_3DGS}

Our method builds upon 3D Gaussian Splatting (3DGS) \cite{kerbl3Dgaussians}. 3DGS represents a static scene as a set of 3D Gaussian primitives with optimizable attributes: a position $\mathbf{p}=(x,y,z)^{T}\in\mathbb{R}^{3}$, a covariance matrix $\Sigma\in\mathbb{R}^{3\times 3}$, an opacity value $o\in[0,1]$, and an RGB view-dependent color $\mathbf{c}\in[0,1]^{3}$ parameterized by SH. For efficient optimization, the covariance matrix is factorized as $\Sigma=RSS^{T}R^{T}$, where $R$ is a rotation matrix describing the orientation of a Gaussian in world space, while the scaling matrix $S=\text{diag}(s_x,s_y,s_z)$ encodes extensions along each axis. Each Gaussian is then associated with a Gaussian density function, defined for every $\mathbf{u}\in\mathbb{R}^{3}$ as:
\begin{equation}
    g(\mathbf{u}|\mathbf{p}, \Sigma) \propto e^{-\frac{1}{2}(\mathbf{u}-\mathbf{p})^T\Sigma^{-1}(\mathbf{u}- \mathbf{p})}.
\end{equation}

To render the scene, 3DGS uses a fast differentiable tile-based rasterizer implemented in CUDA. The visible Gaussians are first projected (\textit{splatted}) onto the 2D image plane of the given camera view. After projection, the color of each pixel in the output image is computed by sorting the splats by depth and accumulating their colors via $\alpha$-blending:
\begin{equation}\label{eq:3DGS_col}
\mathbf{C}=\sum_{i=1}^{N} \mathbf{c}_i \alpha_i \prod_{j=1}^{i-1} (1- \alpha_j),
\end{equation}
where $N$ denotes the number of visible splats and the blending weights $\alpha_i$ are defined as the product of the Gaussian splat opacity with the 2D Gaussian density function evaluated at the pixel value.

The optimization of a 3D Gaussian Splatting scene begins with a collection of RGB images paired with calibrated cameras obtained from COLMAP SfM \cite{schonberger2016structure}. The positions of the Gaussians are initialized with the sparse point cloud produced by SfM. All the Gaussians attributes are optimized by minimizing the reconstruction loss $\mathcal{L}_{\text{rec}}= \lambda\mathcal{L}_1 + (1-\lambda)\mathcal{L}_{\text{D-SSIM}}, \ \lambda > 0$, between the ground truth images and their renderings. During the first training stage, the number of Gaussians is adaptively controlled. The Gaussians in over and under-reconstructed regions are cloned, while those that do not enhance the rendering quality are pruned.
\begin{figure*}[!t]
  \centering
  \begin{adjustbox}{center}
  \includegraphics[width=0.8\textwidth]{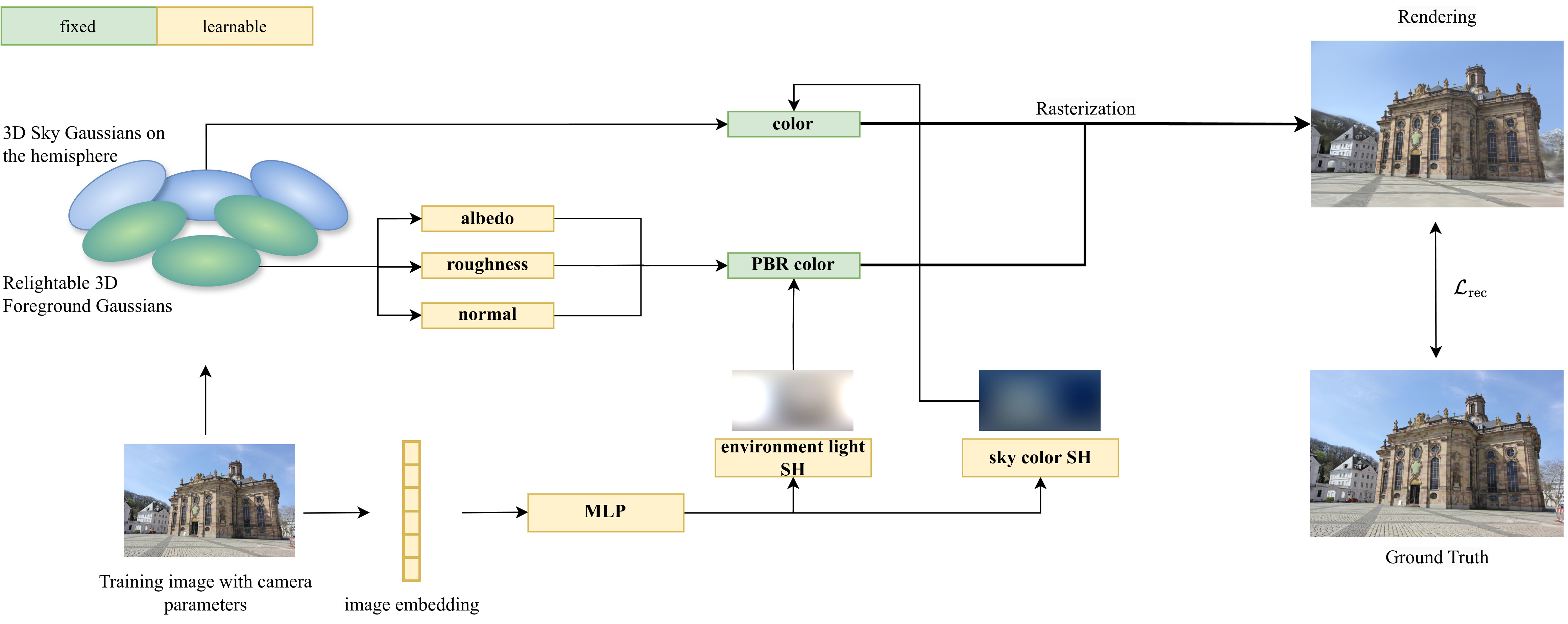}
  \end{adjustbox}
  \caption{\label{fig:pipeline}%
        Training pipeline of R3GW. R3GW learns a relightable 3DGS representation of an outdoor scene captured in the wild. The PBR color of the foreground Gaussians, depending on the surface normals and material properties at their positions, as well as on the environment light, enables relighting of the scene's foreground. In contrast, the color of the sky Gaussians is independent of the scene illumination. The rendered image is formed by rasterizing the sky and the foreground Gaussians in a single pass. The Gaussians are regularized so that the sky Gaussians are responsible for the rendering of the sky pixels, while the foreground Gaussians contribute only to the foreground region. After training, the foreground Gaussians can be rendered under novel lighting conditions by supplying the corresponding environment map as input.}
\end{figure*}
\section{\uppercase{Method}}
\label{sec:method}

R3GW represents the foreground and the sky of an outdoor scene captured in the wild using two separate sets of 3D Gaussians: the foreground and the sky Gaussians. Fig.~\ref{fig:pipeline} overviews our method.

To enable relighting, our approach explicitly models the incoming illumination of the scene as an RGB environment light, represented by an equirectangular environment map and compactly parameterized by SH. Diffuse lighting is typically modeled using second-order SH. Since our goal is to account for additional view-dependent effects, we use  a higher degree of $l_{\text{light}}=4$. We denote the corresponding SH coefficients by $\textbf{l}_{\text{light}}\in\mathbb{R}^{3\times{(l_{\text{light}}+1)^{2}}}$ , and the reconstructed environment light as $L(\cdot)\in[0,\infty)^{3}$. R3GW learns the environment light based on the foreground pixels of the training images. It models the foreground reflections assuming a Cook-Torrance microfacet-based reflectance model \cite{cook1982reflectance} and integrates PBR into the 3DGS representation by computing the color of the foreground Gaussians using the rendering equation \cite{renderingeq}. In contrast, the sky is modeled independently of material properties and surface normals, as its appearance is not influenced by surface interactions. Therefore, we fix the color of the sky relative to the distant lighting and parameterize it with first-order SH, which we denote as $\textbf{l}_{\text{sky}}\in\mathbb{R}^{3\times{(l_{\text{sky}}+1)^{2}}}$, with $l_{\text{sky}}=1$. Sky Gaussians' colors are derived from $\textbf{l}_{\text{sky}}$.
\subsection{Varying Illumination and Sky Appearance}
\label{sec:dynamic_illu_sky_col}

To handle the varying illumination and sky appearance of the scene, following \cite{kaleta2025lumigauss}, each image is associated with a learnable embedding vector $\mathbf{e}$ encoding its appearance information. The environment light and the sky color SH coefficients are predicted per image by an MLP:
\begin{equation}\label{eq:MLP}
    \textbf{l}_{\text{light}}, \textbf{l}_{\text{sky}} = \text{MLP}(\mathbf{e}|\Theta),
\end{equation}
where $\Theta$ is the matrix of the MLP weights, which are optimized alongside the images' latent codes. During training, the image appearance embeddings are stored in an optimizable lookup table.
\subsection{Relightable Foreground Representation}
\label{sec:rel_foreground_rep}

We model the foreground Gaussians as the relightable component of the scene. In addition to the attributes of standard 3DGS, each Gaussian is endowed with the surface normal at its position and with the material properties of the Cook-Torrance BRDF parameterized according to a simplified Disney material model \cite{burley2012physically}: albedo $\mathbf{a}\in[0,1]^{3}$ and roughness $\rho\in[0,1]$. We assume outdoor environments to be dominated by non-metallic surfaces and fix the metallic parameter of the BRDF at zero.

By following \cite{jiang2023gaussianshader}, the normal of a foreground Gaussian is modeled as its shortest axis: $\mathbf{n}=R^{i^*}$, where $R^{i^*}$, $i^*\in\{1,2,3\}$, is the column of the rotation matrix corresponding to the minimum scaling coefficient.

The color of a foreground Gaussian is the light exiting from its position along its viewing direction $\boldsymbol{\omega}_o$, which is given by the rendering equation for non-emitting surfaces \cite{renderingeq}: 
\begin{align}
\mathbf{c}(\boldsymbol{\omega}_o) =\ 
&\underbrace{\frac{\mathbf{a}}{\pi}\int_{\Omega} L(\boldsymbol{\omega}_i)(\mathbf{n}\cdot \boldsymbol{\omega}_i)\,d\boldsymbol{\omega}_i}_{\mathbf{c}_{\text{diffuse}} }\label{eq:fg_gauss_diff_col}\\
&+  \underbrace{\int_{\Omega} L(\boldsymbol{\omega}_i)\frac{DFG}{4(\mathbf{n}\cdot \boldsymbol{\omega}_o)}\,d\boldsymbol{\omega}_i}_{\mathbf{c}_{\text{specular}}}
\label{eq:fg_gauss_spec_col}
\end{align}
where $\Omega$ denotes the upper hemisphere centered around $\mathbf{n}$ and $\boldsymbol{\omega}_i$ is the light incident direction. The functions $D,F,G$, depending on the roughness $\rho$, correspond to the material properties given by normal distribution of the microfacet orientations, the Fresnel reflection, and the geometric attenuation due to the facets shadowing and masking, respectively. Notably, the color of a foreground Gaussian is the sum of a diffuse and a specular color: $\mathbf{c}_{\text{diffuse}},\mathbf{c}_{\text{specular}}$, where only $\mathbf{c}_{\text{specular}}$ depends on the viewing direction $\boldsymbol{\omega_o}$.

The diffuse color is computed using the approach of LumiGauss \cite{kaleta2025lumigauss}, which follows \cite{IrradianceEnvMaps}. By denoting as $\hat{\mathbf{l}}_{\text{light}}\in\mathbb{R}^{3\times9}$ the SH coefficients of the environment light up to the second degree and by $\mathbf{A}\in\mathbb{R}^{9}$ the SH coefficients of the cosine term $(\mathbf{n}\cdot \boldsymbol{\omega_i})$ in Eq.~(\ref{eq:fg_gauss_diff_col}), we compute the diffuse color as: 
\begin{equation}\label{eq:fg_gauss_diff_color_simpl_matrix}
     \mathbf{c}_{\text{diffuse}}=\frac{\mathbf{a}}{\pi}\odot(\mathbf{n}^{T}M(\hat{\mathbf{l}}_{\text{light}},\mathbf{A})\mathbf{n}),
\end{equation}
where $M$ is a symmetric matrix. The definition of $M$, which is derived from $\hat{\mathbf{l}}_{\text{light}}$ and $\mathbf{A}$, can be found in \cite{IrradianceEnvMaps}. For the diffuse color we use only the second-order SH coefficients of the environment light since, as shown in \cite{IrradianceEnvMaps}, the SH coefficients of the cosine term tend to $0$ rapidly after degree $2$.

Following \cite{jiang2023gaussianshader,R3DG2023}, we compute the specular color using the split-sum approximation technique introduced in \cite{SplitSumApprox}, which makes the integral in Eq.~(\ref{eq:fg_gauss_spec_col}) tractable:
\begin{equation}\label{eq:fg_gauss_spec_col_split_sum_approx}
    \mathbf{c}_{\text{specular}}\approx  \underbrace{\int_{\Omega} L(\boldsymbol{\omega}_i)D(\rho, \mathbf{r})\,d\boldsymbol{\omega}_i}_{I_{\text{specular}}}   \underbrace{\int_{\Omega} \frac{DFG}{4(\mathbf{n}\cdot \boldsymbol{\omega}_o)}\,d\boldsymbol{\omega}_i}_{I_{\text{BRDF}}}.
\end{equation}
In the above equation, $I_{\text{specular}}$ is the integral of the environment light $L$ modulated by the normal distribution $D$, which depends on the roughness $\rho$ and the reflection direction $\mathbf{r}=2(\boldsymbol{\omega}_o\cdot\mathbf{n})\mathbf{n}-\boldsymbol{\omega_o}$. The second term, $I_{\text{BRDF}}$, corresponds to the integral of the BRDF specular component, depending on $\rho$ and the cosine of the angle formed by the viewing direction with the surface normal. Given that $I_{\text{BRDF}}$ is independent of the varying environment light, as proposed in \cite{SplitSumApprox}, it is precomputed and stored in a 2D lookup texture. The value of $I_{\text{BRDF}}$ is then obtained as:
\begin{equation}
    I_{\text{BRDF}}=0.04F_1 + F_2,
\end{equation} where $F_1$ and $F_2$ are two scalars derived by interpolating the 2D lookup texture for the Gaussian's roughness $\rho$ and cosine value. The constant 
$0.04$ corresponds to the Fresnel-Schlick reflectivity for non-metallic surfaces. In static lighting scenarios, the integral $I_{\text{specular}}$ can be pre-integrated and stored in the mipmap levels of a cube map \cite{jiang2023gaussianshader,R3DG2023}, following the original split-sum approximation method. However, under changing illumination conditions, representing the environment light for each image as a learnable cube map, would introduce a large number of parameters, making the approach computationally inefficient. Instead, we adopt a simplified approach by applying a smoothing operation to the SH coefficients of the environment light $\mathbf{l}_{\text{light}}$. Specifically, the coefficients are convolved channel-wise with a Gaussian kernel in the SH domain. The blurring strength is set as the square of the roughness. The SH coefficients of the filter are defined as in \cite{chung2005gauss}: 
\begin{equation}
    g^{lm} = g^{l0} = \exp(-l(l+1)\rho^2)
\end{equation}
for every  $l\in\{0,\dots4\}, m\in\{-l,\dots, l\}$. The exponential function in the coefficients of the low-pass filter prevents excessive blurring while effectively smoothing the lighting SH. We approximate the specular light $I_{\text{specular}}$ by evaluating the convolved SH coefficients, $\mathbf{l}_{\text{specular}}$, in the reflection direction $\mathbf{r}$. In summary, our method computes the specular color as:
\begin{equation}\label{eq:spec_color}
    \mathbf{c}_{\text{specular}} = \mathbf{l}_{\text{specular}}^T\mathbf{Y}(\mathbf{r}) \odot (0.04F_1 + F_2),
\end{equation}
where $\mathbf{Y}(\mathbf{r})\in\mathbb{R}^{(l_{\text{light}}+1)^2}$ is the vector of the SH basis functions evaluated in $\mathbf{r}$.

Before rendering, the color of each foreground Gaussian is gamma-corrected with $\gamma=2.2$.
\subsection{Sky Representation}
\label{sec:sky_rep}

Our method represents the sky as the portion of the upper hemisphere surrounding the scene located behind the foreground geometry, effectively treating it as the background of the scene. The sky Gaussians are endowed with the attributes of vanilla 3DGS and, to constrain them to the sky region, their position is parameterized with spherical coordinates. The whole set of sky Gaussians shares a learnable radius attribute $r\in\mathbb{R}^{+}$ and each sky Gaussian has optimizable polar and azimuthal angle attributes, $\theta$, $\phi$, which are clamped during training to the ranges $[0,\frac{\pi}{2}]$ and $[0,\pi]$, respectively. The center of the hemisphere is fixed and estimated as the midpoint of the axis-aligned bounding box of the SfM point cloud. The color of a sky Gaussian in its viewing direction $\boldsymbol{\omega_o}$ is derived by evaluating the SH coefficients:
\begin{equation}\label{eq:sky_gauss_color}
    \mathbf{c}(\boldsymbol{\omega}_o)=\mathbf{l}_{\text{sky}}^T\mathbf{Y}(\boldsymbol{\omega_o}).
\end{equation}

We constrain the sky representation to a subset of the upper hemisphere, as for most of the datasets considering the whole sphere would introduce unnecessary sky Gaussians that do not intersect the view frustum of any training camera. However, the sky model can be easily extended by expanding the ranges of the sky Gaussians' angle attributes.

To initialize the positions of the sky Gaussians, we augment the SfM point cloud, used to initialize the foreground Gaussians' positions, with $N$ additional points sampled uniformly over the designated sky region. Specifically, we first sample the $z$ coordinate of each sky point, compute the corresponding polar angle $\theta$, and then uniformly sample $\phi$. The number of sky Gaussians is chosen proportionally to the estimated sky distance, which is set as the initial value of $r$. Similarly to \cite{kulhanek2024wildgaussians}, we compute it as the $99$th percentile of the distances of the foreground points from their average. See the Appendix for a visualization of the initial set of sky Gaussians.

To track and control the sky Gaussians, each Gaussian is assigned a fixed boolean attribute marking it as sky or foreground.
\def\height{64px}
\def\spysize{42px}
\def\offset{36px}
\def\scale{0.7}
\def\scaleleft{0.7}
\def\distabove{0.3em}
\def\distleft{0.3em}

\newcommand{\qualitativecompdepth}{
\centering
\resizebox{\linewidth}{!}{
    \begin{tikzpicture}[
        >=stealth',
        title/.style={
            anchor=base,
            align=center,
            scale=\scale,
        },
        spy using outlines={rectangle, red, magnification=3, connect spies, size=\spysize}
    ]
    \matrix[matrix of nodes, column sep=0pt, row sep=0pt, ampersand replacement=\&, inner sep=0, outer sep=0, font=\Large] (qualitative) {
        \includegraphics[width=.8\linewidth]{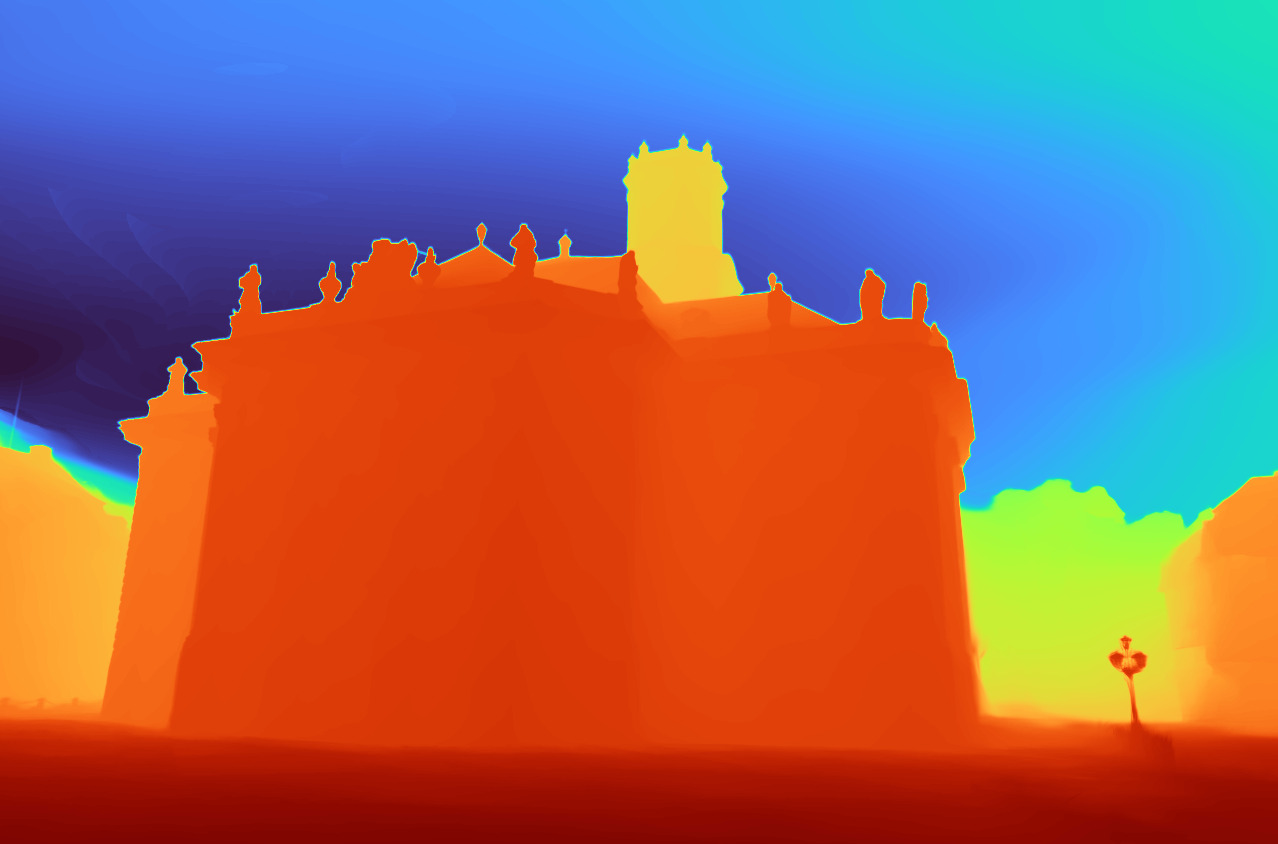} \&
        \includegraphics[width=.8\linewidth]{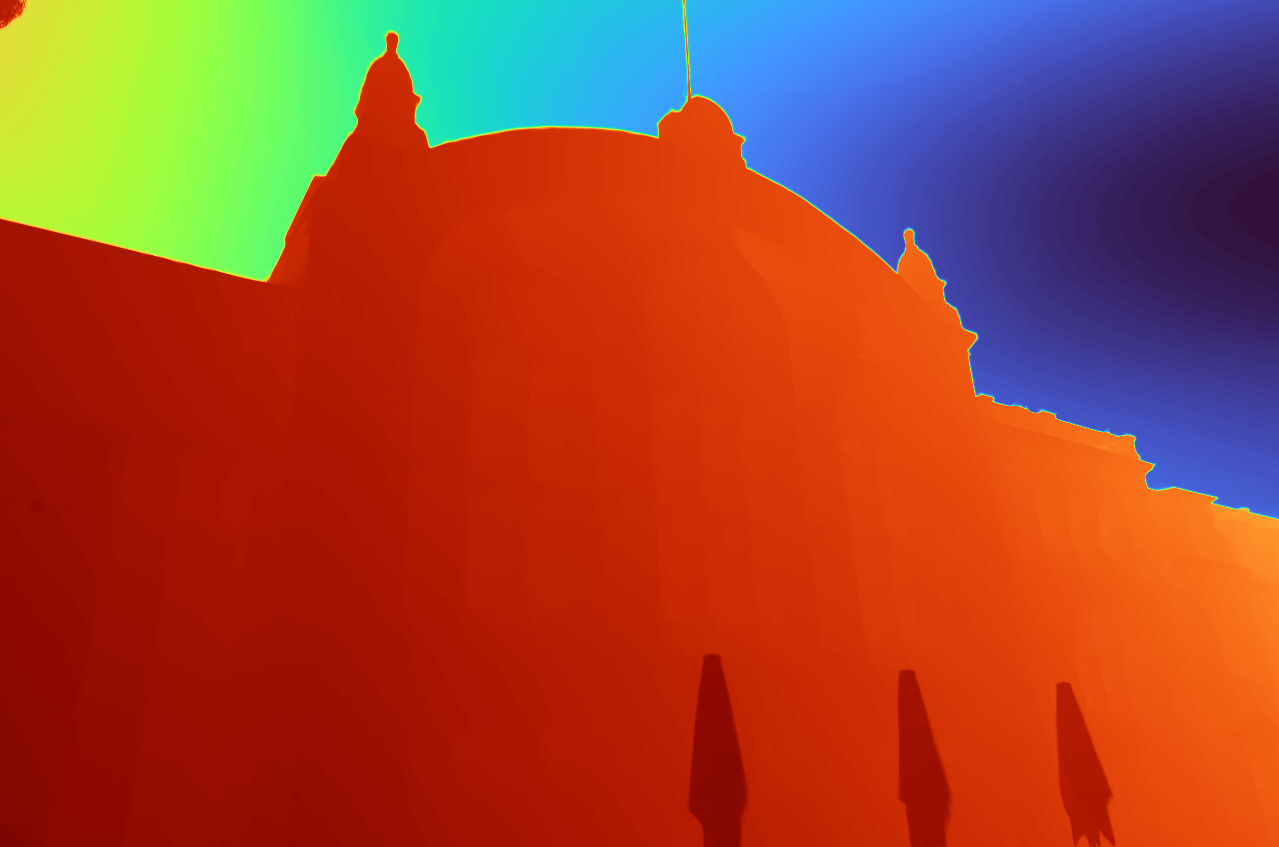} \&
        \includegraphics[width=.8\linewidth]{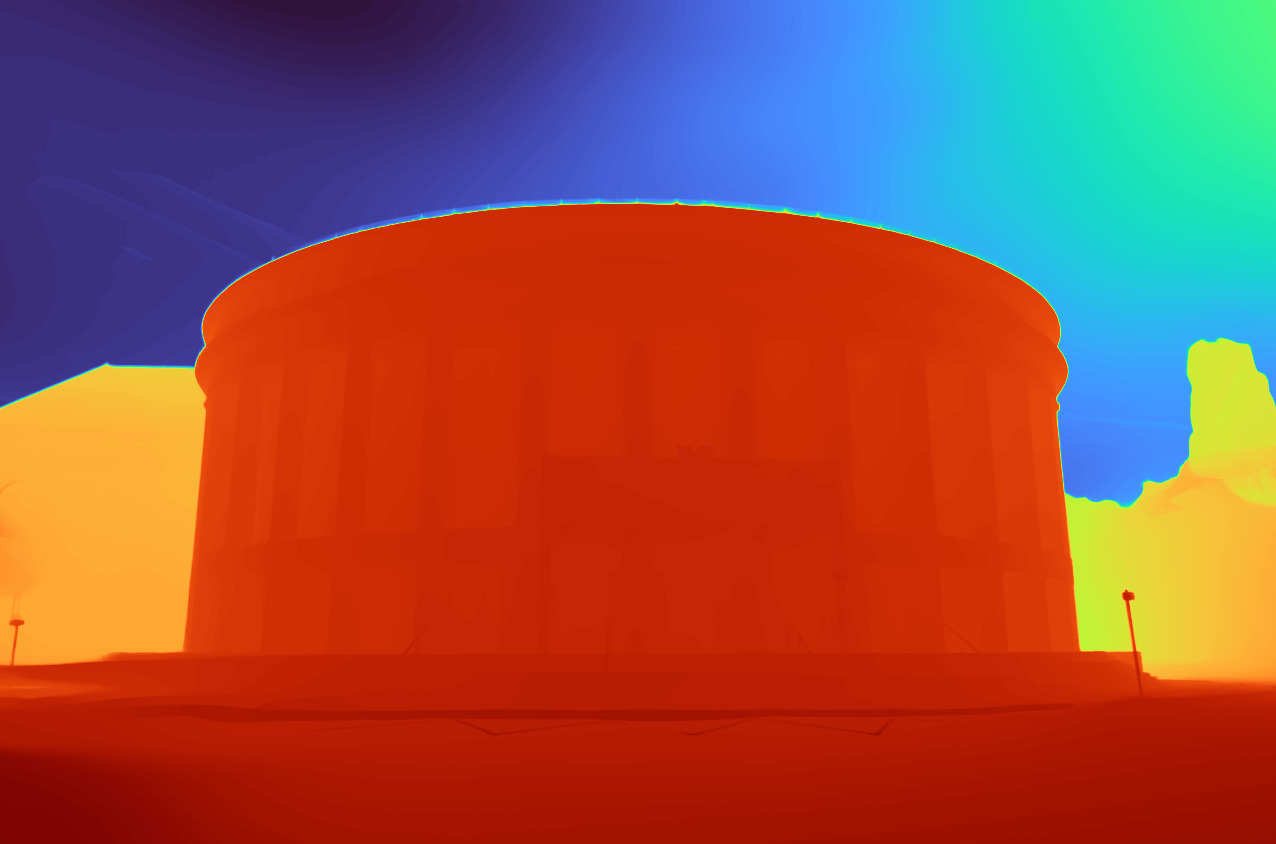} \\
        \includegraphics[width=.8\linewidth]{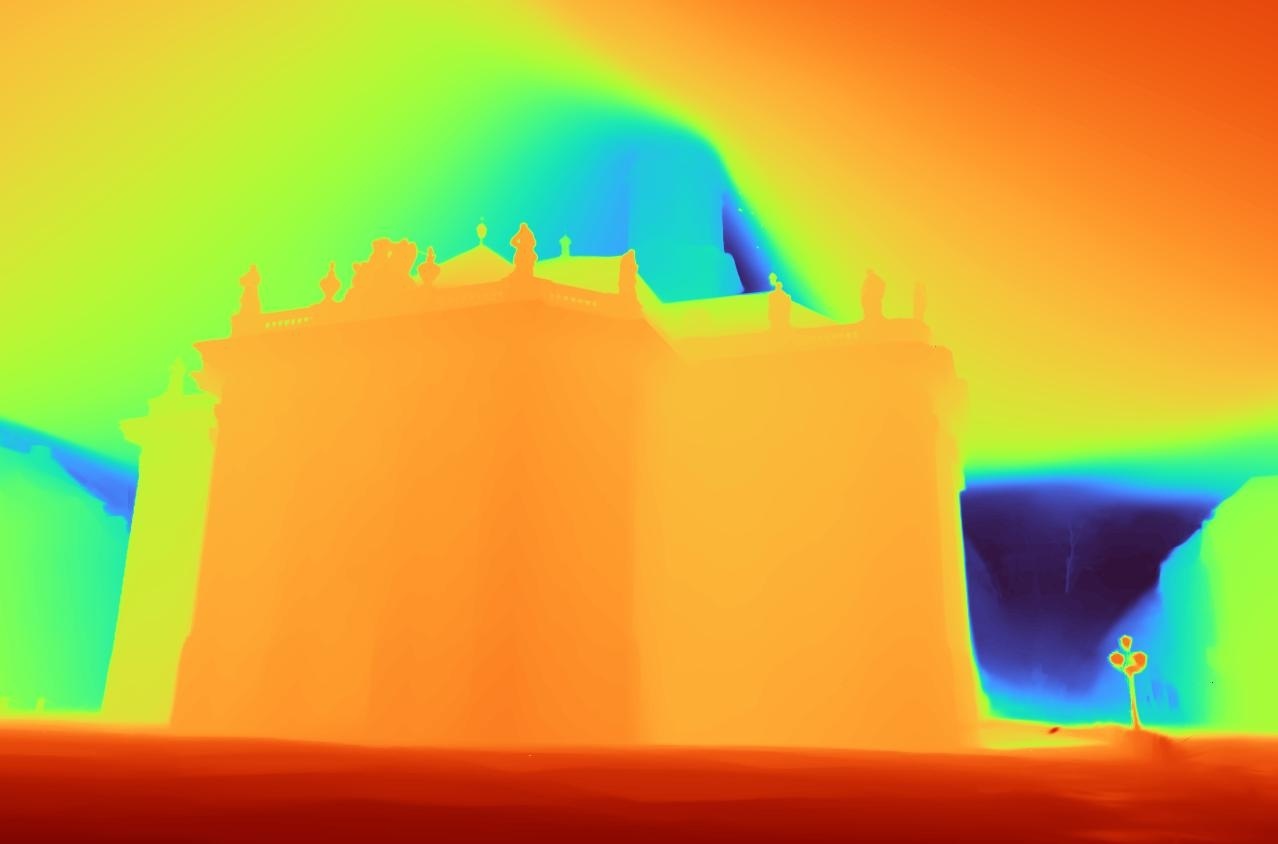} \&
        \includegraphics[width=.8\linewidth]{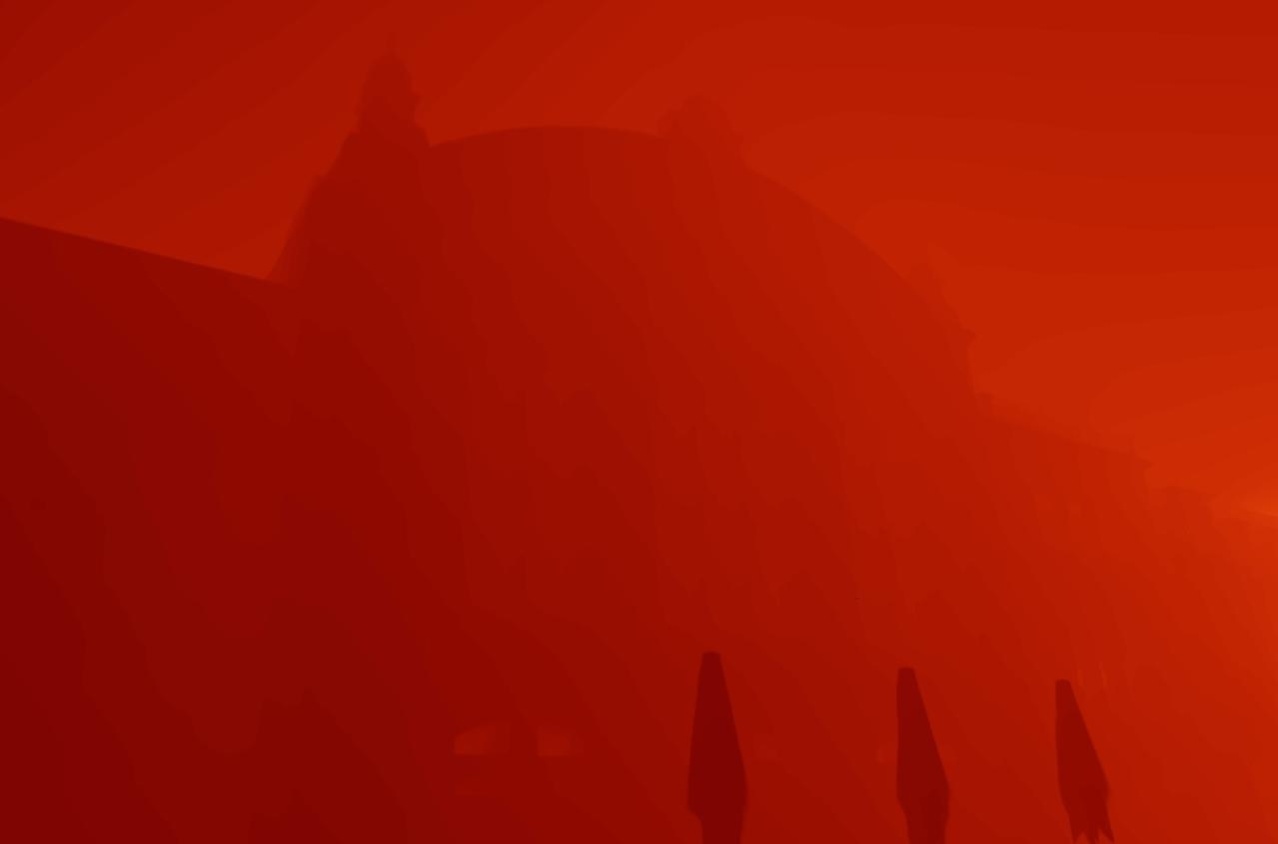} \&
        \includegraphics[width=.8\linewidth]{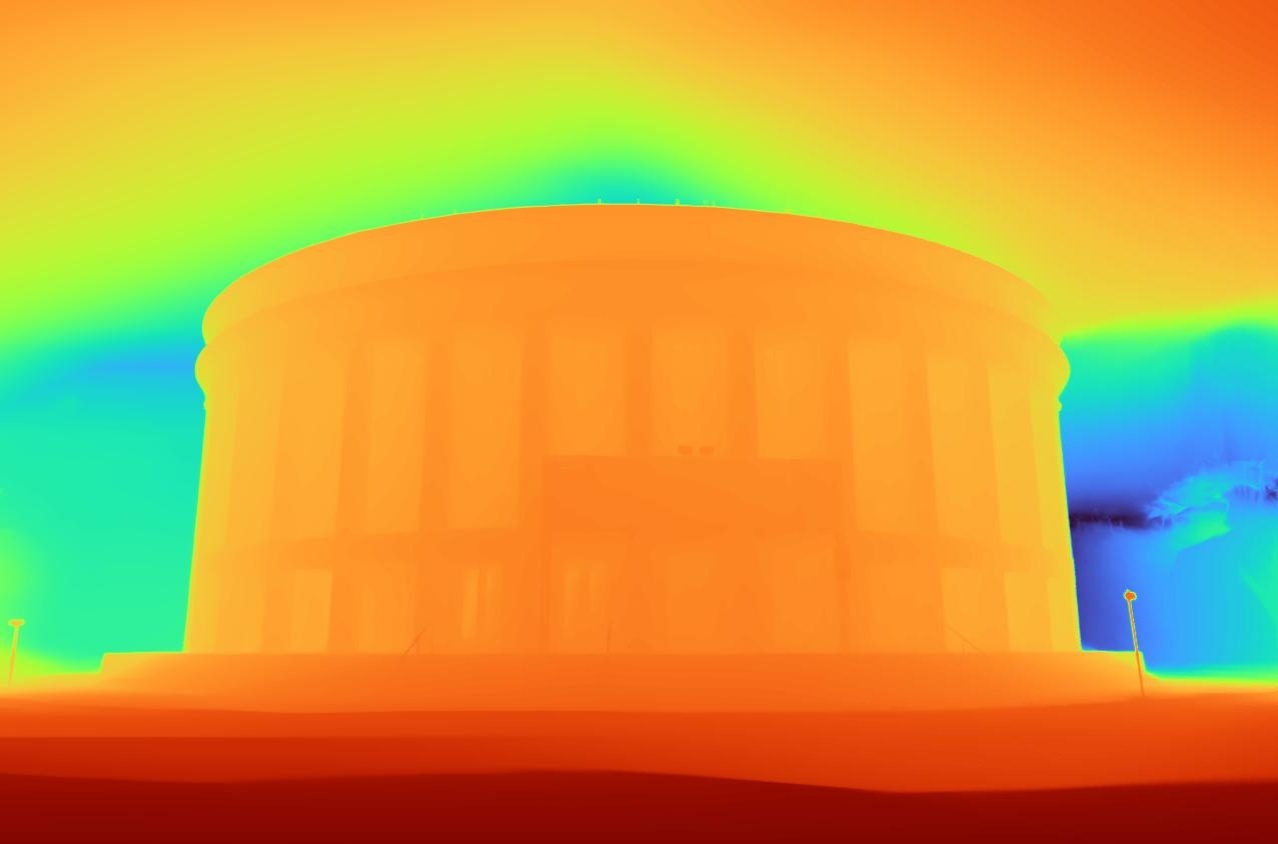}\\
    };

    \node[left=\distleft of qualitative-1-1.west, align=center,scale=\scaleleft, font=\Huge]{\rotatebox{90}{\textbf{Ours}}};
    \node[left=\distleft of qualitative-2-1.west, align=center,scale=\scaleleft, font=\Huge]{\rotatebox{90}{\textbf{LumiGauss}}};

\end{tikzpicture}
}
}

\begin{figure}[b]
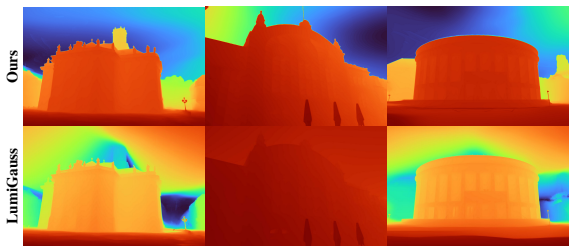

    \centering
    \captionsetup{belowskip=0pt}
    \begin{adjustbox}{center}
    \qualitativecompdepth
    \end{adjustbox}
    \caption{Qualitative comparison of the depth maps predicted by our method and LumiGauss \cite{kaleta2025lumigauss} for the NeRF-OSR scenes. For each novel view, the ground truth is visualized in Fig.~\ref{fig:qualitativecomp_renders}. Compared to LumiGauss, R3GW produces more geometrically consistent depth maps with significantly reduced artifacts in the sky and in the distant regions of the foreground.}
    \label{fig:qualitativecomp_depth_maps}
\end{figure}

\def\height{64px}
\def\spysize{80px}
\def\offset{36px}
\def\scale{0.7}
\def\scaleleft{0.7}
\def\distabove{0.3em}
\def\distleft{0.3em}

\newcommand{\qualitativecompskyrend}{
\centering
\resizebox{\linewidth}{!}{
    \begin{tikzpicture}[
        >=stealth',
        title/.style={
            anchor=base,
            align=center,
            scale=\scale,
        },
        spy using outlines={rectangle, red, magnification=4, every spy on node/.append style={opacity=0}, size=\spysize}
    ]
    \matrix[matrix of nodes, column sep=0pt, row sep=0pt, ampersand replacement=\&, inner sep=0, outer sep=0, font=\Large] (qualitative) {
        \includegraphics[width=.8\linewidth]{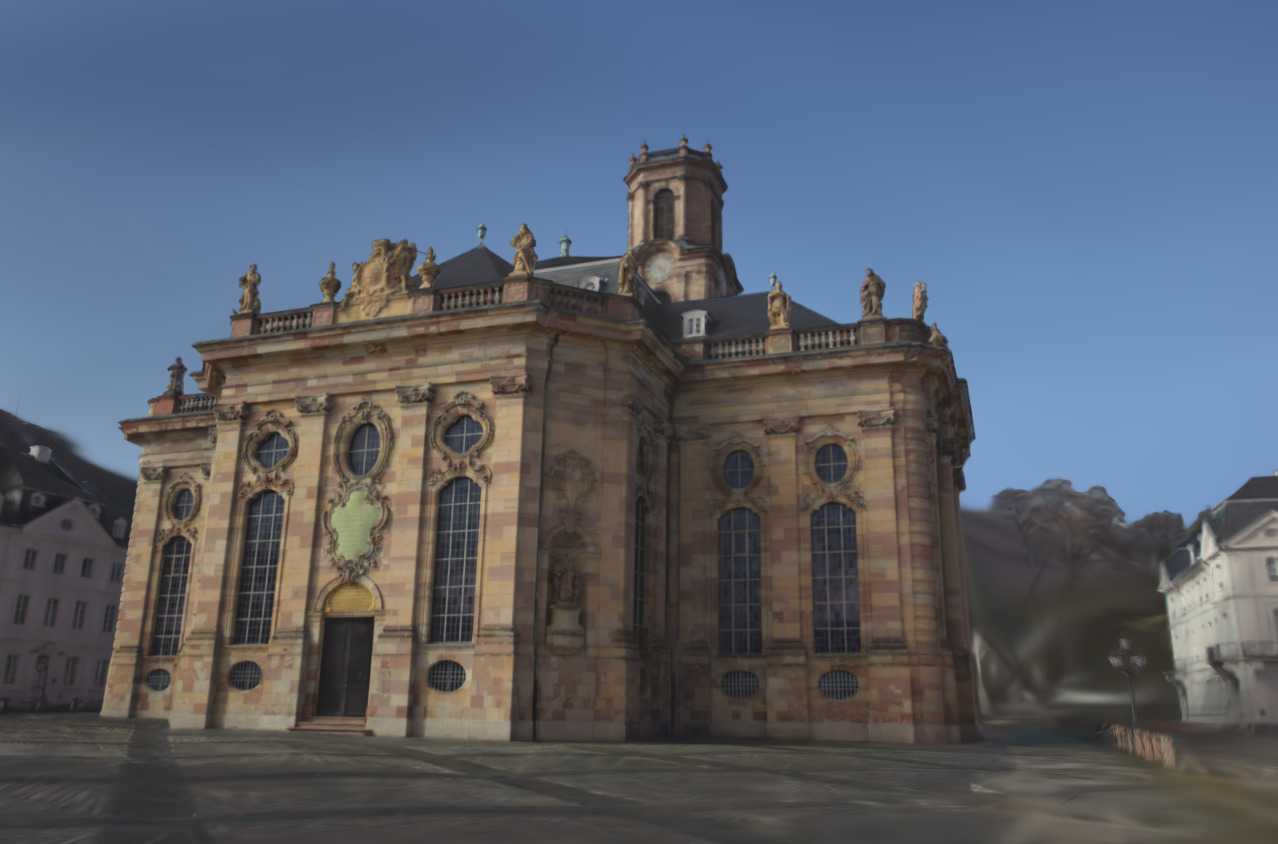} \&
        \includegraphics[width=.8\linewidth]{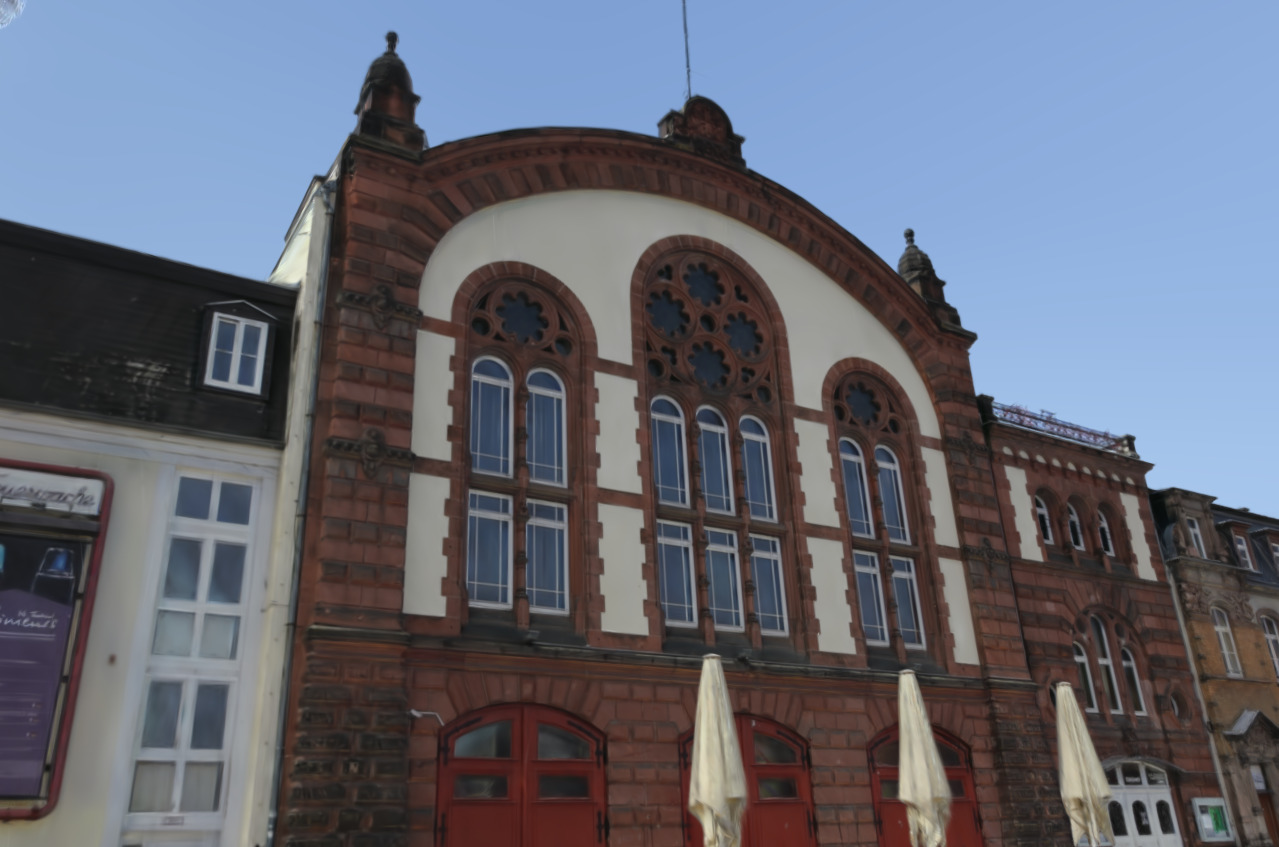} \&
        \includegraphics[width=.8\linewidth]{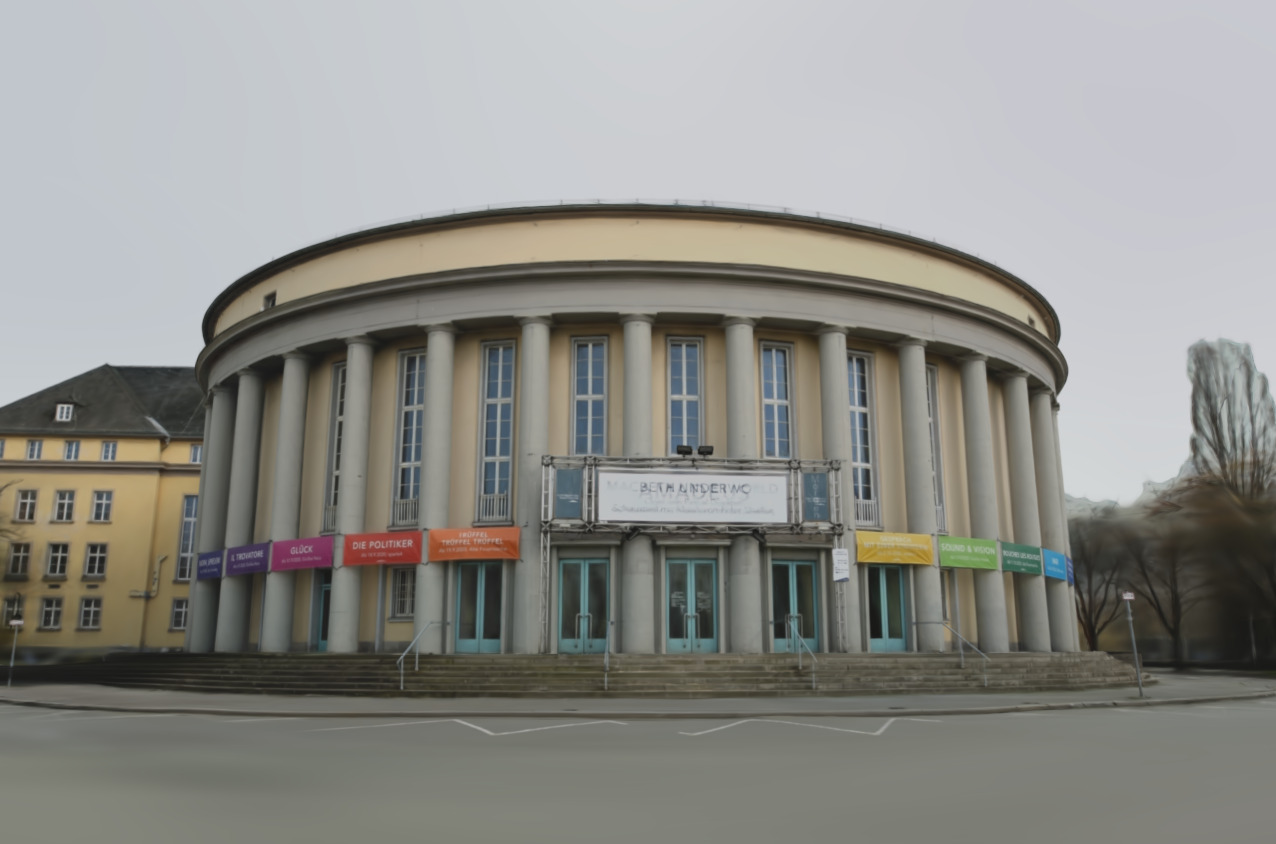} \\
        \includegraphics[width=.8\linewidth]{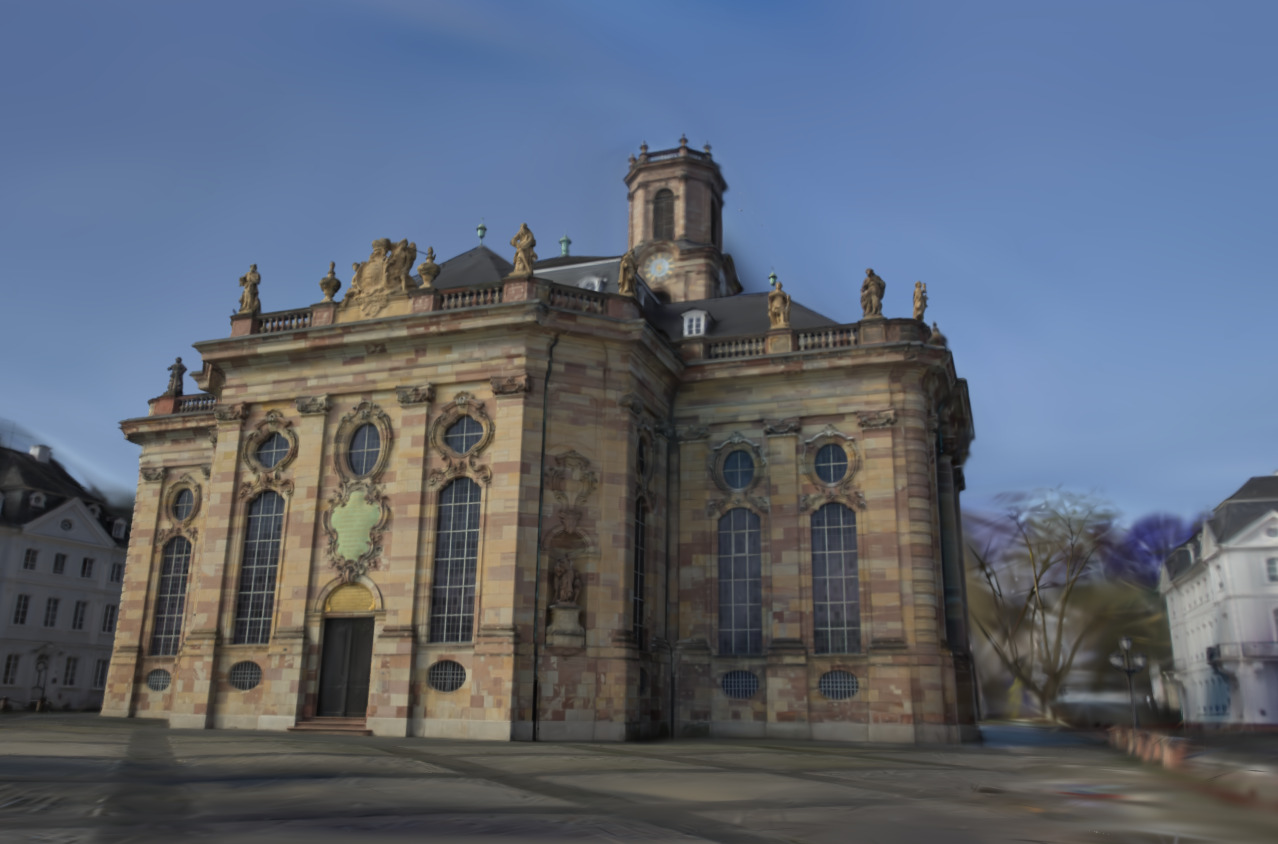} \&
        \includegraphics[width=.8\linewidth]{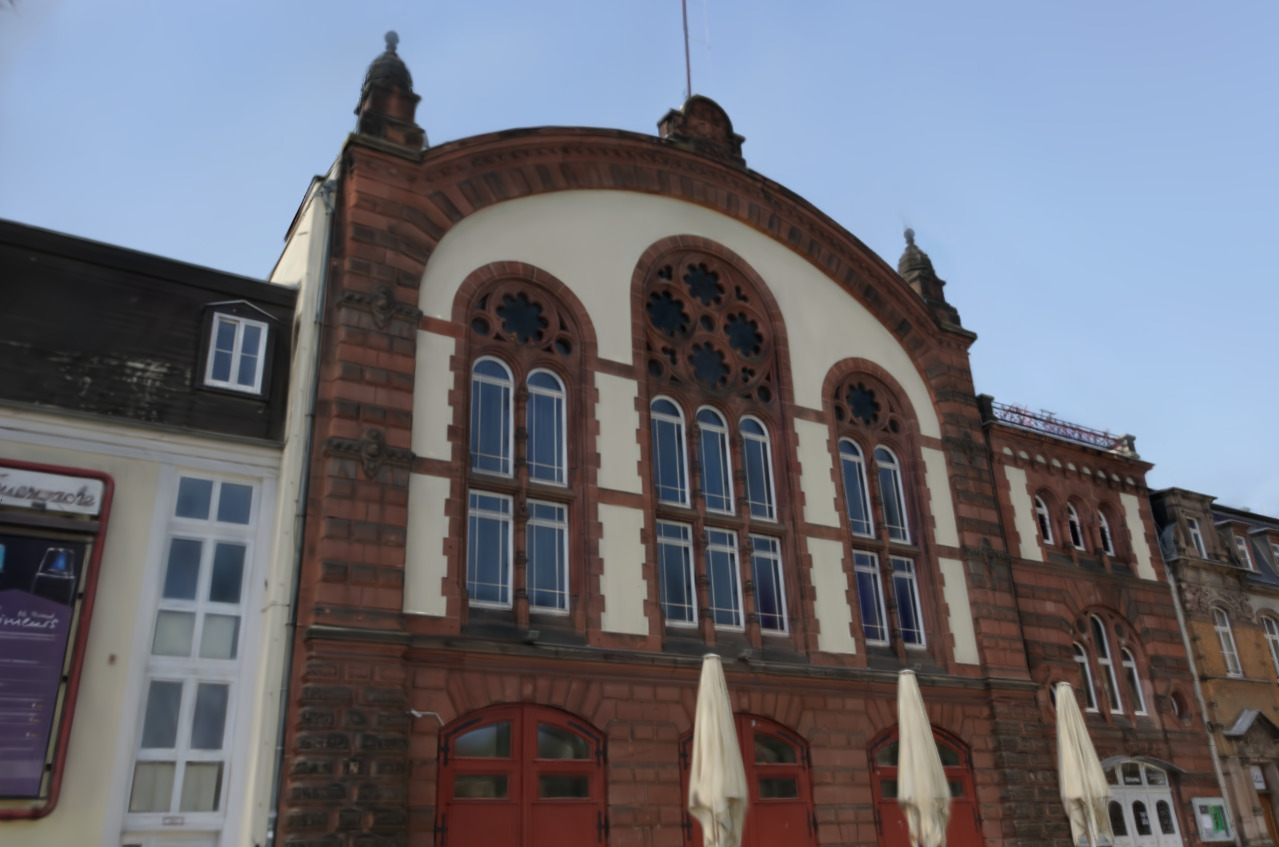} \&
        \includegraphics[width=.8\linewidth]{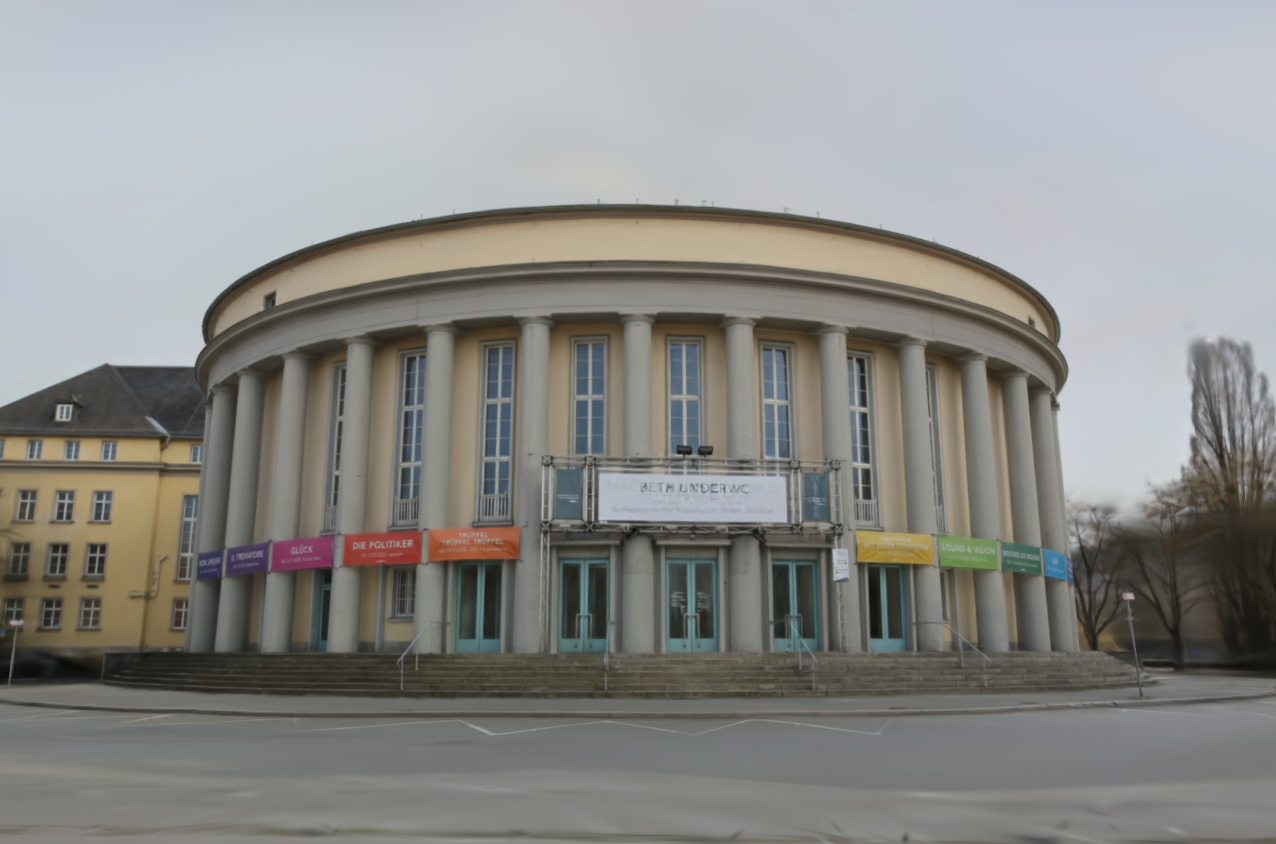}\\
    };

    \node[left=\distleft of qualitative-1-1.west, align=center,scale=\scaleleft, font=\Huge]{\rotatebox{90}{\textbf{Ours}}};
    \node[left=\distleft of qualitative-2-1.west, align=center,scale=\scaleleft, font=\Huge]{\rotatebox{90}{\textbf{LumiGauss}}};

    \foreach \n in {1,...,2}{
       \edef\temp{\noexpand\spy[anchor=center,magnification=5] on ([xshift=1.3em,yshift=3.2
        em] qualitative-\n-1.center) in node  [anchor=south east] at (qualitative-\n-1.south east);}
        \temp
    }

    \foreach \n in {1,...,2}{
       \edef\temp{\noexpand\spy[anchor=center,magnification=5] on ([xshift=3.5em,yshift=2.7
        em] qualitative-\n-2.center) in node  [anchor=south east] at (qualitative-\n-2.south east);}
        \temp
    }
    \foreach \n in {1,...,2}{
       \edef\temp{\noexpand\spy[anchor=center,magnification=3] on ([xshift=1.5em,yshift=3.2
        em] qualitative-\n-3.center) in node  [anchor=south east] at (qualitative-\n-3.south east);}
        \temp
    }

\end{tikzpicture}
}
}

\begin{figure}[t]
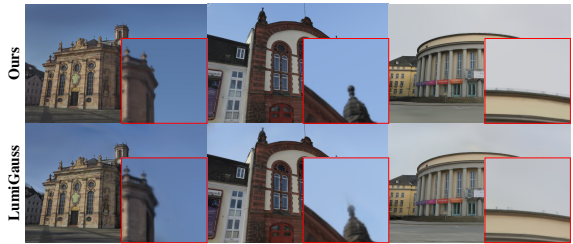

    \centering
    \captionsetup{belowskip=0pt}
    \begin{adjustbox}{center}
    \qualitativecompskyrend
    \end{adjustbox}
    \caption{Qualitative comparison of rendering quality at the sky-foreground boundary between our method and LumiGauss \cite{kaleta2025lumigauss} for the NeRF-OSR scenes. For a clear visualization, we render the novel views for both R3GW and LumiGauss using trained illuminations. Our sky Gaussians are rendered using the corresponding trained sky color SH. R3GW yields sharper building outlines with fewer halos as a result of an improved depth reconstruction (see Fig.~\ref{fig:qualitativecomp_depth_maps}).}
    \label{fig:qualitativecomp_skyrend}
\end{figure}

\subsection{Losses}
\label{sec:loss}

The learnable parameters of our scene representation are the attributes of the sky and foreground Gaussians (except for the color and the boolean identifier), along with the appearance embeddings and MLP parameters. These are optimized by minimizing the reconstruction loss of 3DGS introduced in Sec.~\ref{sec:preliminaries_3DGS}, $\mathcal{L}_{\text{rec}}$,  combined with the following regularizations.

To constrain the environment light to take positive values, we use the regularization introduced by \cite{kaleta2025lumigauss}:
\begin{equation}\label{eq:loss_light}
    \mathcal{L}_{\text{light}} = \frac{1}{M}\sum_{i=1}^{M}\lVert \min(0, L(\boldsymbol{\omega_i})) \rVert^{2}_{2},
\end{equation}
where $\{\omega_i\}_{i=1}^{M}$ are incident light directions sampled uniformly on the unit upper hemisphere.

To align the normals of the foreground Gaussians with the surface normals, we use the normal consistency loss of \cite{huang20242d}:
\begin{equation}\label{eq:normal_loss}
    \mathcal{L}_{\text{normal}}=\sum_{i=1}^{N}\alpha_i(1-\mathbf{n}_i^{T}\mathbf{N}),
\end{equation}
where $\{\alpha_i\}_{i=1}^N$ are the blending weights of Eq.~(\ref{eq:3DGS_col}), $\mathbf{n}_i$ is the normal of the visible foreground Gaussian $i$, and $\mathbf{N}$ is the reference surface normal estimated from the gradient of the depth map. Furthermore, to encourage flattening of the foreground Gaussians, we apply the following loss from \cite{chen2023neusg} to the smallest scaling coefficient:
\begin{equation}\label{eq:scales_loss}
    \mathcal{L}_{\text{scale}} = \lVert \min(s_x,s_y, s_z)\rVert_1.
\end{equation}

The sky Gaussians must be guided to contribute only to the rendering of the sky, while the foreground Gaussians should not be moved in the sky region. Otherwise, the model may learn the scene illumination based on physically implausible BRDF features and sky normals. Therefore, we apply the following loss to the sky and foreground pixels of the renderings  $\hat{I}_{\text{fg}}$ and $\hat{I}_{\text{sky}}$, which are obtained by rendering all Gaussians with the foreground Gaussians’ colors set to zero in $\hat{I}_{\text{sky}}$, and the colors of the sky Gaussians set to zero in $\hat{I}_{\text{fg}}$:
\begin{equation}\label{eq:loss_sky_brdf}
    \mathcal{L}_{\text{fg-sky}} = \mathcal{L}_1((1 - M_{\text{sky}})\hat{I}_{\text{fg}}, \mathbf{0}) + \mathcal{L}_1(M_{\text{sky}}\hat{I}_{\text{sky}}, \mathbf{0}),
\end{equation}
where $M_{\text{sky}}$ is a binary sky segmentation mask corresponding to the view. 

We allow the background to move only farther away from the foreground. To enforce this, we introduce the following loss to penalize the case in which $\bar{d}_{\text{foreground}}$, the average depth of the visible foreground Gaussians, exceeds the average depth of the visible sky Gaussians $\bar{d}_{\text{sky}}$:
\begin{equation}\label{eq:loss_depth}
    \mathcal{L}_{\text{sky-depth}} = e^{-\gamma(\bar{d}_{\text{sky}}-\bar{d}_{\text{foreground}})}, \ \gamma>0.
\end{equation}

In summary, the training objective function is:
\begin{multline}\label{eq:final_loss}
    \mathcal{L}= \mathcal{L}_{\text{rec}} 
    + \lambda_{\text{light}} \mathcal{L}_{\text{light}} 
    + \lambda_{\text{normal}} \mathcal{L}_{\text{normal}} \\
    + \lambda_{\text{scale}} \mathcal{L}_{\text{scale}} 
    + \lambda_{\text{fg-sky}} \mathcal{L}_{\text{fg-sky}} 
    + \lambda_{\text{sky-depth}} \mathcal{L}_{\text{sky-depth}},
\end{multline}
where we set:  $\lambda_{\text{light}} = 1.0$, $\lambda_{\text{normal}} = 0.05$, $\lambda_{\text{scale}} = 1.0$, $\lambda_{\text{fg-sky}} = 0.5$, $\lambda_{\text{sky-depth}} = 0.005$.

\section{\uppercase{Experiments}}
\label{sec:experiments}
\begin{table*}[!t]
\caption{Quantitative comparison of our method with the baselines on the NeRF-OSR test scenes. 
For LumiGauss \cite{kaleta2025lumigauss} and NeRF-OSR \cite{rudnev2022nerf} the $*$ denotes the results presented in the papers by the authors.
We group methods into NeRF-based and Gaussian Splatting (GS)-based approaches, as NeRF models typically achieve higher visual fidelity but are more computationally expensive to train and render.
The last section shows the evaluation results of the ablation studies described in Sec.~\ref{sec:ablation}.}
\label{tab:eval_metrics}
\centering
\newcommand{\na}{-}
\NewDocumentCommand{\B}{}{\fontseries{b}\selectfont}
\resizebox{\textwidth}{!}{
\begin{tabular}{|l|c|c|c|c|c|c|c|c|c|c|c|c|c|c|c|c|}
\hline
\multirow{2}{*}{\textbf{Method}} 
& \multicolumn{4}{c|}{\textbf{PSNR}$\uparrow$} 
& \multicolumn{4}{c|}{\textbf{SSIM}$\uparrow$} 
& \multicolumn{4}{c|}{\textbf{MSE}$\downarrow$}
& \multicolumn{4}{c|}{\textbf{MAE}$\downarrow$} \\
\cline{2-17}
& lk2 & lwp & st & avg 
& lk2 & lwp & st & avg 
& lk2 & lwp & st & avg 
& lk2 & lwp & st & avg \\
\hline
\multicolumn{17}{|l|}{\textbf{NeRF-based}} \\ \hline
$\text{NeRF-OSR}^{*}$ \cite{rudnev2022nerf} 
& 18.72 & 16.65 & 15.43 & 16.93 
& 0.468 & 0.501 & 0.517 & 0.495
& 0.014 & 0.024 & 0.029 & 0.022
& \B 0.090 & 0.114 & 0.113 & 0.106 \\
\hline
NeRF-OSR \cite{rudnev2022nerf} 
& 18.80 & 15.30 & 15.59 & 16.56
& 0.501 & 0.514 & 0.511 & 0.509
&  0.011 & 0.034 & 0.027 & 0.024 
& 0.090 & 0.132 & \B 0.111 & 0.111\\
\hline
SR-TensoRF \cite{chang2024fast}
& 17.30 & 16.74 & 15.63 & 16.56
&  0.542 & \B 0.653 & 0.632 & 0.609
& 0.021 & 0.024 & 0.030 & 0.025
& 0.096 & \B 0.093 & 0.111 & \B 0.100 \\
\hline
FEGR \cite{wang2023fegr}
& 21.53 & 17.57 & 17.00 & 18.70
&  \na & \na & \na & \na 
& 0.007 & 0.018 & 0.023 & 0.016
& \na & \na & \na & \na\\
\hline
SOL-NeRF \cite{sun2023sol}
& 21.23 & 17.58 & \B 18.18 & 19.00
& \B 0.749 & 0.618 & \B 0.680 & \B 0.682
& 0.008 & 0.028 & \B 0.019 & 0.018
& - & - & - & - \\
\hline
NeuSky \cite{gardner2024sky}
& \B22.50 & \B 18.31 & 16.66 & \B 19.16
&  \na & \na & \na & \na
& \B 0.005 & \B 0.016 & 0.023 & \B 0.015
& \na & \na & \na & \na\\
\noalign{\hrule height 1.5pt}
\multicolumn{17}{|l|}{\textbf{GS-based}} \\ \hline
$\text{LumiGauss}^{*}$ \cite{kaleta2025lumigauss}
&  19.59  &  18.01  &  \B 17.02  & \B 18.21 
& 0.700 &  \B 0.778 &  \B 0.729 & 0.736
& 0.012 &  \B 0.017 & \B  0.021 & \B 0.017
& 0.085  & \B 0.096 &  \B 0.107 & \B 0.096 \\
\hline
LumiGauss \cite{kaleta2025lumigauss}
&  19.37  &  \B 18.18  & 16.89 & 18.15
&  0.703 & 0.774 & 0.724 & 0.734
&  0.012 & 0.017 & 0.022 & 0.017
&  0.085  &  0.096 & 0.111 & 0.097 \\
\hline
LumiGauss w/out shadows \cite{kaleta2025lumigauss}
&  19.03  &  17.76  & 16.34 & 17.71
&  0.693 & 0.761 & 0.709 & 0.721
&  0.013 & 0.018 & 0.024 & 0.018
&  0.090  &  0.103 & 0.122 & 0.105 \\
\hline
Ours & \B 20.18 &  17.93 & 16.22 & 18.11
& \B 0.731 & 0.768 & 0.718 & \B 0.739
& \B 0.010 & 0.018 & 0.025 & 0.018
& \B 0.078 & 0.098 &  0.117 & 0.098 \\
\noalign{\hrule height 1.5pt}
\multicolumn{17}{|l|}{\textbf{Ablations}} \\ \hline
Ours w/out $\mathbf{c}_{\text{specular}}$ (Eq.~(\ref{eq:spec_color})) & 20.03 & 17.38 & 15.35 & 17.59
& 0.722 & 0.760 & 0.706 & 0.729
& 0.011 & 0.020 & 0.032 & 0.021
& 0.080 & 0.102 &  0.128 & 0.103 \\
\hline
Ours w/out $\mathcal{L}_{\text{sky-depth}}$ (Eq.~(\ref{eq:loss_depth})) & 20.08 & 17.86 & 16.19 & 18.04
& 0.728 & 0.761 & 0.717 & 0.735
& 0.011 & 0.019 & 0.025 & 0.018
& 0.079 & 0.101 &  0.116 & 0.099 \\
\hline
\end{tabular}
}
\end{table*}

\def\height{64px}
\def\spysize{42px}
\def\offset{36px}
\def\scale{0.7}
\def\scaleleft{0.7}
\def\distabove{0.3em}

\newcommand{\qualitativecomp}{
\centering
\resizebox{\linewidth}{!}{
    \begin{tikzpicture}[
        >=stealth',
        title/.style={
            anchor=base,
            align=center,
            scale=\scale,
        },
        spy using outlines={rectangle, red, magnification=3, connect spies, size=\spysize}
    ]
    \matrix[matrix of nodes, column sep=0pt, row sep=0pt, ampersand replacement=\&, inner sep=-0.1, outer sep=-0.1, font=\Large] (qualitative) {
        \includegraphics[width=0.3\textwidth]{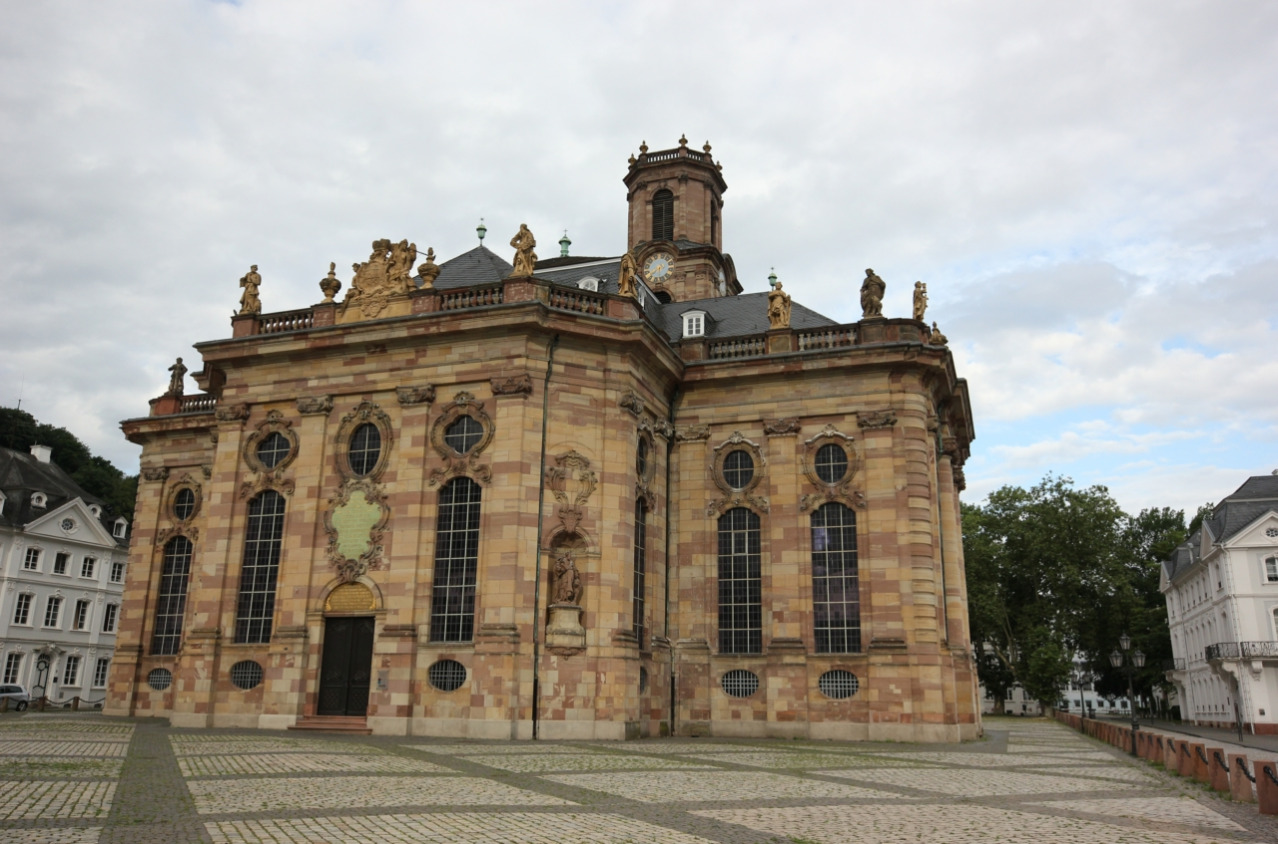} \&
        \includegraphics[width=0.3\textwidth]{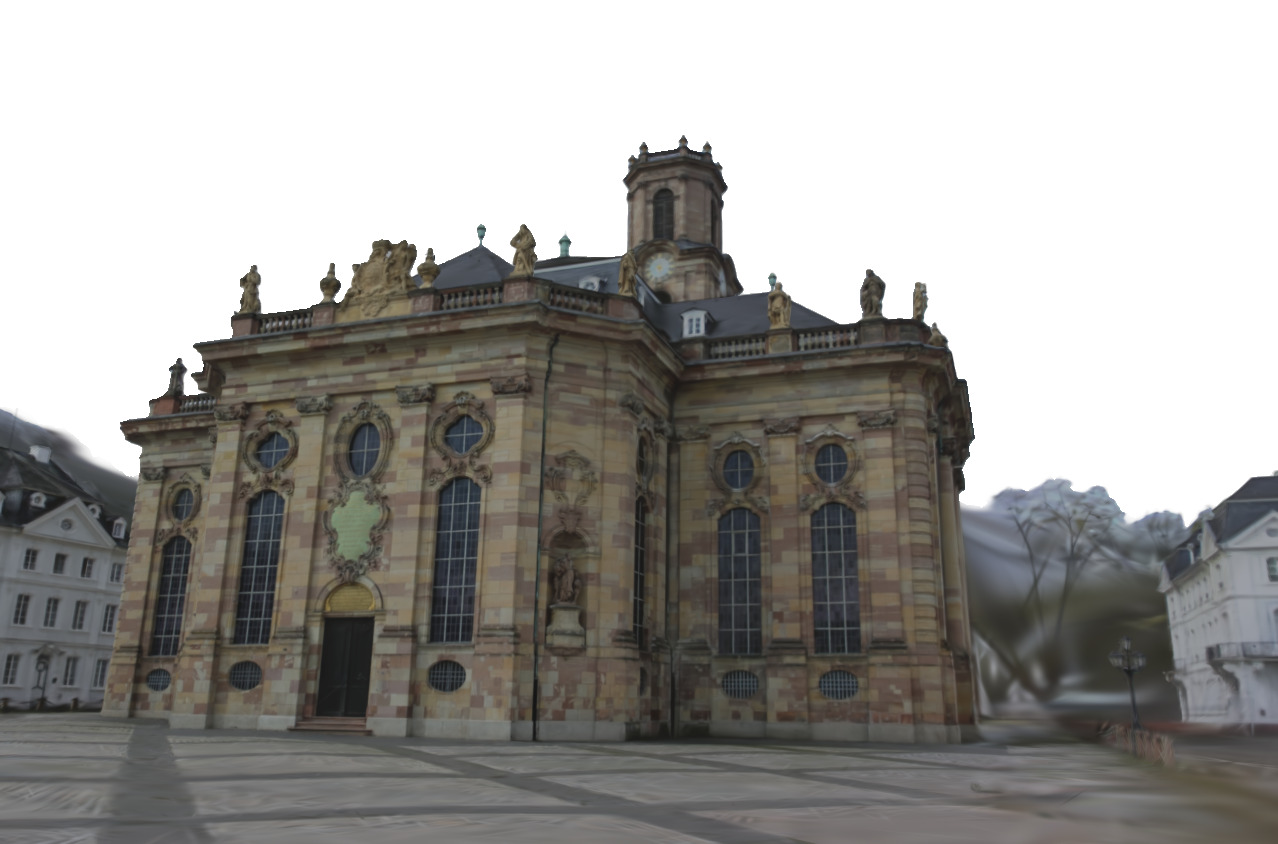} \&
        \includegraphics[width=0.3\textwidth]{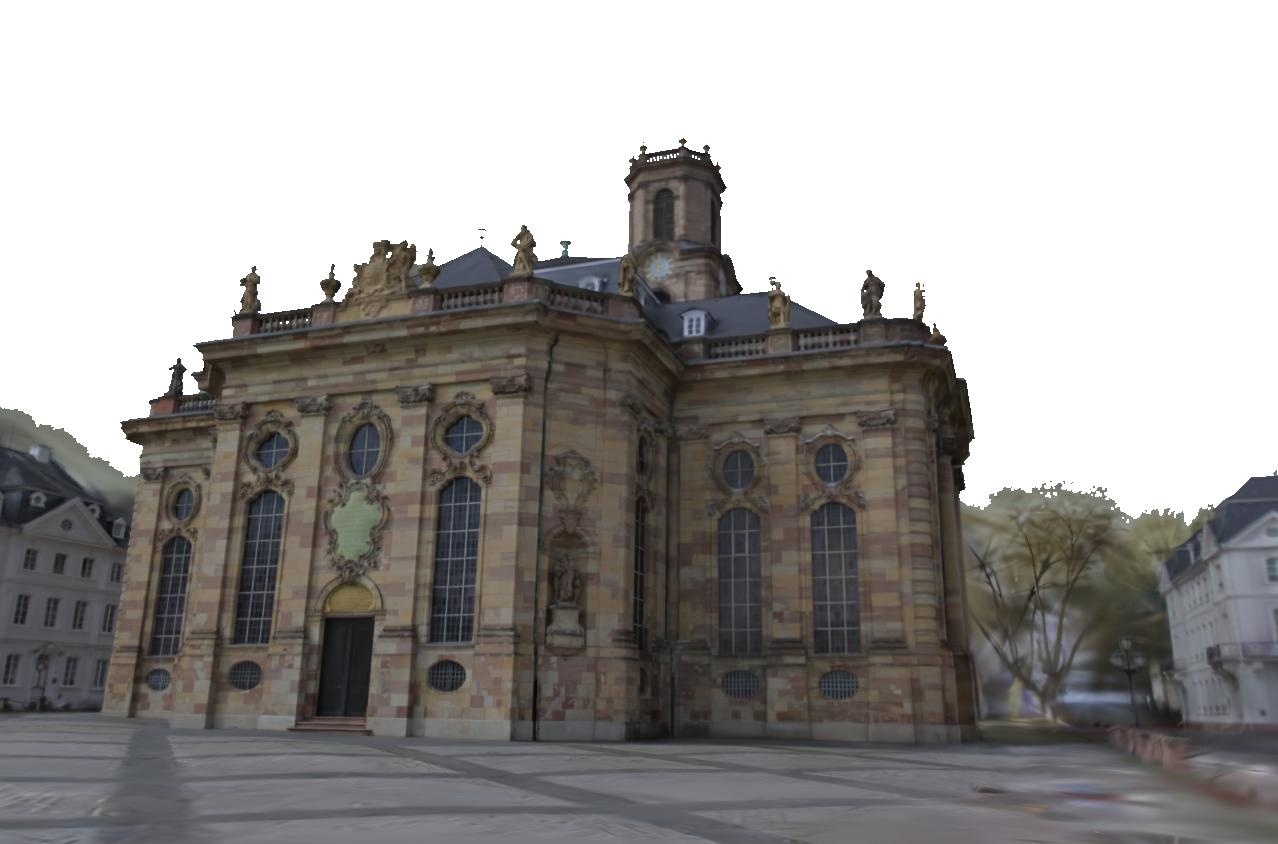} \&
        \includegraphics[width=0.3\textwidth]{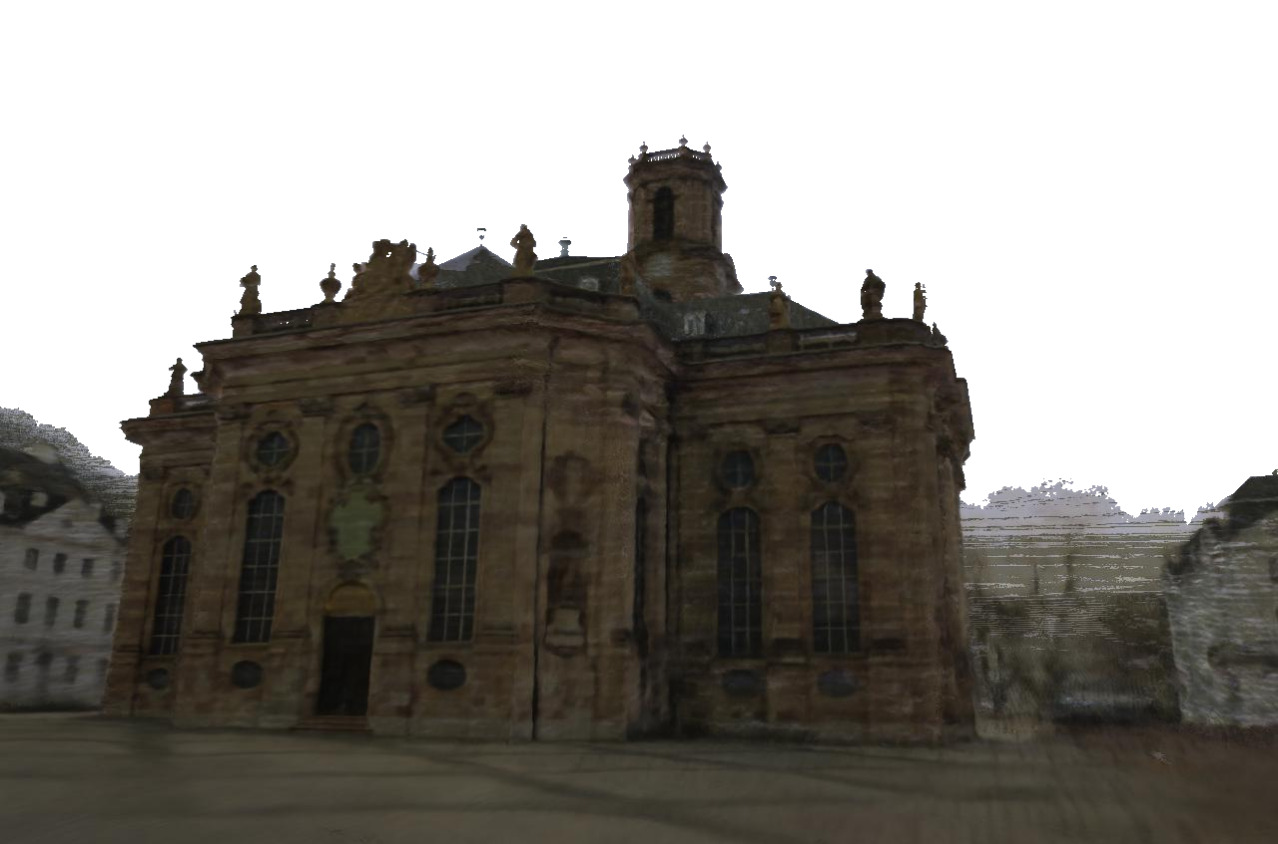} \\
        \includegraphics[width=0.3\textwidth]{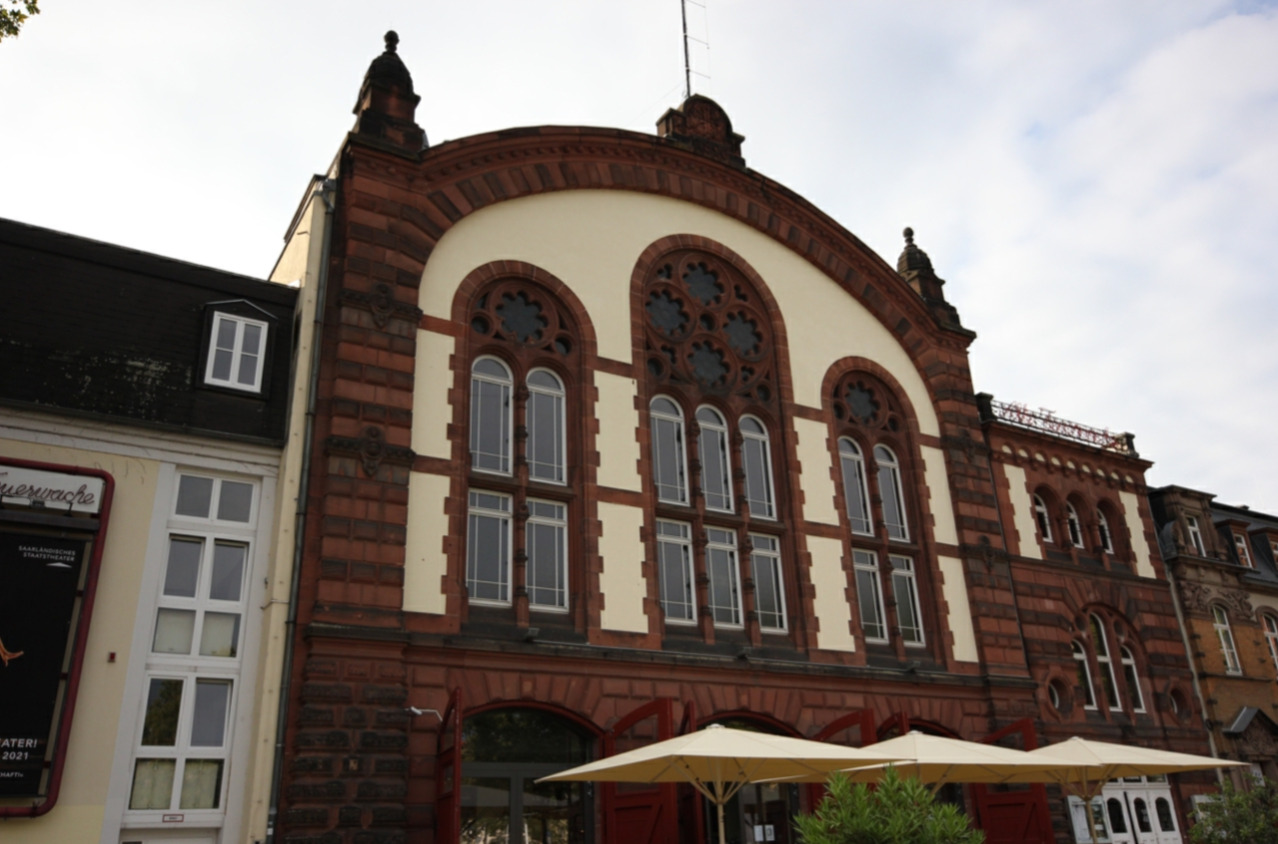} \&
        \includegraphics[width=0.3\textwidth]{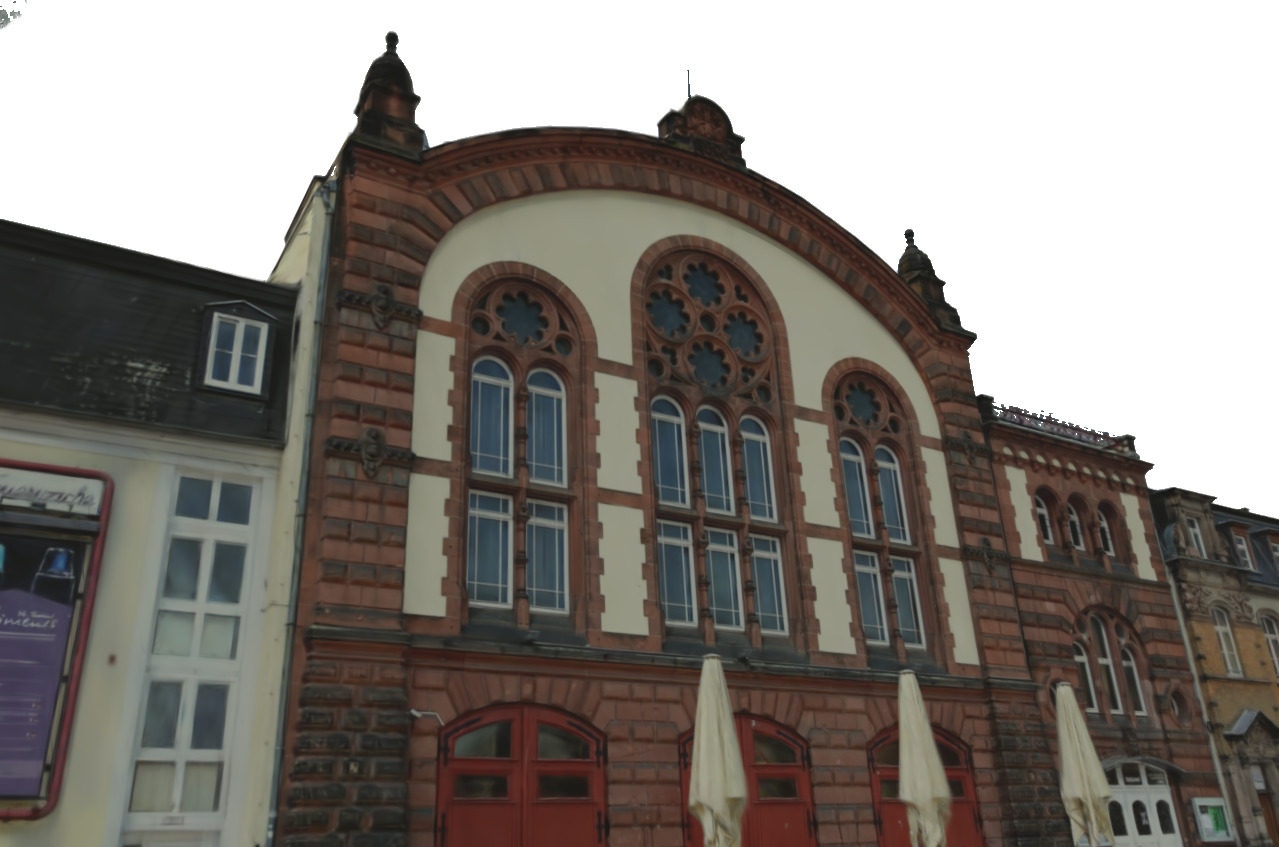} \&
        \includegraphics[width=0.3\textwidth]{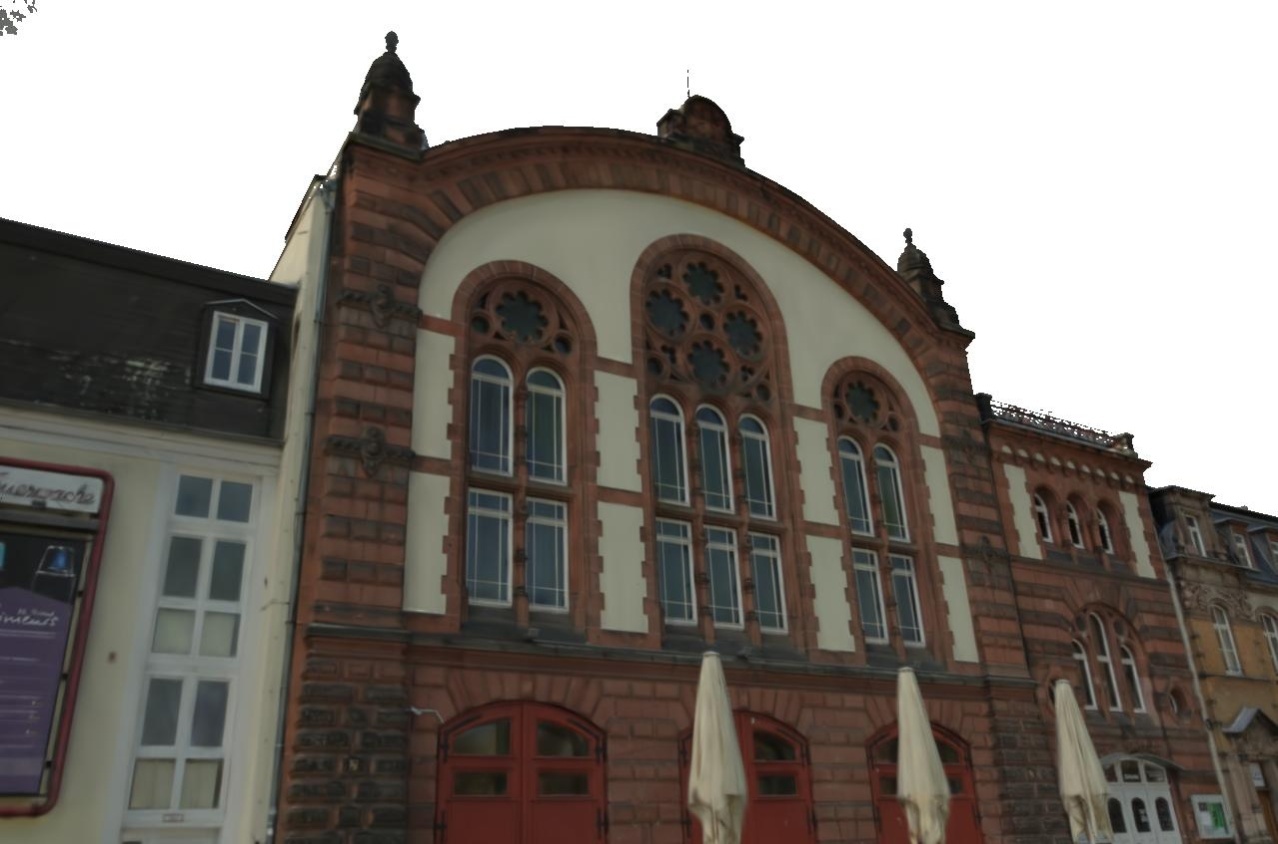}\&
        \includegraphics[width=0.3\textwidth]{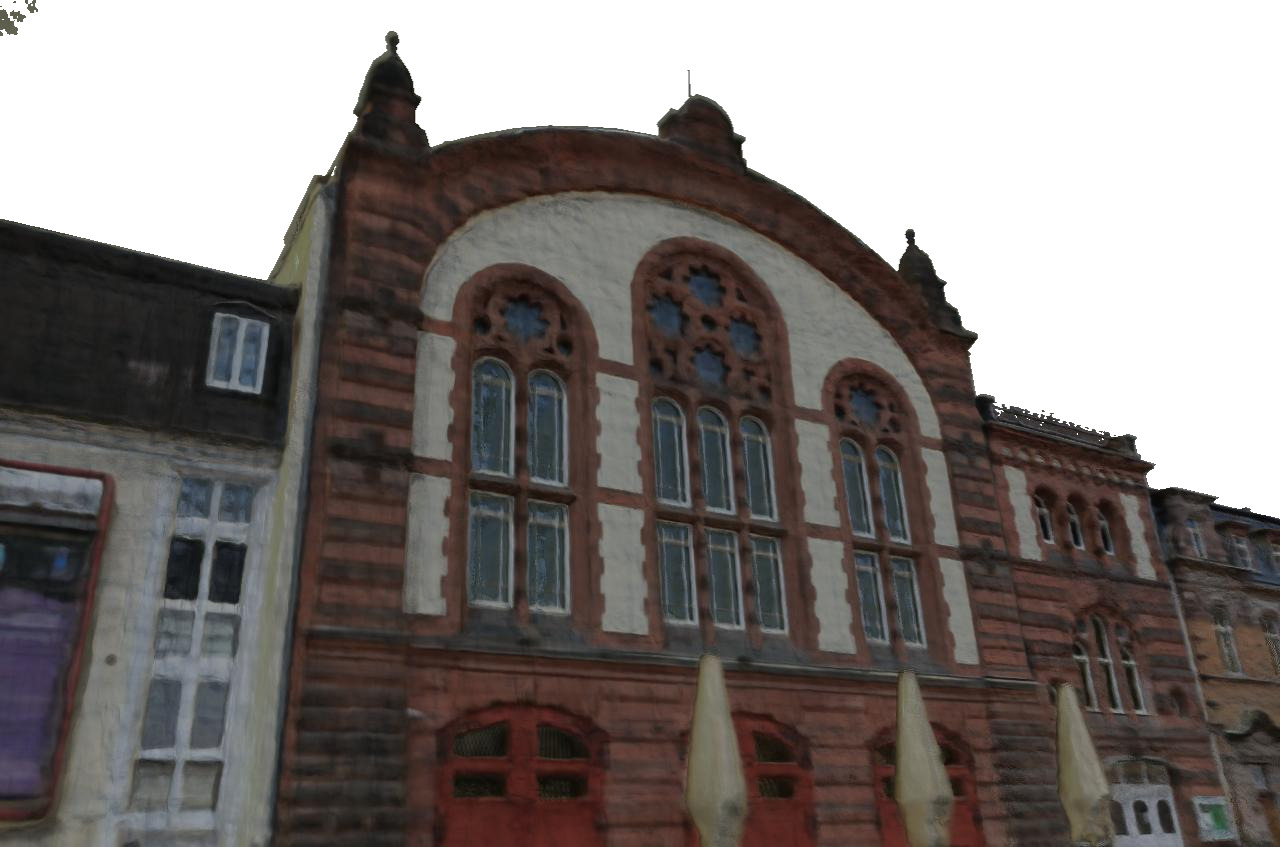} \\
        \includegraphics[width=0.3\textwidth]{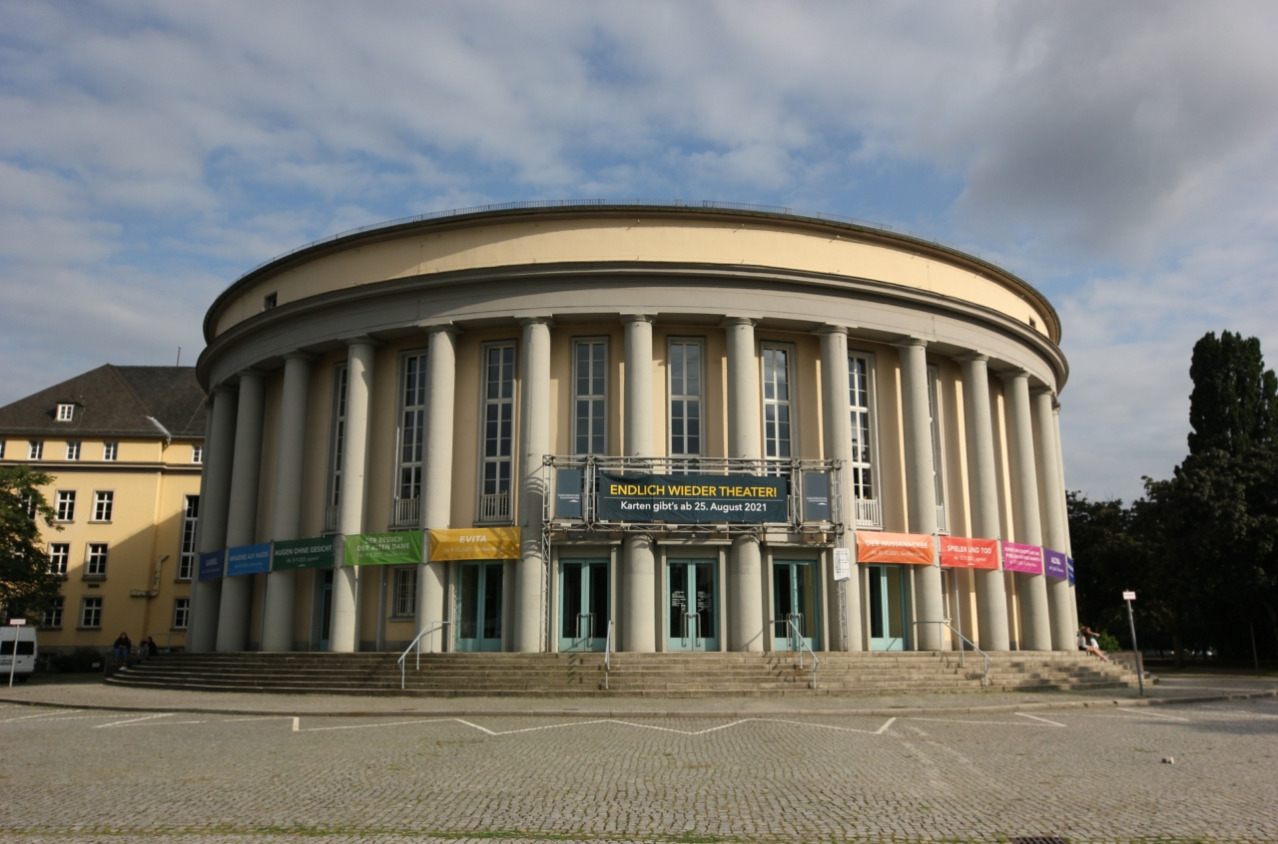}\&
        \includegraphics[width=0.3\textwidth]{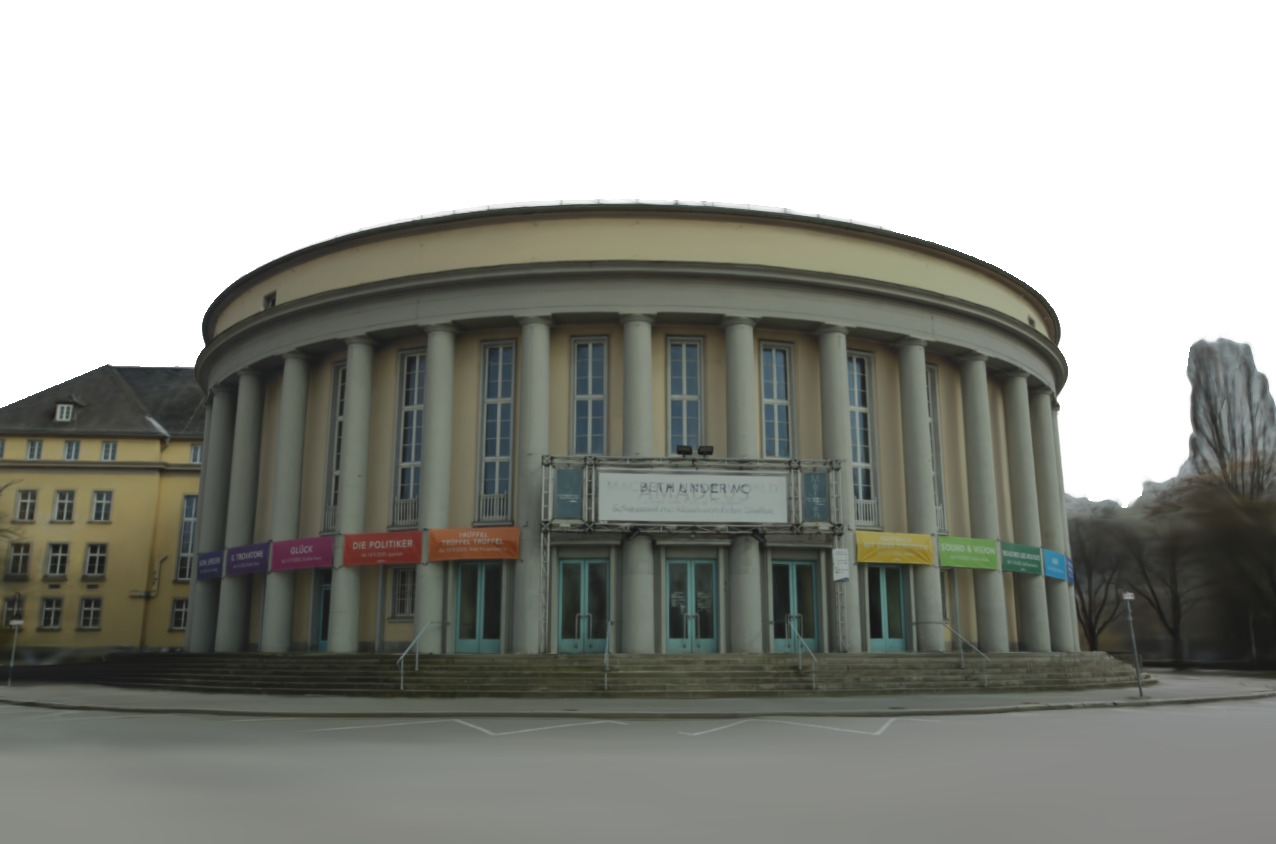} \&
        \includegraphics[width=0.3\textwidth]{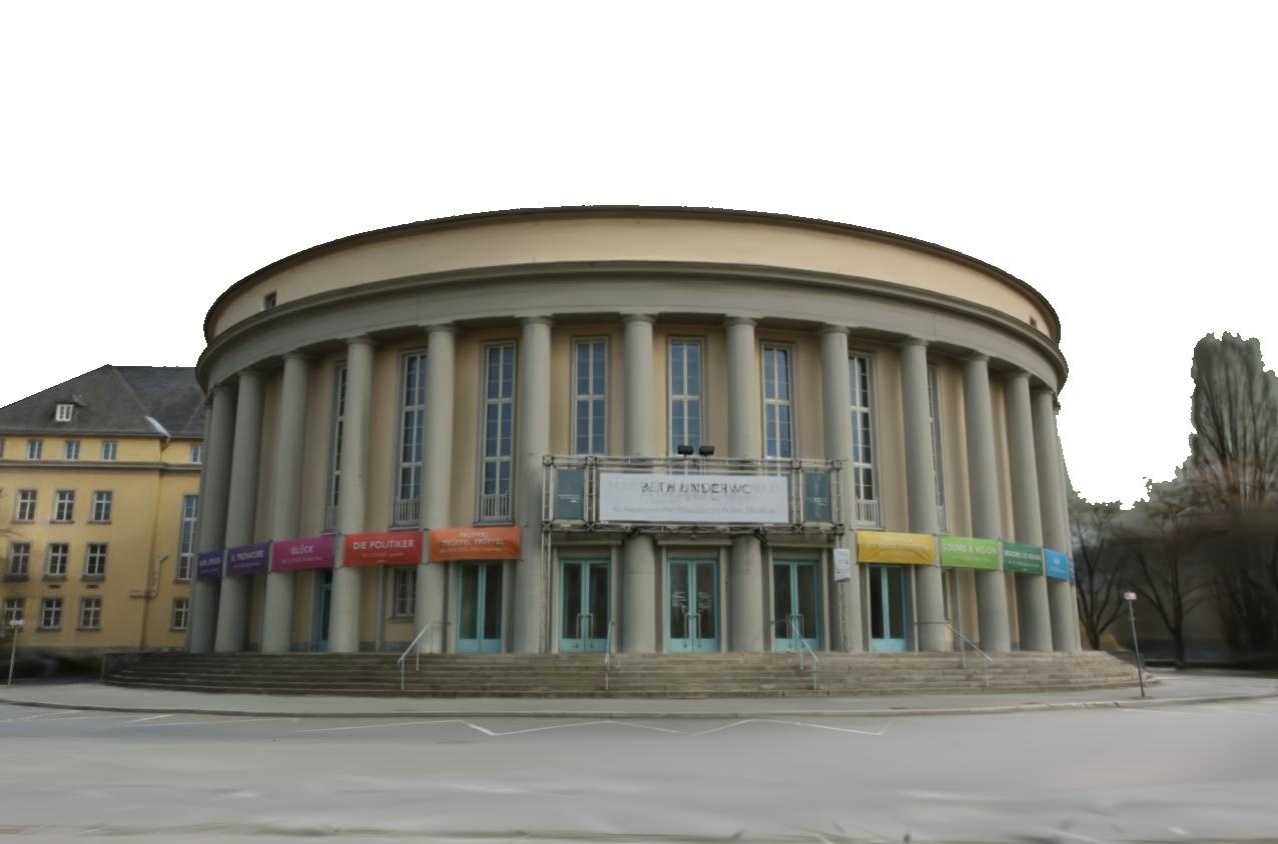} \&
        \includegraphics[width=0.3\textwidth]{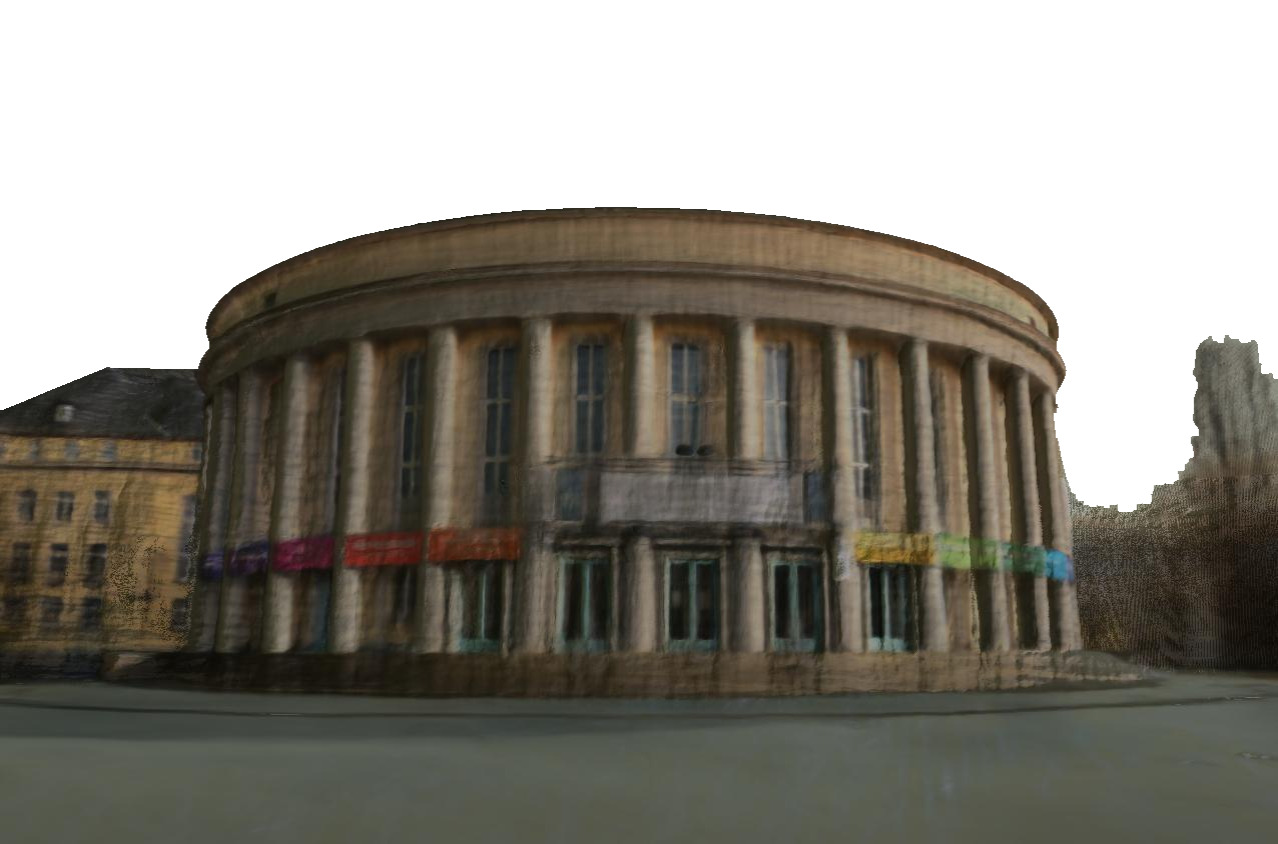} \\
    }; 

    \node[above=\distabove+0.1em of qualitative-1-1.north, title, font=\Large] {Ground Truth};
    \node[above=\distabove+0.1em of qualitative-1-2.north, title, font=\Large] {\textbf{Ours}};
    \node[above=\distabove+0.1em of qualitative-1-3.north, title, font=\Large] {\textbf{LumiGauss}};
    \node[above=\distabove+0.1em of qualitative-1-4.north, title, font=\Large] {\textbf{NeRF-OSR}};

    \foreach \n in {1,...,4}{
        \edef\temp{\noexpand\spy[anchor=center,magnification=3] on ([xshift=-2.0em,yshift=0.0
        em] qualitative-1-\n.center) in node[below left=3.75px and 3.75px] at (qualitative-1-\n.north east);}
        \temp
    }

    \foreach \n in {1,...,4}{
        \edef\temp{\noexpand\spy[anchor=center,magnification=3] on ([xshift=0.6em,yshift=-0.4em] qualitative-2-\n.center) in node[below left=3.75px and 3.75px] at (qualitative-2-\n.north east);}
        \temp
    }
    \foreach \n in {1,...,4}{
        \edef\temp{\noexpand\spy[anchor=center,magnification=3] on ([xshift=0.7em,yshift=0.0em] qualitative-3-\n.center) in node[below left=3.75px and 3.75px] at (qualitative-3-\n.north east);}
        \temp
    }

\end{tikzpicture}
}
}

\afterpage{%
\begin{figure*}[!htbp]
    \centering
    \begin{adjustbox}{center}
    \qualitativecomp
    \end{adjustbox}
    \caption{Qualitative comparison of rendered novel views and illuminations of the NeRF-OSR test scenes (from top to bottom: \emph{lk2}, \emph{lwp}, \emph{st}) synthesized by our method, LumiGauss \cite{kaleta2025lumigauss} (shadowed version), and NeRF-OSR \cite{rudnev2022nerf}. We show novel views employed for the evaluation, rendered using the ground truth environment maps. The sky is masked out in all the renderings.}
    \label{fig:qualitativecomp_renders}
\end{figure*}
}

\def\height{64px}
\def\spysize{42px}
\def\offset{36px}
\def\scale{0.7}
\def\scaleleft{0.7}
\def\distabove{0.3em}
\def\distleft{0.3em}

\newcommand{\qualitativecompalbedonormal}{
\centering
\resizebox{\linewidth}{!}{
    \begin{tikzpicture}[
        >=stealth',
        title/.style={
            anchor=base,
            align=center,
            scale=\scale,
        },
        spy using outlines={rectangle, red, magnification=3, connect spies, size=\spysize}
    ]
    \matrix[matrix of nodes, column sep=0pt, row sep=0pt, ampersand replacement=\&, inner sep=0, outer sep=0, font=\Large] (qualitative) {
        \includegraphics[width=.8\linewidth]{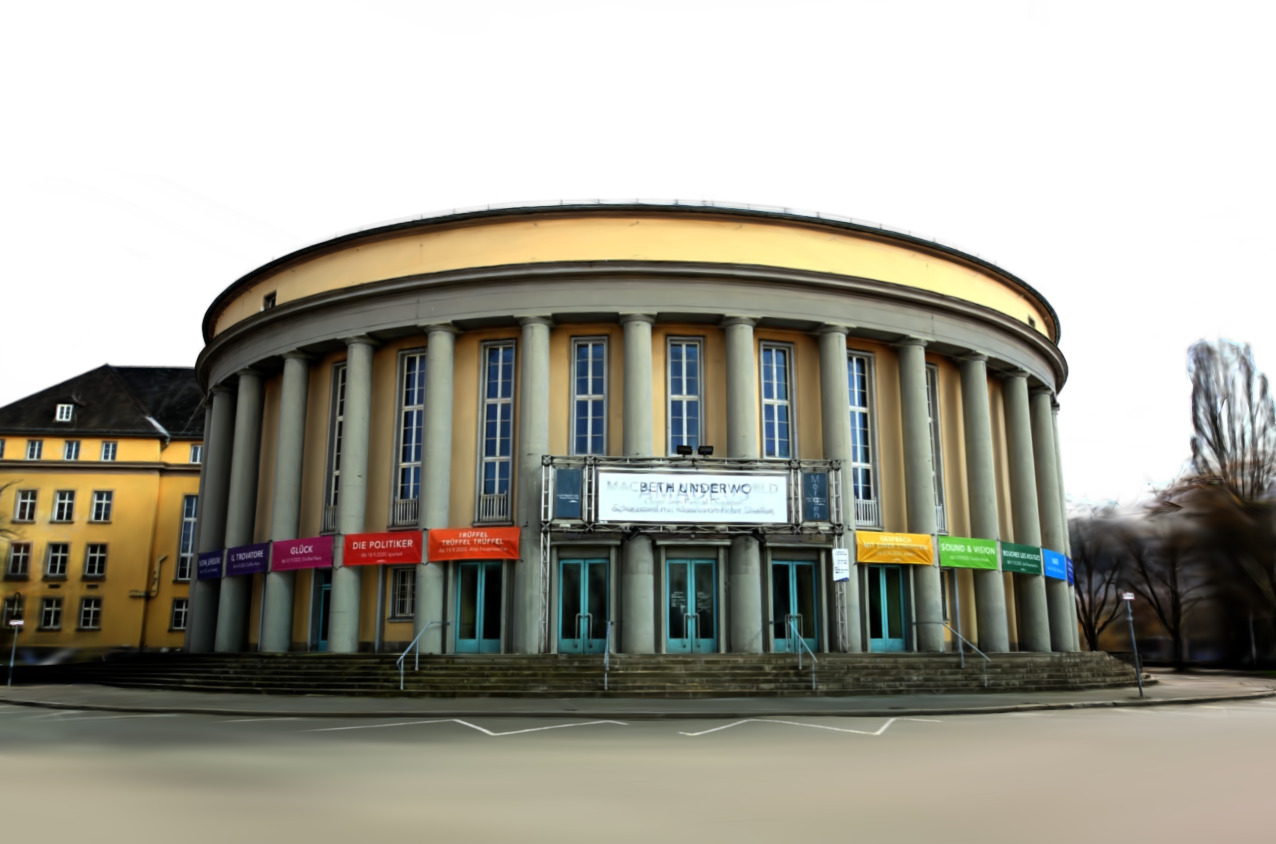} \&
        \includegraphics[width=.8\linewidth]{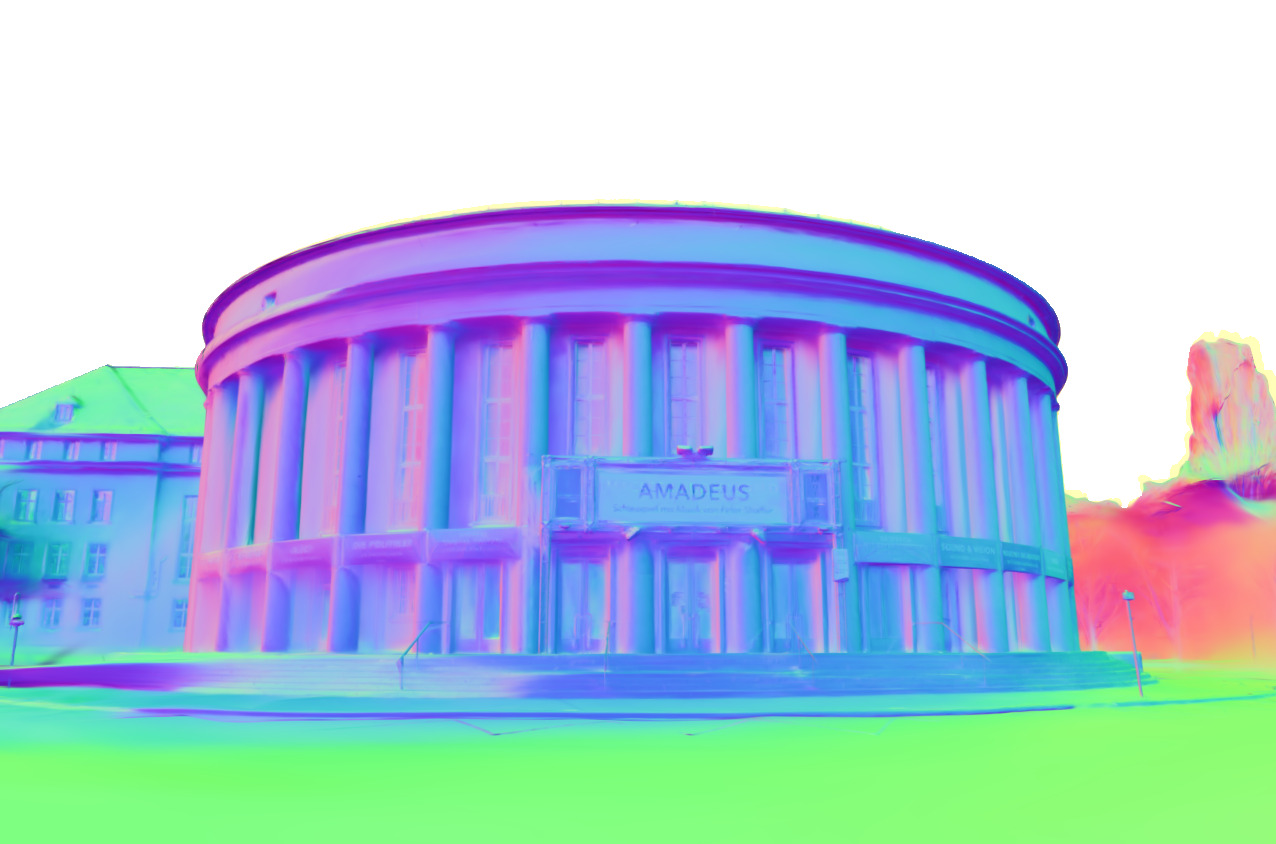} \\
        \includegraphics[width=.8\linewidth]{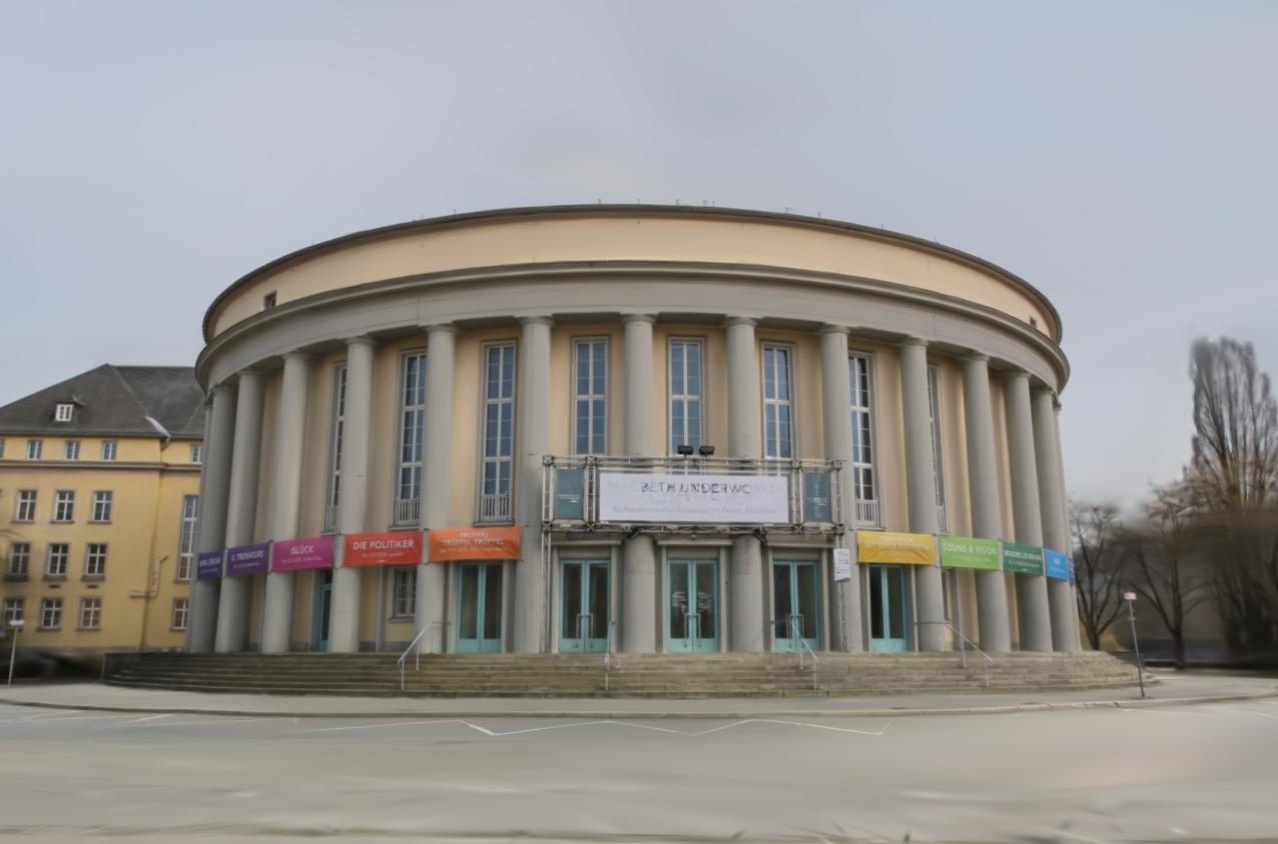} \&
        \includegraphics[width=.8\linewidth]{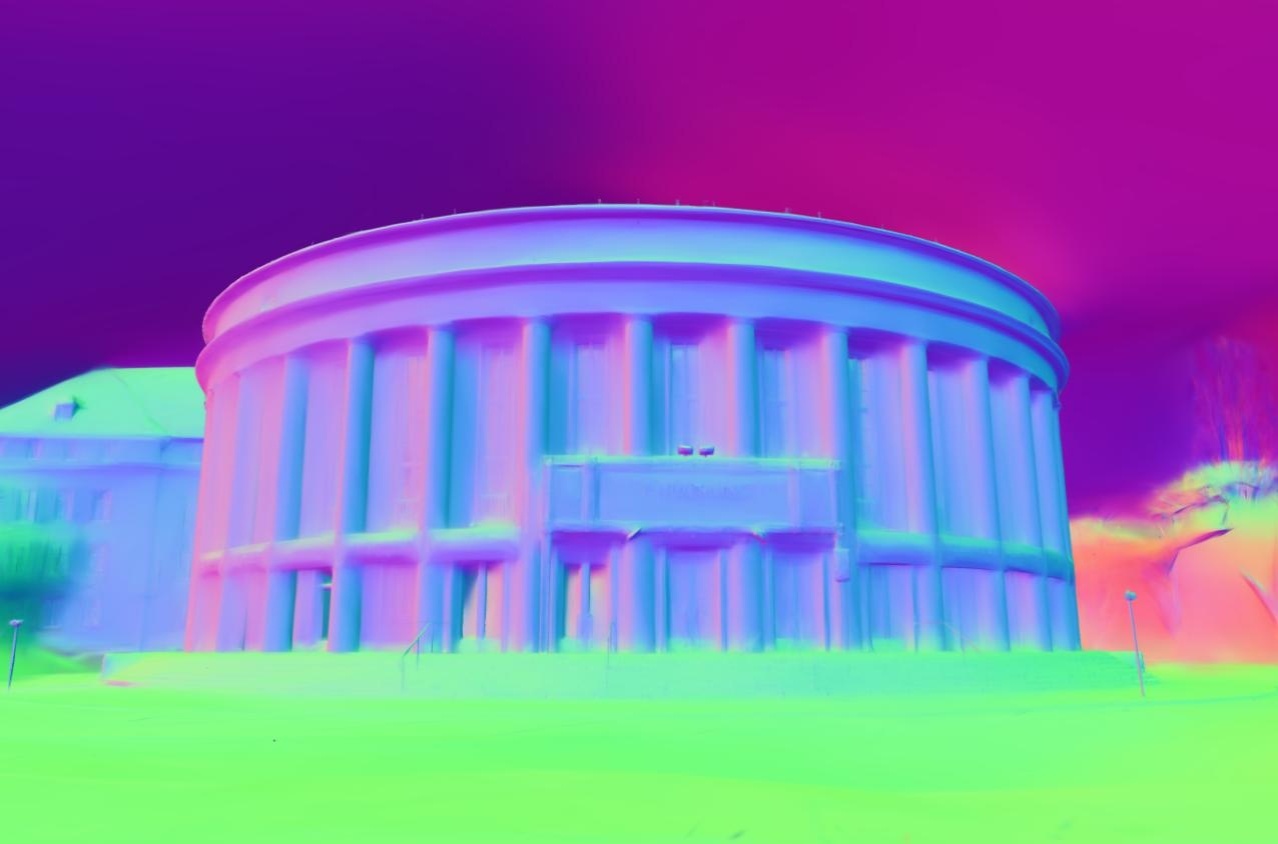} \\
        \includegraphics[width=.8\linewidth]{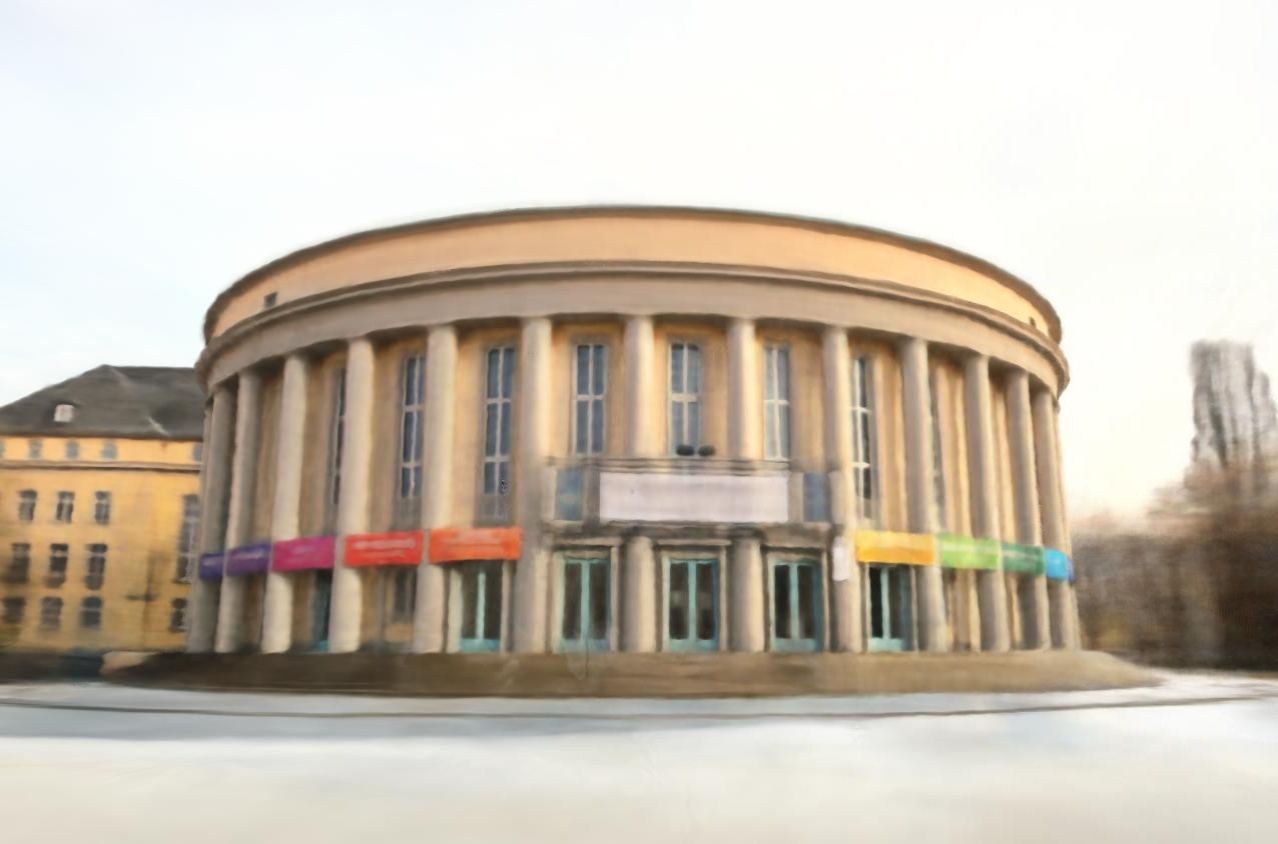} \&
        \includegraphics[width=.8\linewidth]{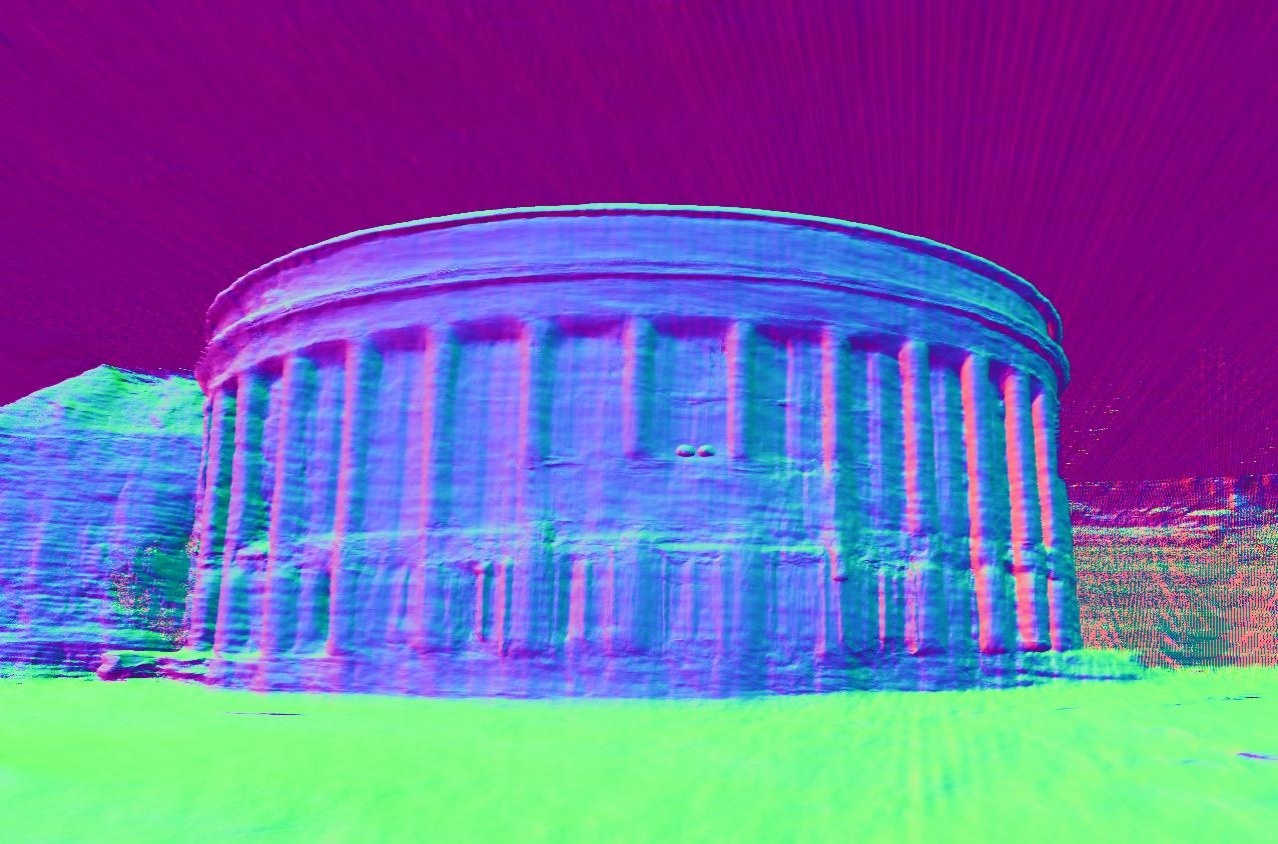}\\
    }; 

    \node[above=\distabove of qualitative-1-1.north, title, font=\Huge] {Albedo};
    \node[above=\distabove of qualitative-1-2.north, title, font=\Huge] {Surface Normals};

    \node[left=\distleft of qualitative-1-1.west, align=center,scale=\scaleleft, font=\Huge]{\rotatebox{90}{\textbf{Ours}}};
    \node[left=\distleft of qualitative-2-1.west, align=center,scale=\scaleleft, font=\Huge]{\rotatebox{90}{\textbf{LumiGauss}}};
    \node[left=\distleft of qualitative-3-1.west, align=center,scale=\scaleleft, font=\Huge]{\rotatebox{90}{\textbf{NeRF-OSR}}};

\end{tikzpicture}
}
}

\begin{figure}[!tb]
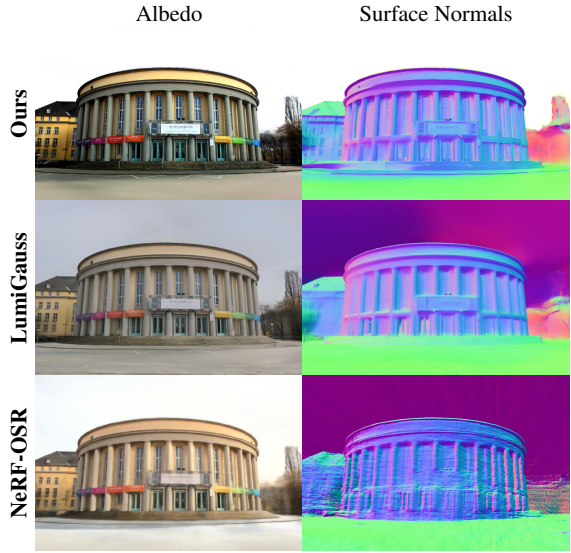

    \centering
    \begin{adjustbox}{center}
    \qualitativecompalbedonormal
    \end{adjustbox}
    \caption{Qualitative comparison of the albedo map and surface normals predicted by our method, NeRF-OSR \cite{rudnev2022nerf}, and LumiGauss \cite{kaleta2025lumigauss} for the NeRF-OSR scene \emph{st}. The ground truth novel view is visualized in Fig.~\ref{fig:qualitativecomp_renders}. R3GW synthesizes sharper surface normal maps than LumiGauss and NeRF-OSR. Furthermore, our sky Gaussians are not attributed with spurious albedo and surface normals.}
    \label{fig:qualitativecomp_albedo_normal_maps}
\end{figure}

\subsection{Dataset, Evaluation, and Baselines}
\label{sec:data}

To evaluate our approach, we strictly follow the evaluation protocol of NeRF-OSR \cite{rudnev2022nerf} and LumiGauss \cite{kaleta2025lumigauss}. We use the NeRF-OSR \cite{rudnev2022nerf} dataset, which contains eight outdoor scenes captured in the wild and provides ground truth LDR environment maps for each lighting condition in the test set of a scene. We use the official data split for Ludwigskirche (\emph{lk2}), Staatstheater (\emph{st}), and Landwehrplatz (\emph{lwp}). For each scene, we compute the average PSNR, MSE, MAE, and SSIM over five testing viewpoints. We select one view for each test lighting condition and render it using the corresponding ground truth environment map as illumination. Due to the implementation design for the image appearance embeddings, the MLP that predicts the SH coefficients of the sky color along with those of the environment light is unavailable for the test images. Therefore, when rendering the test views with the ground truth environment maps, we fix the color of the sky Gaussians to white. For the computation of the evaluation metrics, we mask out the occluders, sky, and vegetation using the segmentation masks provided in the NeRF-OSR dataset \cite{rudnev2022nerf,kaleta2025lumigauss}. We benchmark our approach against state-of-the-art methods described in Sec.~\ref{sec:related_work_rel_3drep_w}.

For each scene, we use the camera parameters and SfM point cloud provided by \cite{kaleta2025lumigauss}. We reproduced the results for LumiGauss \cite{kaleta2025lumigauss} and NeRF-OSR \cite{rudnev2022nerf}. For the other baselines, we report the results presented in their respective papers, which were obtained using the same evaluation protocol. Our approach and LumiGauss were trained for \num{40 000} iterations, while NeRF-OSR was trained for \num{500 000} iterations. We conducted all the experiments on an NVIDIA A100 GPU with 40 GB. See the Appendix for additional implementation details.
\subsection{Results}
\label{sec:results}

We present a comprehensive quantitative evaluation of our method against the baselines in Table~\ref{tab:eval_metrics}. The table is structured to highlight the trade-off between reconstruction quality and computational efficiency across the methods built on NeRF and Gaussian Splatting (GS). 

Compared to the NeRF-based approaches, R3GW attains competitive reconstruction quality: while it achieves higher average PSNR and SSIM scores than NeRF-OSR \cite{rudnev2022nerf} and SR-TensoRF \cite{chen2022tensorf}, it remains slightly below NeuSky \cite{gardner2024sky} in PSNR. Nonetheless, NeuSky requires about 14 hours of training per scene compared to only 2 hours for R3GW. This illustrates the efficiency advantage of our 3DGS-based formulation even when exact runtime comparisons are limited.
Notably, our method achieves the highest average SSIM across all methods that report this metric. Since R3GW does not explicitly model shadows, we additionally compare it to the unshadowed version of the GS baseline, LumiGauss \cite{kaleta2025lumigauss}. Our approach outperforms LumiGauss on the scene \emph{lk2} across all evaluation metrics, for both the shadowed and unshadowed versions of the method. On the \emph{lwp} and \emph{st} scenes, R3GW achieves performance on par with or better than that of the unshadowed LumiGauss, with higher SSIM and lower MAE scores. Averaged across all scenes, our method outperforms LumiGauss unshadowed.  We attribute the remaining difference with LumiGauss shadowed to the absence of explicit shadow modeling in our current formulation, highlighting a promising direction for future extensions.

Fig.~\ref{fig:qualitativecomp_renders} demonstrates that R3GW reconstructs novel views of the scenes with high fidelity, including architectural details, while reproducing realistic novel illuminations. Notably, all methods render colors that deviate from the ground truth, as reflected in the relatively low PSNR scores. We attribute this to the fact that the models do not predict the illumination for the test views, but they are directly provided with the ground truth environment map, which is manually aligned with the scene reconstruction. Fig.~\ref{fig:qualitativecomp_albedo_normal_maps} further demonstrates that our method successfully decomposes the material and geometry of the scene, predicting plausible albedo maps and sharper surface normals than the baselines. Unlike the baselines, R3GW computes the diffuse color of the foreground Gaussians with albedo in linear space and gamma-corrects the sum of the diffuse and specular colors. This results in different tone and contrast in albedo maps, which are predicted by blending the albedo attributes of the foreground
Gaussians and clamped to $[0,1]$ for visualization. We present an additional visual comparison of our method with LumiGauss \cite{kaleta2025lumigauss} and NeRF-OSR \cite{rudnev2022nerf} in the Appendix of this paper.

Fig.~\ref{fig:qualitativecomp_depth_maps} compares the depth maps predicted by our method and LumiGauss \cite{kaleta2025lumigauss} to qualitatively assess how our decoupled sky representation affects depth estimation. Unlike LumiGauss, our approach prevents sky Gaussians from being placed close to the camera position, which reduces depth artifacts. The insets in Fig.~\ref{fig:qualitativecomp_skyrend} show that this results in sharper and halo-free rendering of the foreground outlines.
\def\height{64px}
\def\offset{36px}
\def\scale{0.7}
\def\scaleleft{0.7}
\def\distabove{0.3em}

\newcommand{\qualitativerelightingext}{
\centering
\resizebox{\linewidth}{!}{
    \begin{tikzpicture}[
        >=stealth',
        title/.style={
            anchor=base,
            align=center,
            scale=\scale,
        },
    ]
    \matrix[matrix of nodes, column sep=-0.5pt, row sep=0pt, ampersand replacement=\&, inner sep=0, outer sep=0, font=\Large] (qualitative) {
        \begin{tikzpicture}[baseline=(main.base)]
            \node[inner sep=0pt, outer sep=0] (main) at (0,0) {\includegraphics[width=0.8\textwidth]{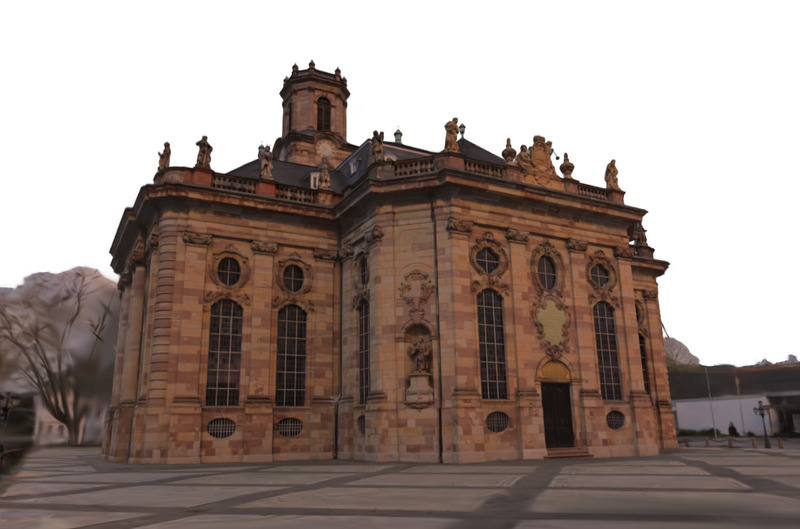}};
            \node[anchor=north west] (overlay1) at (main.north west) {\includegraphics[width=0.3\textwidth]{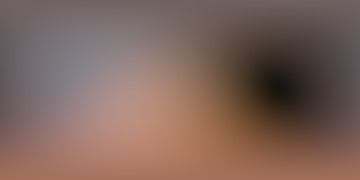}};
        \end{tikzpicture}\&
        \includegraphics[width=0.8\textwidth]{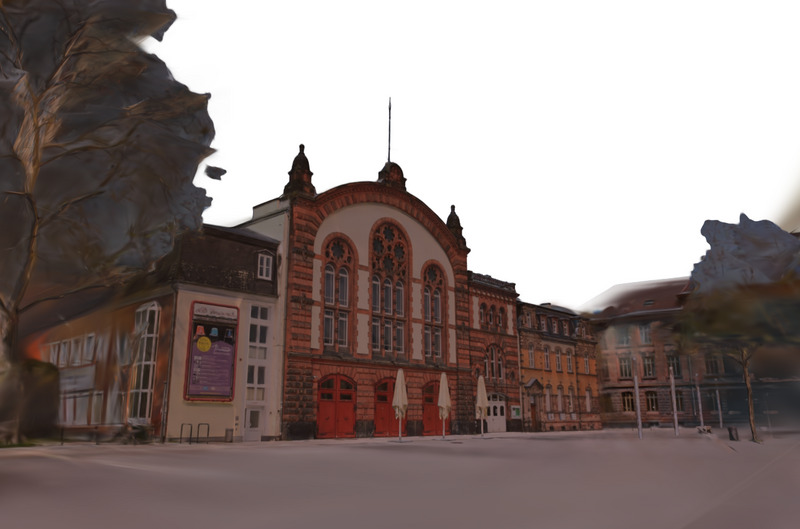}\&
        \includegraphics[width=0.8\textwidth]{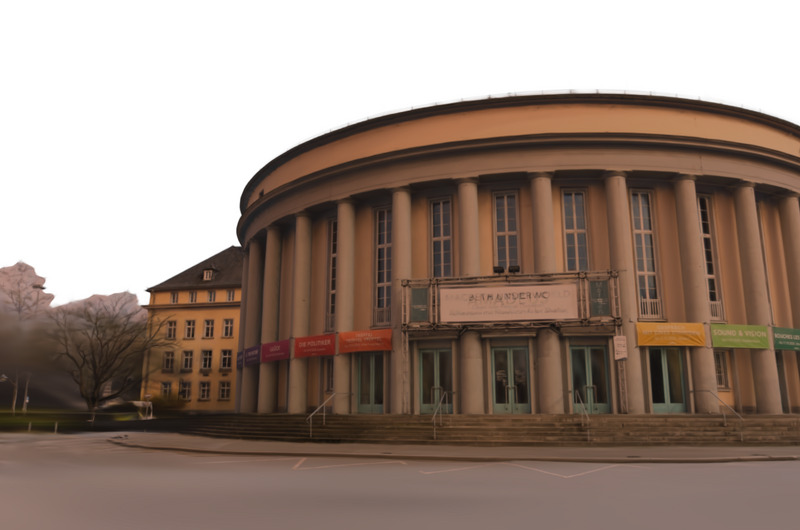}\\
        \begin{tikzpicture}[baseline=(main.base)]
            \node[inner sep=0pt, outer sep=0] (main) at (0,0) {\includegraphics[width=0.8\textwidth]{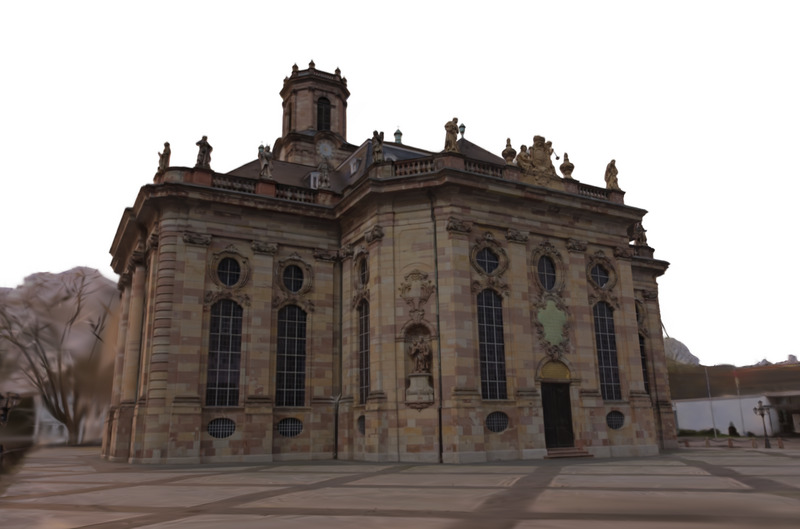}};
            \node[anchor=north west] (overlay1) at (main.north west) {\includegraphics[width=0.3\textwidth]{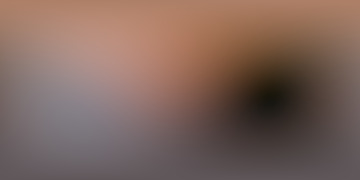}};
        \end{tikzpicture}\&
        \includegraphics[width=0.8\textwidth]{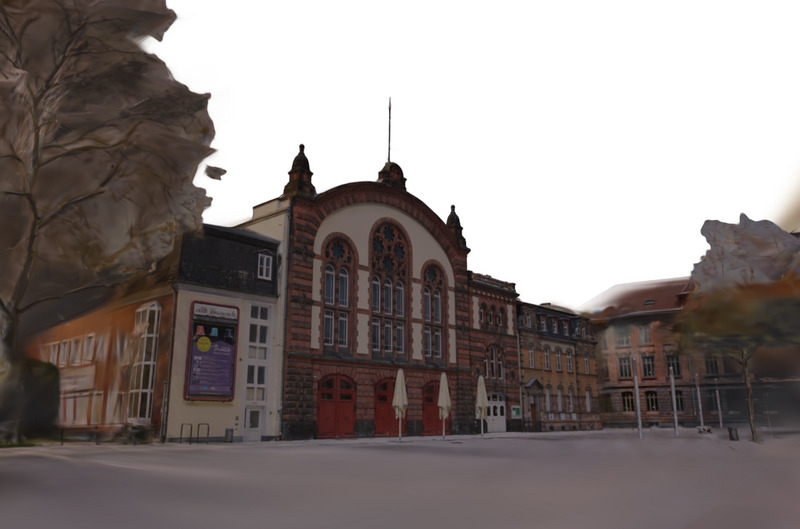}\&
        \includegraphics[width=0.8\textwidth]{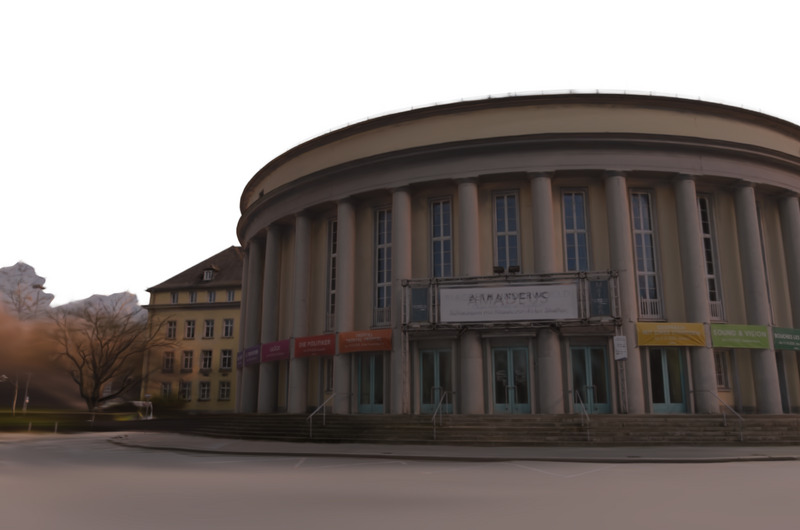}\\
    }; 

\end{tikzpicture}
}
}

\begin{figure}[!htb]
        \centering
        \begin{adjustbox}{valign=c,center}
        \qualitativerelightingext 
        \end{adjustbox}
        \caption{Novel views of the NeRF-OSR scenes relighted under a novel HDR environment map. The first and second row show renderings with two different sun orientations, produced by projecting onto fourth-order SH the same environment map and rotating it around the y-axis (top-left corner of each row).
        }
        \label{fig:relit_external_envmaps}
\end{figure}
\subsection{Ablation Studies}
\label{sec:ablation}

We evaluate the contribution of the specular component $\mathbf{c}_{\text{specular}}$ (Eq.~(\ref{eq:spec_color})) in the foreground Gaussians' color attribute through an ablation study. To isolate its effect, we trained the model assuming purely diffuse reflections by disabling $\mathbf{c}_{\text{specular}}$ and reducing the degree of the environmental light SH to $2$. Table~\ref{tab:eval_metrics} compares our model to its ablated version. For all the scenes, the view-dependent specular color enhances our model's ability to synthesize novel views and illuminations. Importantly, removing the specular view-dependent component from the color of the foreground Gaussians results in a drop in both PSNR and SSIM.

We further ablate the loss term $\mathcal{L}_{\text{sky-depth}}$ (Eq.~(\ref{eq:loss_depth})). The comparison with the full R3GW model in Table~\ref{tab:eval_metrics} indicates that $\mathcal{L}_{\text{sky-depth}}$ improves both PSNR and SSIM scores.
\def\height{64px}
\def\offset{36px}
\def\scale{0.7}
\def\scaleleft{0.7}
\def\distabove{0.5em}

\newcommand{\qualitativerelightingexptrain}{
\centering
\resizebox{\linewidth}{!}{
    \begin{tikzpicture}[
        >=stealth',
        title/.style={
            anchor=base,
            align=center,
            scale=1.5,
        },
    ]
    \matrix[matrix of nodes, column sep=-0.5pt, row sep=0pt, ampersand replacement=\&, inner sep=0, outer sep=0, font=\Large] (qualitative) {
        \includegraphics[width=0.8\textwidth]{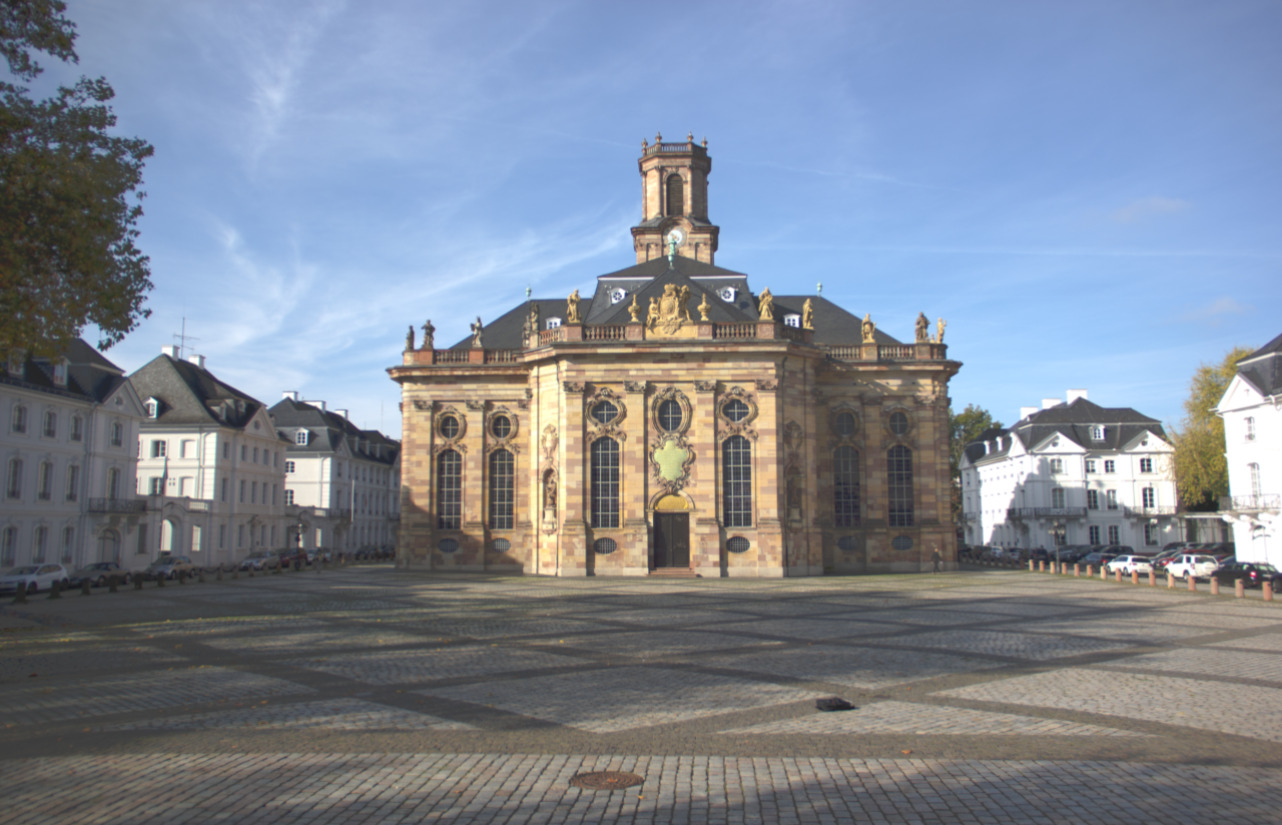}\&
        \begin{tikzpicture}[baseline=(main.base)]
            \node[inner sep=0pt, outer sep=0] (main) at (0,0) {\includegraphics[width=0.8\textwidth]{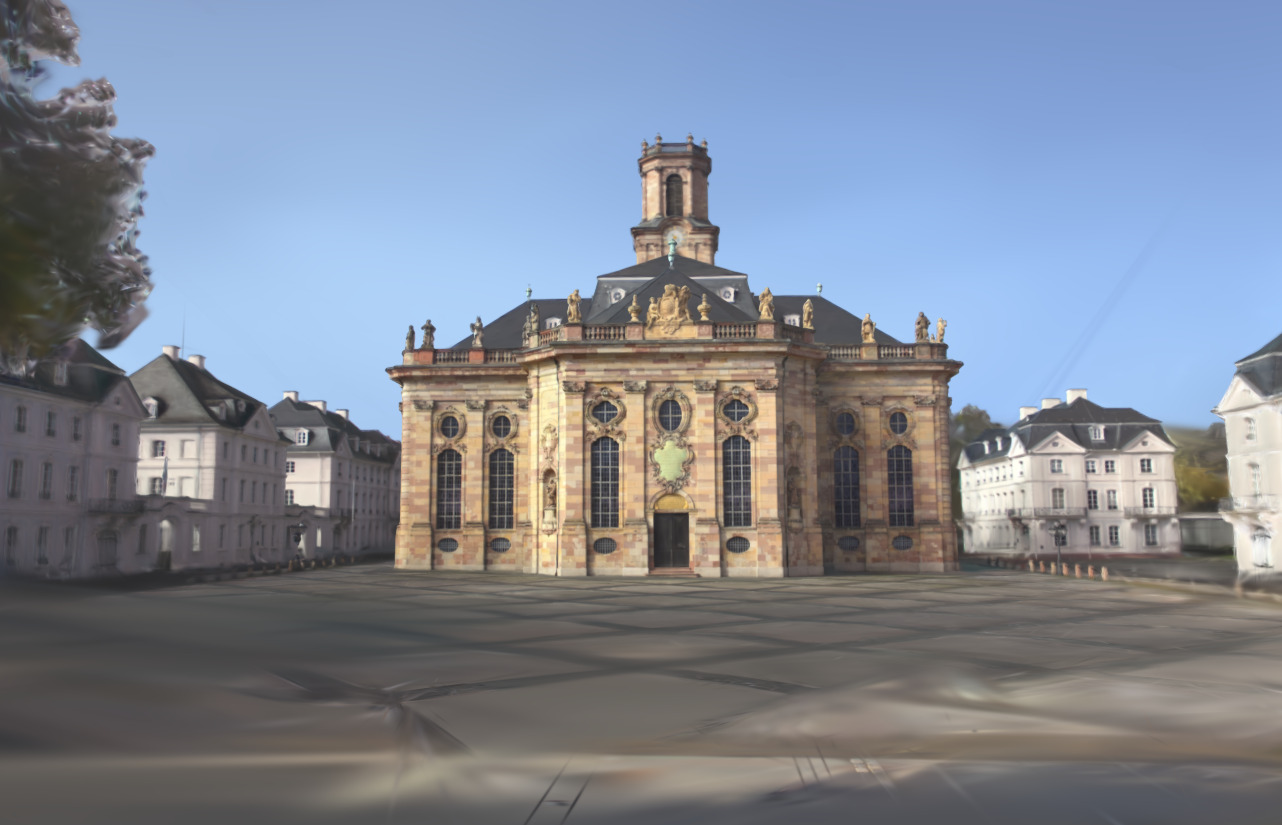}};
            \node[anchor=north west] (overlay1) at (main.north west) {\includegraphics[width=0.3\textwidth]{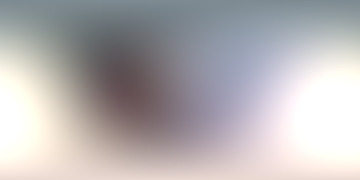}};
        \end{tikzpicture}\&
        \includegraphics[width=0.8\textwidth]{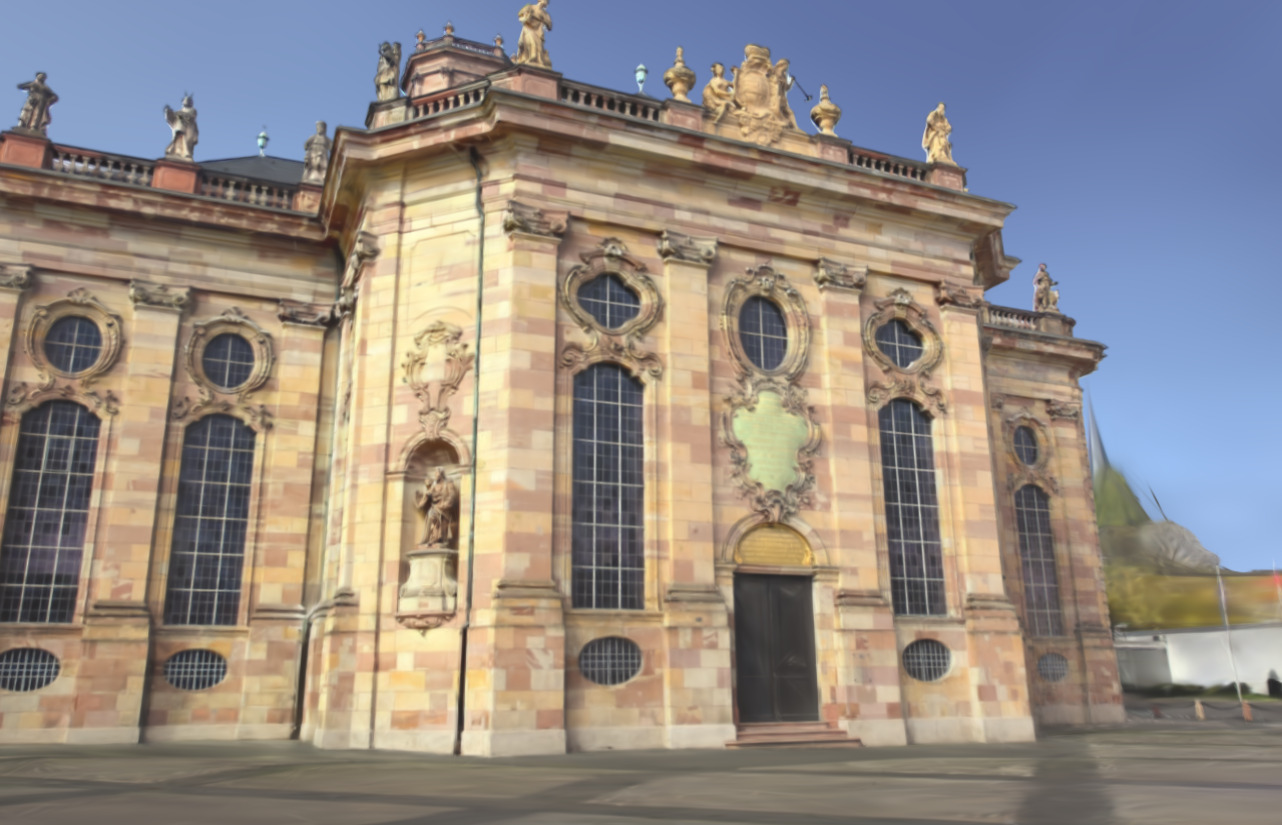}\\
        \includegraphics[width=0.8\textwidth]{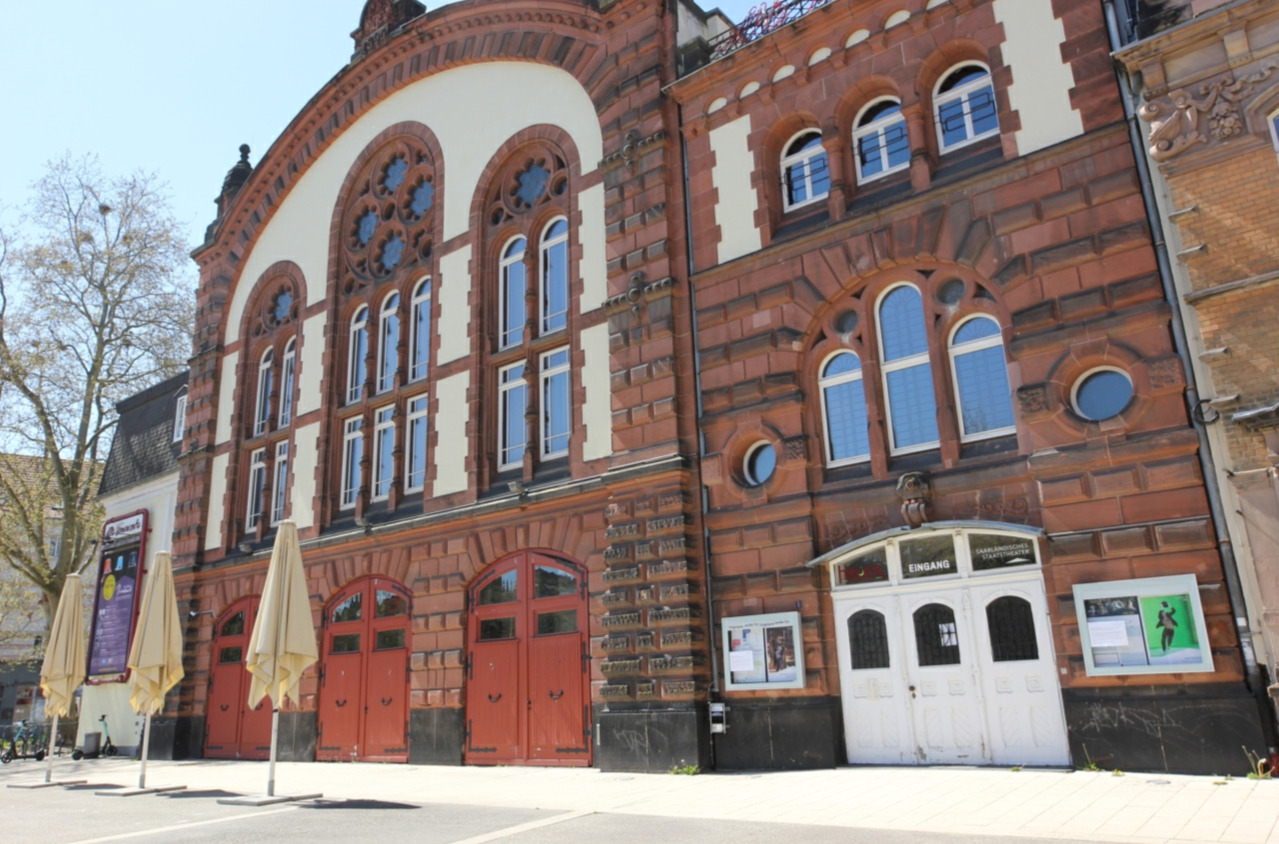}\&
        \begin{tikzpicture}[baseline=(main.base)]
            \node[inner sep=0pt, outer sep=0] (main) at (0,0) {\includegraphics[width=0.8\textwidth]{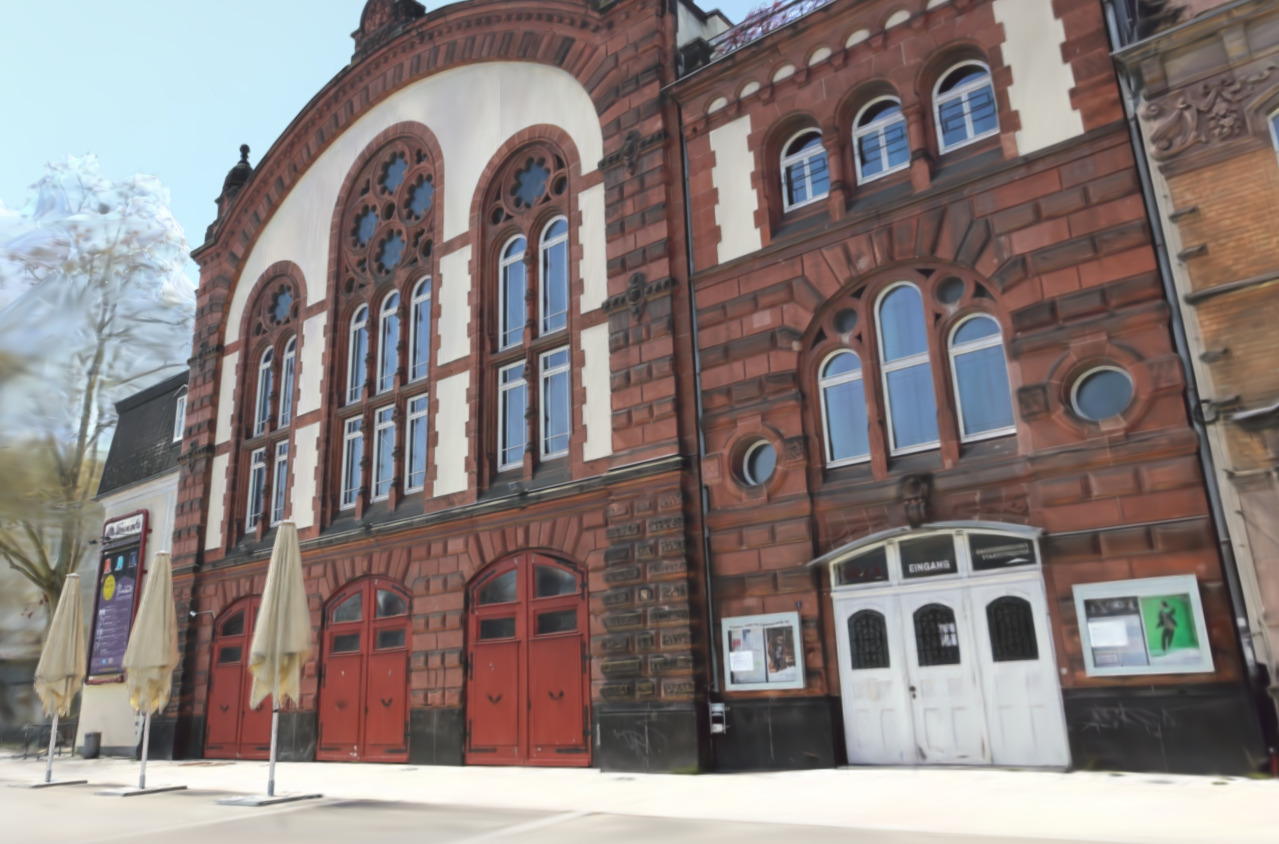}};
            \node[anchor=north west] (overlay1) at (main.north west) {\includegraphics[width=0.3\textwidth]{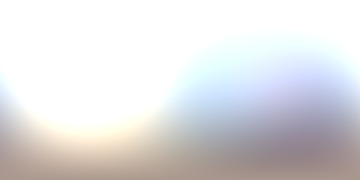}};
        \end{tikzpicture}\&
        \includegraphics[width=0.8\textwidth]{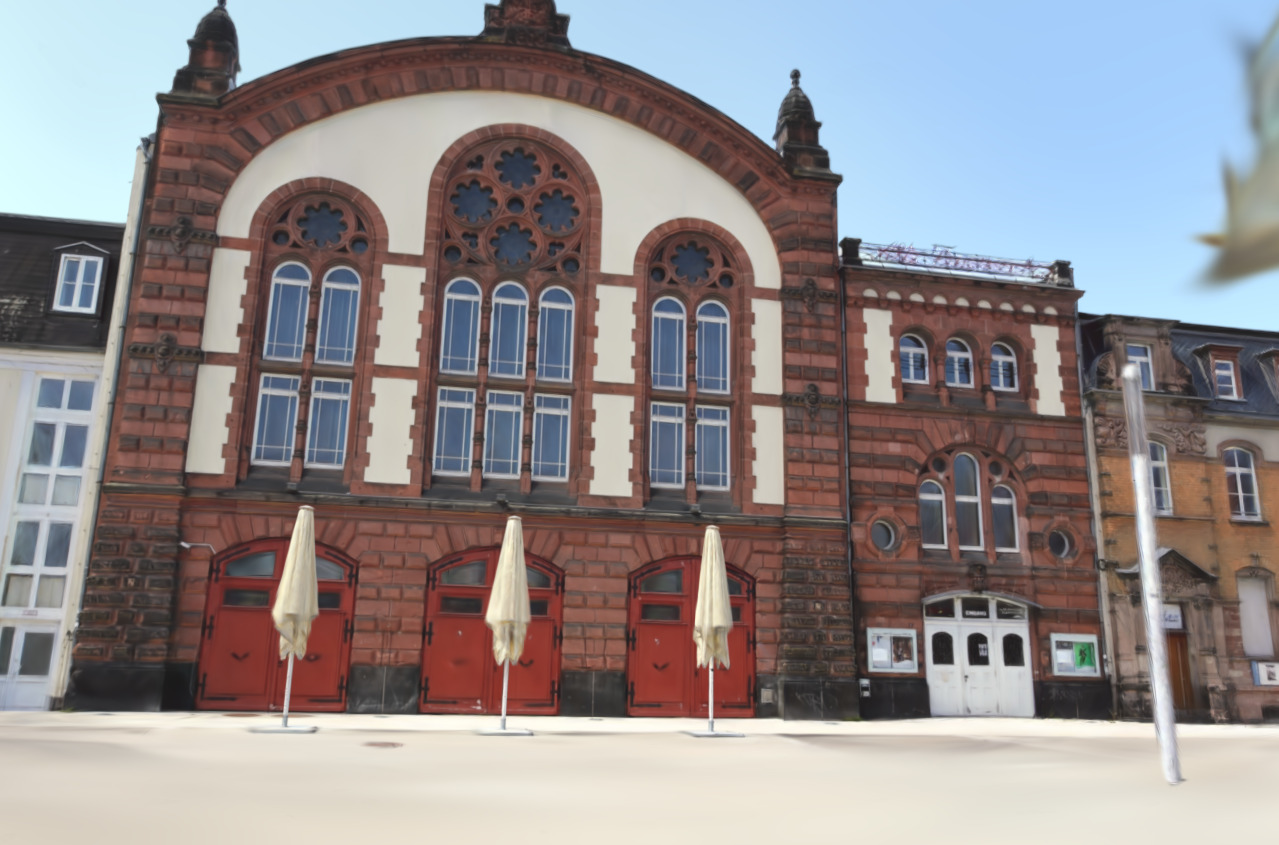}\\
        \includegraphics[width=0.8\textwidth]{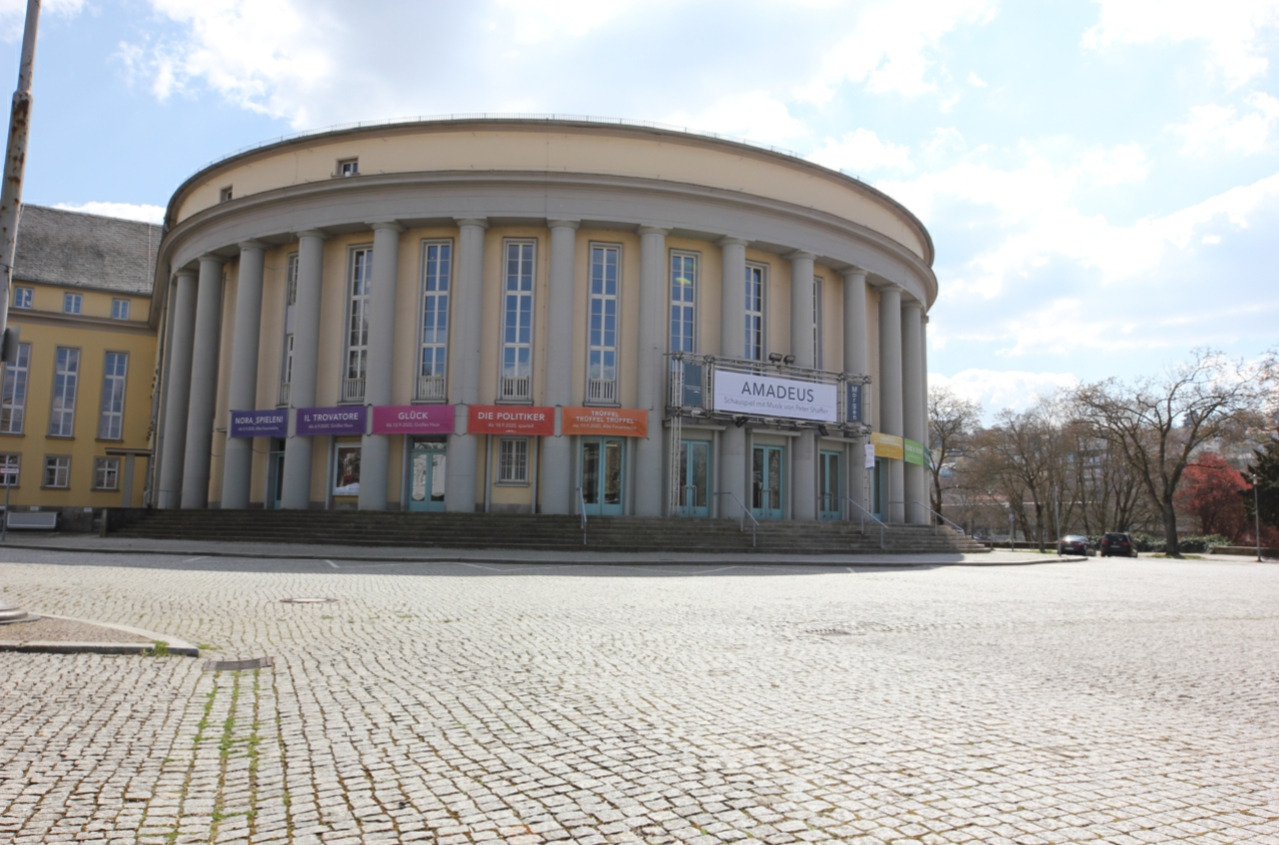}\&
        \begin{tikzpicture}[baseline=(main.base)]
            \node[inner sep=0pt, outer sep=0] (main) at (0,0) {\includegraphics[width=0.8\textwidth]{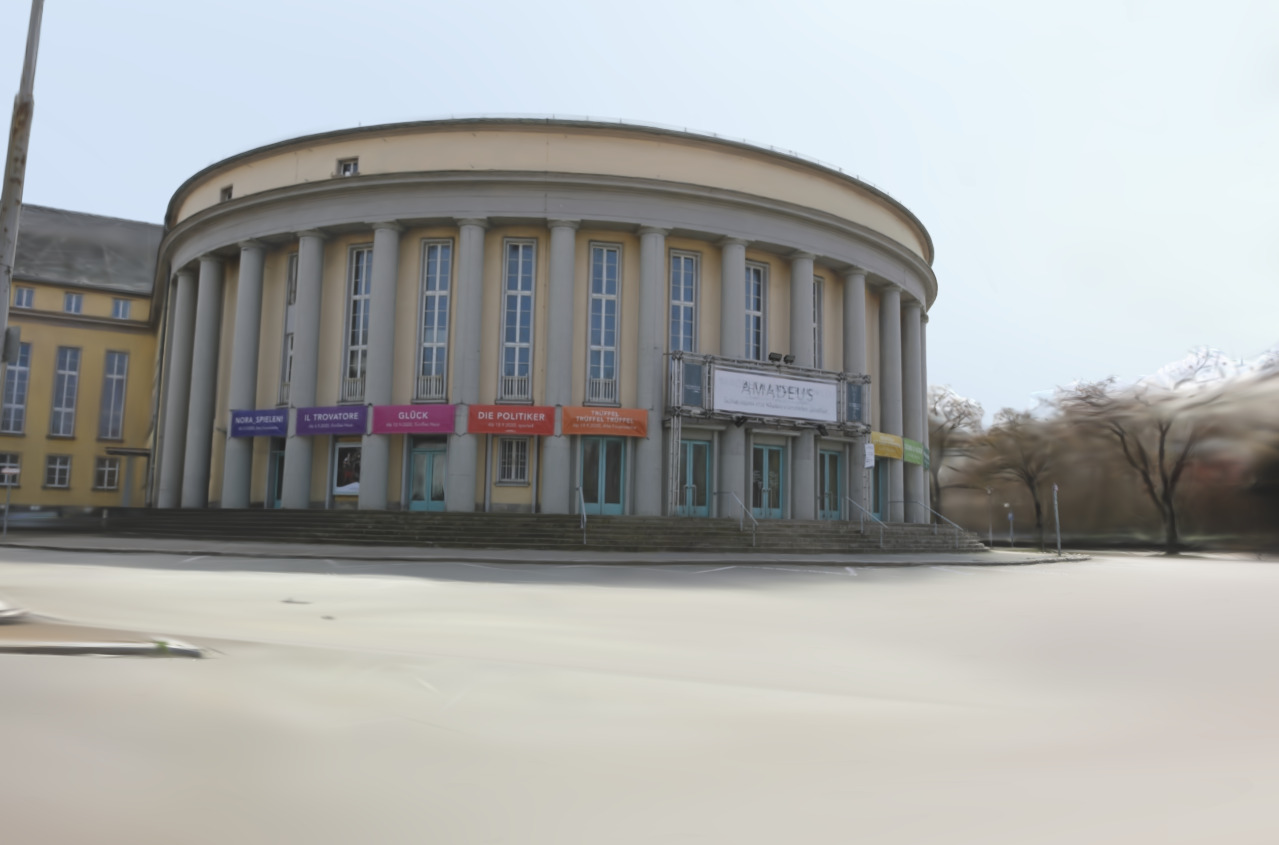}};
            \node[anchor=north west] (overlay1) at (main.north west) {\includegraphics[width=0.3\textwidth]{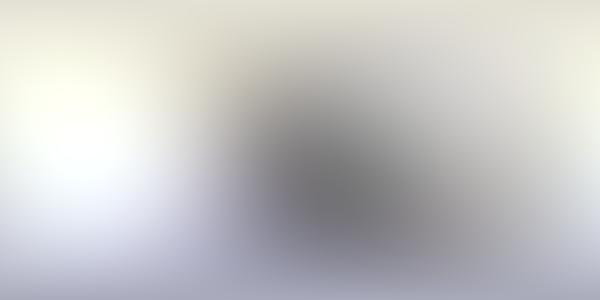}};
        \end{tikzpicture}\&
        \includegraphics[width=0.8\textwidth]{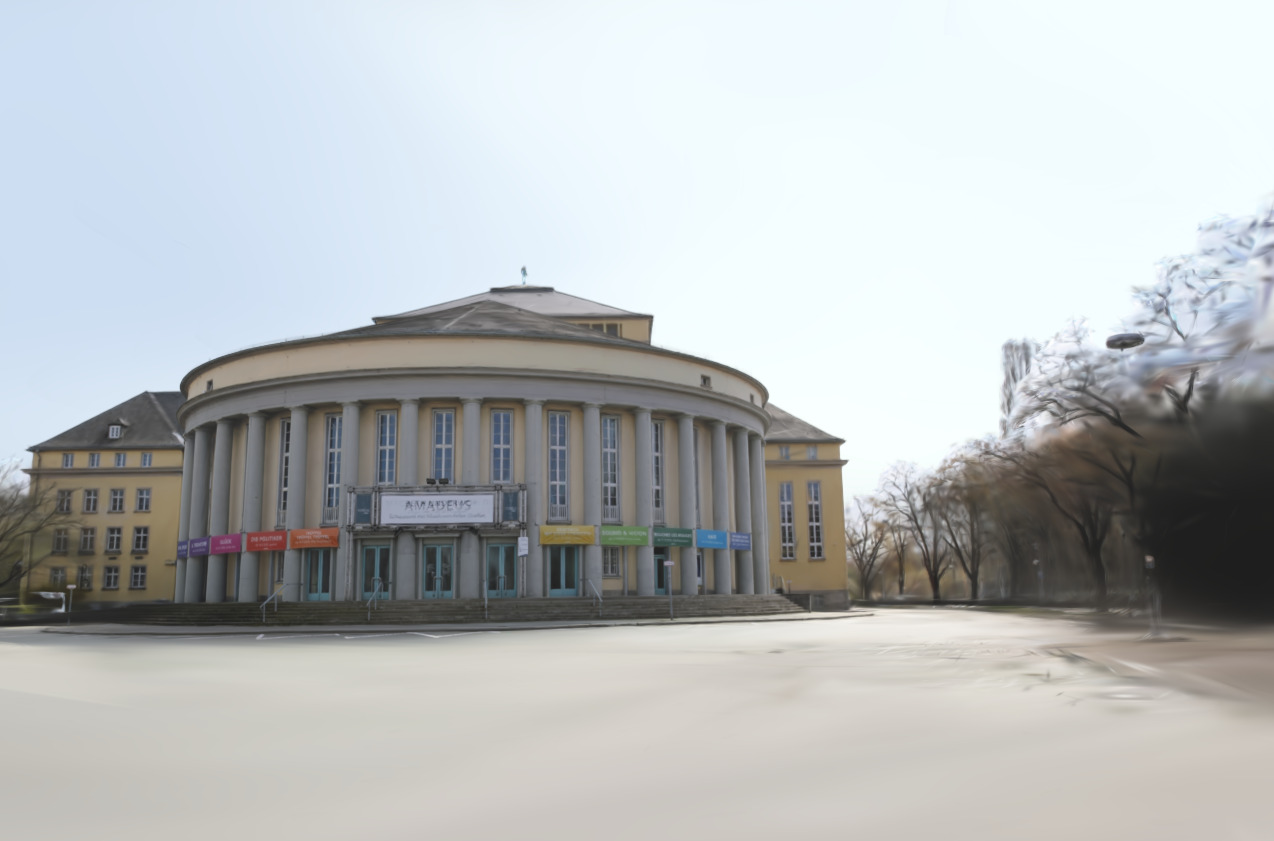}\\
    }; 

    \node[above=\distabove+0.2em of qualitative-1-1.north, title, font=\Huge] {Training Ground Truth};
    \node[above=\distabove+0.2em of qualitative-1-2.north, title, font=\Huge] {Reconstructed View and Light};
    \node[above=\distabove+0.2em of qualitative-1-3.north, title, font=\Huge] {Relighted Novel View};

\end{tikzpicture}
}
}

\begin{figure}[!t]
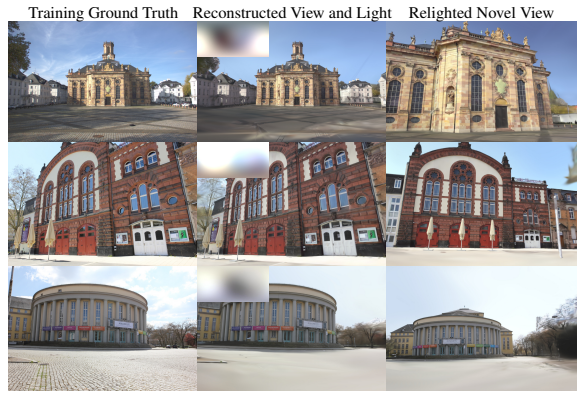

        \centering
        \begin{adjustbox}{valign=c,center}
        \qualitativerelightingexptrain 
        \end{adjustbox}
        \caption{Novel views of the NeRF-OSR scenes relighted with trained environment maps (top-left corner of the second column).
        }
        \label{fig:relit_trained_envmaps}
\end{figure}
\subsection{Relighting}
\label{sec:relighting}

To demonstrate our model's relighting capabilities, we render novel views of the NeRF-OSR scenes using a novel HDR environment map as illumination, see Fig.~\ref{fig:relit_external_envmaps}. During training, our method predicts an environment map for each image. We use these learned environment maps to relight novel views of the scenes. At training time, R3GW also predicts the color of the sky per image. Thus, to illustrate one way of employing the sky Gaussians at inference, we render them using the trained sky color. Fig.~\ref{fig:relit_trained_envmaps} shows that the learned illuminations remain consistent when changing the scene viewpoint.
\section{\uppercase{Limitations and Future Work}}
\label{sec:limitation}

Our method does not explicitly model cast shadows or indirect lighting effects. Thus, incorporating a dedicated shadow modeling component presents a promising opportunity to further enhance realism and to disentangle surface material appearance and incident illumination in the presence of prominent shadows. A natural extension of our work would be to augment the environment light with a sun directional light and adapt shadow mapping techniques to the 3DGS setting. Introducing a directional light could also address our method’s current limitation in reproducing sharp specular highlights, which can occur under clear sky or sunny conditions and are challenging to capture with SH-based illumination only. Furthermore, since the sky is part of the distant lighting illuminating the scene, we believe that relating its appearance to the environment light could further enhance the relighting capabilities of our approach.
\section{\uppercase{Conclusions}}
\label{sec:conclusions}

We have presented R3GW, a new relighting pipeline for Gaussian Splatting that is capable of handling the complexities of outdoor scenes acquired in unconstrained environments, bridging the gap between real-world capture and controllable rendering. The foreground Gaussians enhanced with a PBR color enable relighting and view-dependent appearance modeling of the scene foreground, while the sky Gaussians, accounting for the reconstruction of the sky, preserve the physical plausibility of the learned scene representation and boost the accuracy of the recovered geometry. The experimental results have shown that our approach performs comparably to the baselines and accurately reproduces novel illuminations along with novel views. 
\section*{\uppercase{Acknowledgments}}
This work has partly been funded by the European Commission (Horizon Europe projects Luminous, grant no.\ 101135724 and Spirit, grant no.\ 101070672).

\bibliographystyle{apalike}
{\small
\bibliography{R3GWbib}}

@article{kerbl3Dgaussians,
  title={{3D Gaussian Splatting for Real-Time Radiance Field Rendering}},
  author={Kerbl, Bernhard and Kopanas, Georgios and Leimk{\"u}hler, Thomas and Drettakis, George},
  journal={ACM Trans. on Graphics},
  volume={42},
  number={4},
pages={1--14},
  year={2023}
}

@inproceedings{R3DG2023,
    author    = {Gao, Jian and Gu, Chun and Lin, Youtian and Zhu, Hao and Cao, Xun and Zhang, Li and Yao, Yao},
    title     = {{Relightable 3D Gaussian: Real-Time Point Cloud Relighting with BRDF Decomposition and Ray Tracing}},
    booktitle = {Proc. of European Conference on Computer Vision},
    pages = {73--89},
    year  = {2024}
}

@inproceedings{liang2023gs,
  title={{GS-IR: 3D Gaussian Splatting for Inverse Rendering}},
  author={Liang, Zhihao and Zhang, Qi and Feng, Ying and Shan, Ying and Jia, Kui},
  booktitle={Proc. of the IEEE/CVF Conference on Computer Vision and Pattern Recognition},
  pages={21644--21653},
  year={2024}
}

@inproceedings{jiang2023gaussianshader,
  title={{GaussianShader: 3D Gaussian Splatting with Shading Functions for Reflective Surfaces}},
  author={Jiang, Yingwenqi and Tu, Jiadong and Liu, Yuan and Gao, Xifeng and Long, Xiaoxiao and Wang, Wenping and Ma, Yuexin},
  booktitle={Proc. of the IEEE/CVF Conference on Computer Vision and Pattern Recognition},
  pages={5322--5332},
  year={2024}
}

@inproceedings{renderingeq,
  title={{The Rendering Equation}},
  author={Kajiya, James T},
  booktitle={Proc. 13th annual Conference on Computer Graphics and Interactive Techniques},
  pages={143--150},
  year={1986},
  doi = {10.1145/15886.15902},
  url = {https://doi.org/10.1145/15886.15902}
}

@inproceedings{IrradianceEnvMaps,
  title={{An Efficient Representation for Irradiance Environment Maps}},
  author={Ramamoorthi, Ravi and Hanrahan, Pat},
  booktitle={Proc. 28th Annual Conference on Computer Graphics and Interactive Techniques},
  pages={497--500},
  year={2001},
  doi = {10.1145/383259.383317},
  url = {https://doi.org/10.1145/383259.383317}
}

@inproceedings{kaleta2025lumigauss,
                title     = {{LumiGauss: Relightable Gaussian Splatting in the Wild}},
                author    = {
                    Kaleta, Joanna 
                    and Kania, Kacper 
                    and Trzci{\'n}ski, Tomasz
                    and Kowalski, Marek
                },
                booktitle = {Proc. of IEEE/CVF Winter Conference on Applications of Computer Vision},
                pages={1--10},
                year      = {2025}
            }

@inproceedings{SplitSumApprox,
  author    = {Karis, Brian},
  title     = {{Real Shading in Unreal Engine 4}},
  booktitle = {ACM SIGGRAPH 2013 Courses: Physically Based Shading in Theory and Practice},
  year      = {2013},
  url       = {https://blog.selfshadow.com/publications/s2013-shading-course/karis/s2013_pbs_epic_notes_v2.pdf}
}

@article{chen2023neusg,
  title={{NeuSG: Neural Implicit Surface Reconstruction with 3D Gaussian Splatting Guidance}},
  author={Chen, Hanlin and Li, Chen and Lee, Gim Hee},
  journal={arXiv preprint arXiv:2312.00846},
  year={2023}
}

@article{mildenhall2021nerf,
  title={{NeRF: Representing Scenes as Neural Radiance Fields for View Synthesis}},
  author={Mildenhall, Ben and Srinivasan, Pratul P and Tancik, Matthew and Barron, Jonathan T and Ramamoorthi, Ravi and Ng, Ren},
  journal={Communications of the ACM},
  volume={65},
  number={1},
  pages={99--106},
  year={2020},
  publisher={ACM New York, NY, USA}
}

@inproceedings{rudnev2022nerf,
  title={{NeRF for Outdoor Scene Relighting}},
  author={Rudnev, Viktor and Elgharib, Mohamed and Smith, William and Liu, Lingjie and Golyanik, Vladislav and Theobalt, Christian},
  booktitle={Proc. of European Conference on Computer Vision},
  pages={615--631},
  year={2022}
}

@inproceedings{chang2024fast,
  title={{Fast Sun-aligned Outdoor Scene Relighting based on TensoRF}},
  author={Chang, Yeonjin and Kim, Yearim and Seo, Seunghyeon and Yi, Jung and Kwak, Nojun},
  booktitle={Proc. of IEEE/CVF Winter Conference on Applications of Computer Vision},
  pages={3626--3636},
  year={2024}
}

@inproceedings{chen2022tensorf,
  title={{TensoRF: Tensorial Radiance Fields}},
  author={Chen, Anpei and Xu, Zexiang and Geiger, Andreas and Yu, Jingyi and Su, Hao},
  booktitle={Proc. of European Conference on Computer Vision},
  pages={333--350},
  year={2022}
}

@inproceedings{sun2023sol,
  title={{SOL-NeRF: Sunlight Modeling for Outdoor Scene Decomposition and Relighting}},
  author={Sun, Jia-Mu and Wu, Tong and Yang, Yong-Liang and Lai, Yu-Kun and Gao, Lin},
  booktitle={SIGGRAPH Asia 2023 Conference Papers},
  pages={1--11},
  year={2023}
}

@techreport{chung2005gauss,
  title={{Gauss-Weierstrass Kernel Smoothing on Unit Sphere}},
  author={Chung, Moo K},
  institution={University of Wisconsin-Madison},
  year={2005}
}

@inproceedings{gardner2024sky,
  title={{The Sky’s the Limit: Relightable Outdoor Scenes via a Sky-Pixel Constrained Illumination Prior and Outside-In Visibility}},
  author={Gardner, James AD and Kashin, Evgenii and Egger, Bernhard and Smith, William AP},
  booktitle={Proc. of European Conference on Computer Vision},
  pages={126--143},
  year={2024}
}

@article{kulhanek2024wildgaussians,
  title={{Wildgaussians: 3D Gaussian Splatting in the Wild}},
  author={Kulhanek, Jonas and Peng, Songyou and Kukelova, Zuzana and Pollefeys, Marc and Sattler, Torsten},
  journal={arXiv preprint arXiv:2407.08447},
  year={2024}
}

@inproceedings{zhang2024gaussian,
  title={{Gaussian in the Wild: 3D Gaussian Splatting for Unconstrained Image Collections}},
  author={Zhang, Dongbin and Wang, Chuming and Wang, Weitao and Li, Peihao and Qin, Minghan and Wang, Haoqian},
  booktitle={Proc. of European Conference on Computer Vision},
  pages={341--359},
  year={2024}
}

@article{xu2024splatfacto,
  title={ {Splatfacto-W: A Nerfstudio Implementation of Gaussian Splatting for Unconstrained Photo Collections}},
  author={Xu, Congrong and Kerr, Justin and Kanazawa, Angjoo},
  journal={arXiv preprint arXiv:2407.12306},
  year={2024}
}

@article{cook1982reflectance,
  title={{A Reflectance Model for Computer Graphics}},
  author={Cook, Robert L and Torrance, Kenneth E.},
  journal={ACM Trans. on Graphics},
  volume={1},
  number={1},
  pages={7--24},
  year={1982}
}

@inproceedings{schonberger2016structure,
  title={{Structure-from-Motion Revisited}},
  author={Schonberger, Johannes L and Frahm, Jan-Michael},
  booktitle={Proc. of the IEEE/CVF Conference on Computer Vision and Pattern Recognition},
  pages={4104--4113},
  year={2016}
}

@article{philip2019multi,
  title={{Multi-view Relighting Using a Geometry-Aware Network}},
  author={Philip, Julien and Gharbi, Micha{\"e}l and Zhou, Tinghui and Efros, Alexei A and Drettakis, George},
  journal={ACM Trans. on Graphics},
  volume={38},
  number={4},
  pages={78--1},
  year={2019}
}

@inproceedings{yu2020self,
  title={{Self-supervised Outdoor Scene Relighting}},
  author={Yu, Ye and Meka, Abhimitra and Elgharib, Mohamed and Seidel, Hans-Peter and Theobalt, Christian and Smith, William AP},
  booktitle={Proc. of European Conference on Computer Vision},
  pages={84--101},
  year={2020}
}

@inproceedings{Haiyang2025GaRe,
  author={Haiyang, Bai and Jiaqi, Zhu and Songru, Jiang and Wei, Huang and Tao,Lu and Yuanqi,Li and Jie,Guo and Runze, Fu and Yanwen, Guo and Lijun,Chen},
  title={{GaRe: Relightable 3D Gaussian Splatting for Outdoor Scenes from Unconstrained Photo Collections}},
  booktitle = {Proc. of the IEEE/CVF International Conference on Computer Vision},
  year={2025},
}

@inproceedings{huang20242d,
  title={{2D Gaussian Splatting for Geometrically Accurate Radiance Fields}},
  author={Huang, Binbin and Yu, Zehao and Chen, Anpei and Geiger, Andreas and Gao, Shenghua},
  booktitle={ACM SIGGRAPH 2024 Conference Papers},
  pages={1--11},
  year={2024}
}

@article{kingma2014adam,
  title={{Adam: A Method for Stochastic Optimization}},
  author={Kingma, Diederik P and Ba, Jimmy},
  journal={arXiv preprint arXiv:1412.6980},
  year={2014}
}

@inproceedings{grubert2025improving,
  title={{Improving Adaptive Density Control for 3D Gaussian Splatting}},
  author={Grubert, Glenn and Barthel, Florian and Hilsmann, Anna and Eisert, Peter},
  booktitle={Proc. of the International Conference on Computer Vision Theory and Applications},
  pages={610--621},
  year={2025}
}

@article{chen2024gi,
  title={{GI-GS: Global Illumination Decomposition on Gaussian Splatting for Inverse Rendering}},
  author={Chen, Hongze and Lin, Zehong and Zhang, Jun},
  journal={arXiv preprint arXiv:2410.02619},
  year={2024}
}

@article{wu20253d,
  title={{3D Gaussian Inverse Rendering with Approximated Global Illumination}},
  author={Wu, Zirui and Chen, Jianteng and Li, Laijian and Wu, Shaoteng and Zhu, Zhikai and Xu, Kang and Oswald, Martin R and Song, Jie},
  journal={arXiv preprint arXiv:2504.01358},
  year={2025}
}

@inproceedings{burley2012physically,
  author    = {Burley, Brent},
  title     = {{Physically Based Shading at Disney}},
  booktitle = {ACM SIGGRAPH 2012 Courses: Physically Based Shading in Theory and Practice},
  year      = {2012},
  url       = {https://blog.selfshadow.com/publications/s2012-shading-course/burley/s2012_pbs_disney_brdf_notes_v3.pdf}
}

@article{chen2022vision,
  title={{Vision Transformer Adapter for Dense Predictions}},
  author={Chen, Zhe and Duan, Yuchen and Wang, Wenhai and He, Junjun and Lu, Tong and Dai, Jifeng and Qiao, Yu},
  journal={arXiv preprint arXiv:2205.08534},
  year={2022}
}

@inproceedings{wang2023fegr,
title = {{Neural Fields meet Explicit Geometric Representations for Inverse Rendering of Urban Scenes}}, 
author = {Zian Wang and Tianchang Shen and Jun Gao and Shengyu Huang and Jacob Munkberg and Jon Hasselgren and Zan Gojcic and Wenzheng Chen and Sanja Fidler},
booktitle = {Proc. of the IEEE/CVF Conference on Computer Vision and Pattern Recognition},
  pages={8370--8380},
year = {2023}
}
\section*{\uppercase{Appendix}}
\begin{figure}[t]
  \centering
  \includegraphics[width=.6\linewidth]{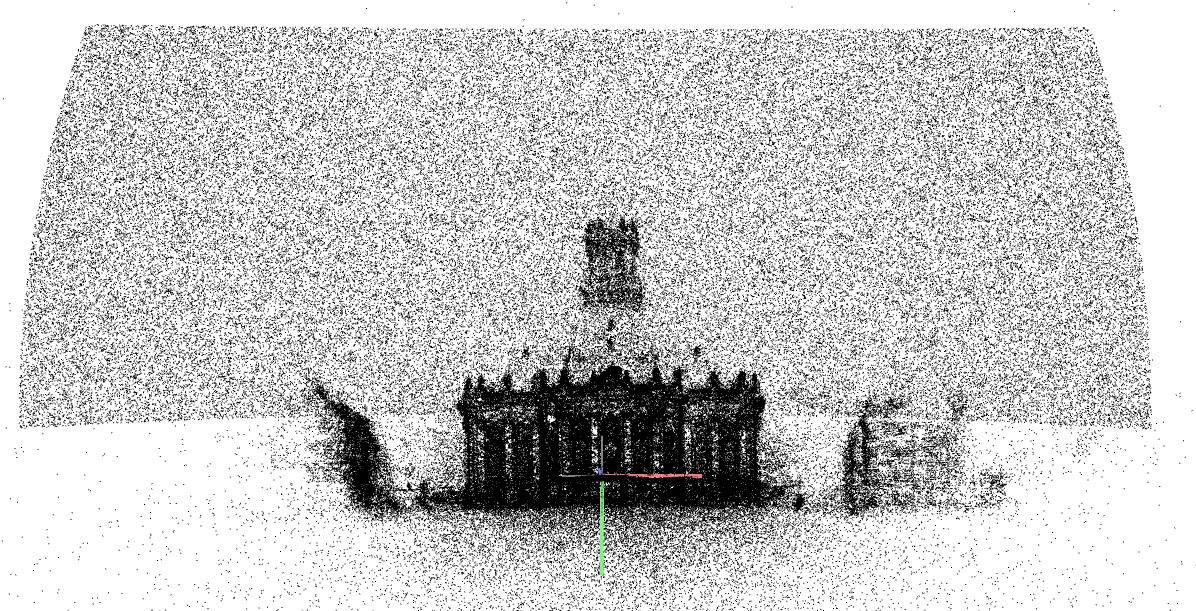}
  %
  %
  \caption{\label{fig:skyGauss}
           Initial positions of the foreground and sky Gaussians. The foreground SfM point cloud is augmented with the sky Gaussians initial positions sampled uniformly on the background of the upper hemisphere surrounding the scene.}
\end{figure}
\section*{Additional Implementation Details}
\label{sec:implementation_details}
\def\height{64px}
\def\spysize{42px}
\def\offset{36px}
\def\scale{0.7}
\def\scaleleft{0.7}
\def\distabove{0.3em}

\newcommand{\qualitativecompplus}{
\centering
\resizebox{\linewidth}{!}{
    \begin{tikzpicture}[
        >=stealth',
        title/.style={
            anchor=base,
            align=center,
            scale=\scale,
        },
        spy using outlines={rectangle, red, magnification=3, connect spies, size=\spysize}
    ]
    \matrix[matrix of nodes, column sep=0pt, row sep=0pt, ampersand replacement=\&, inner sep=-0.1, outer sep=-0.1, font=\Large] (qualitative) {
        \includegraphics[width=0.3\textwidth]{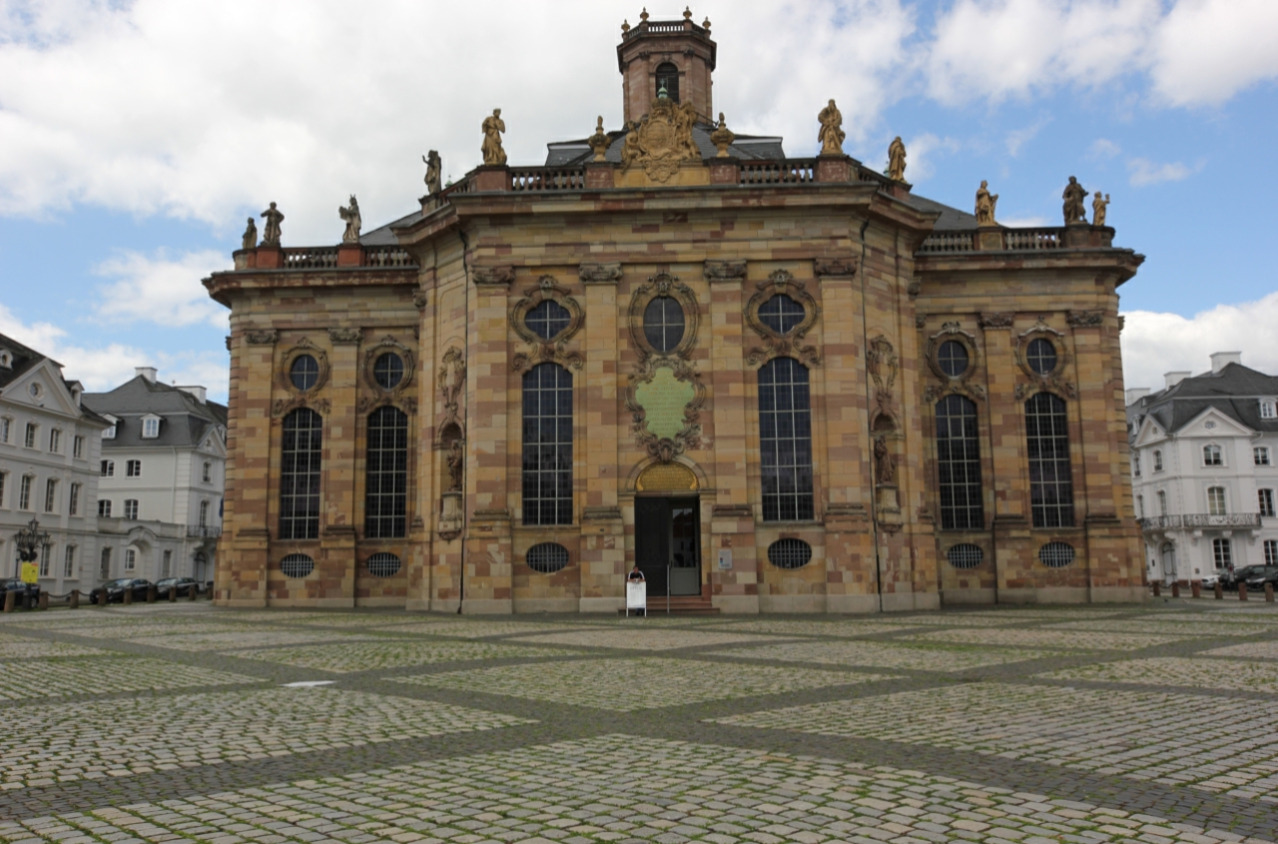} \&
        \includegraphics[width=0.3\textwidth]{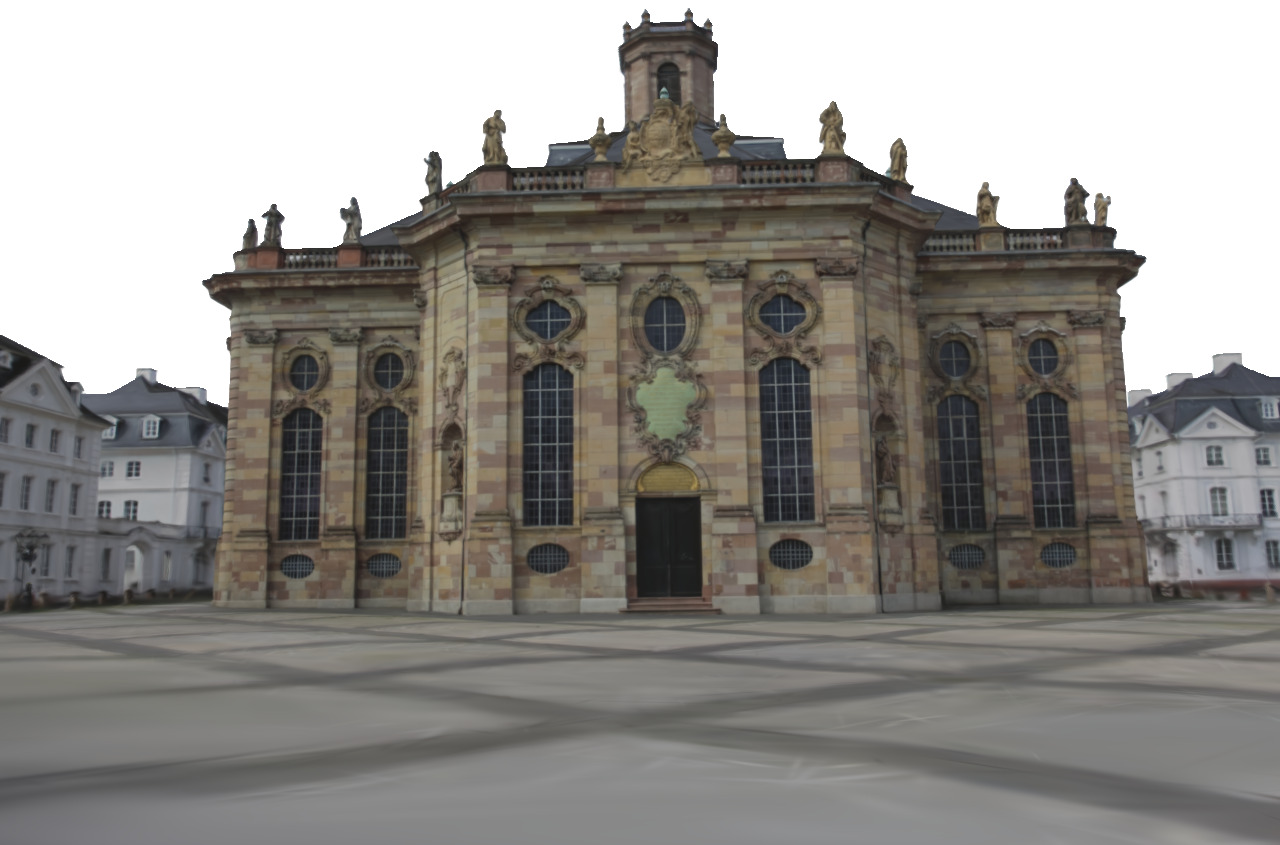} \&
        \includegraphics[width=0.3\textwidth]{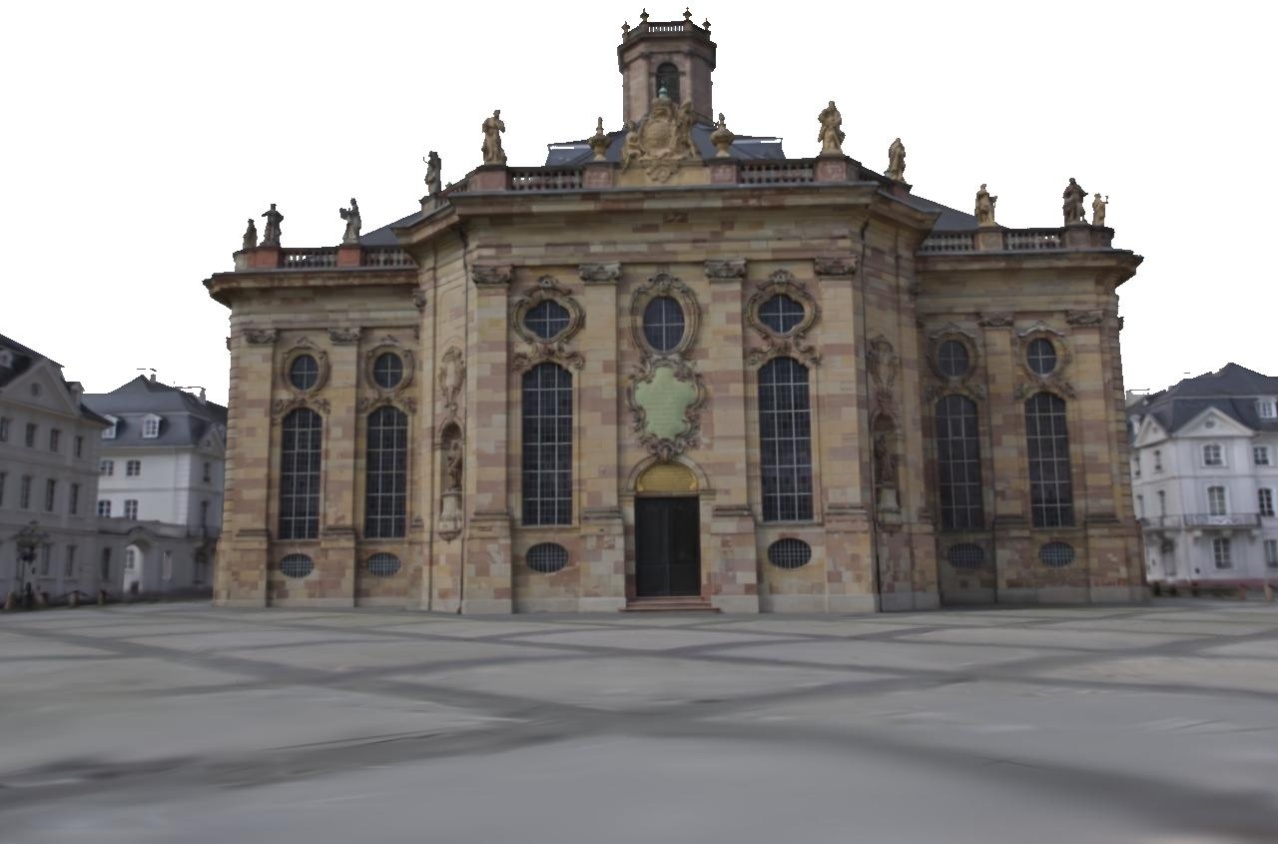} \&
        \includegraphics[width=0.3\textwidth]{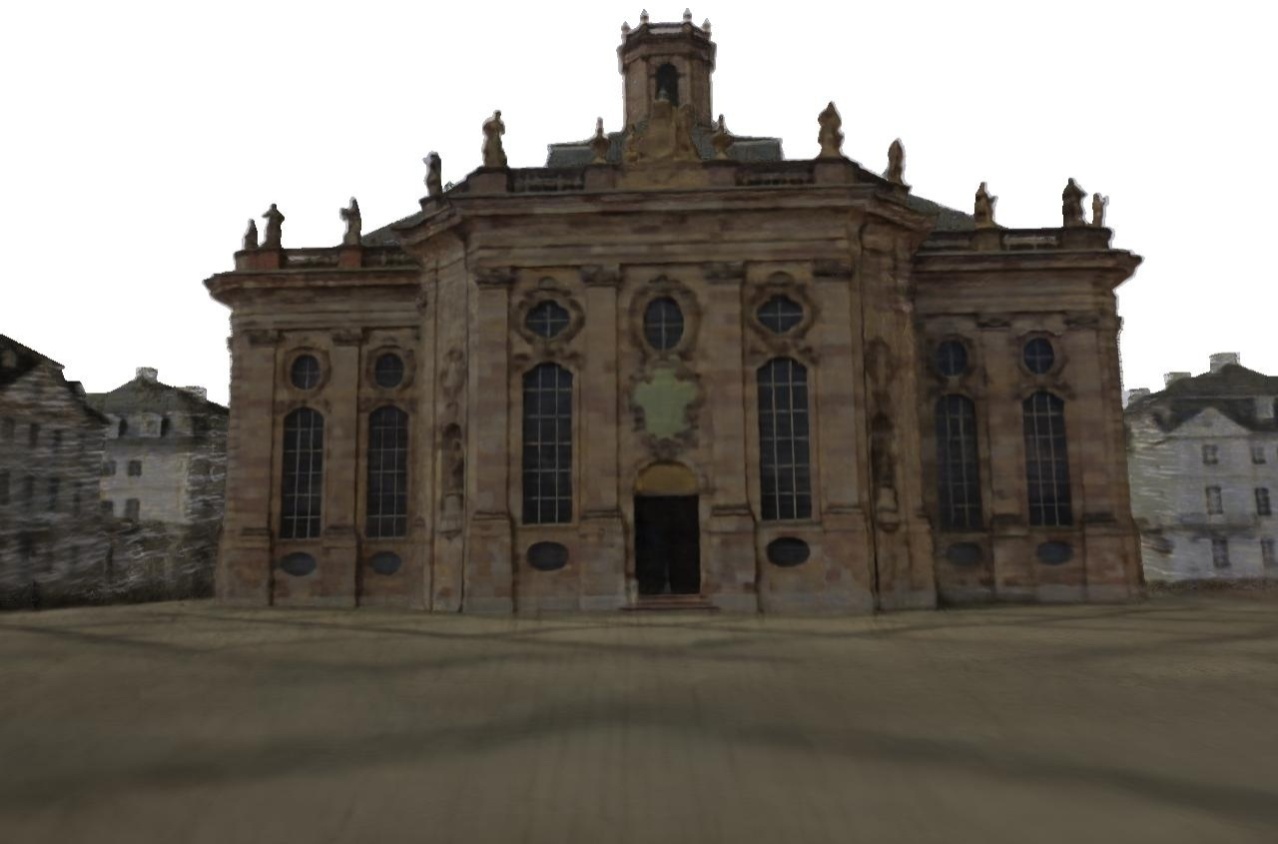} \\
        \includegraphics[width=0.3\textwidth]{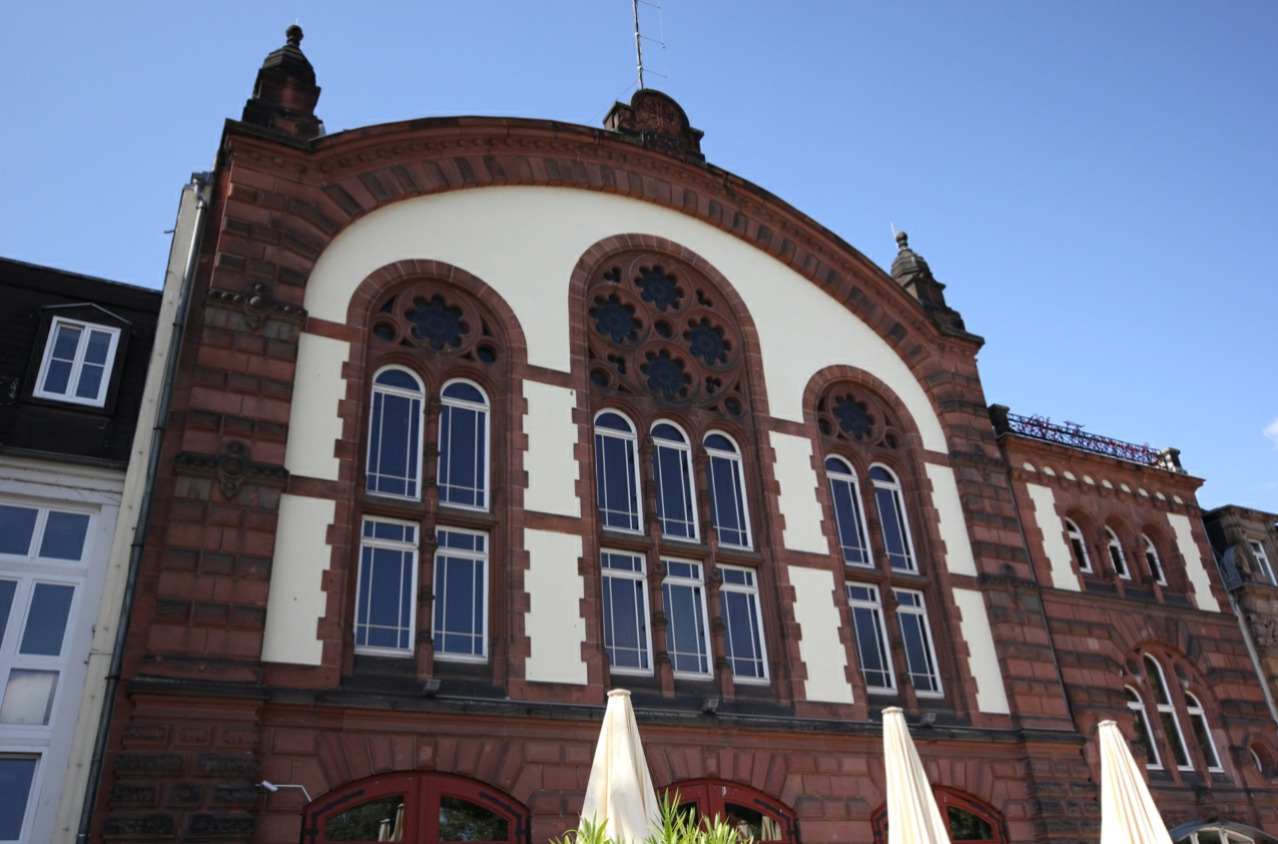} \&
        \includegraphics[width=0.3\textwidth]{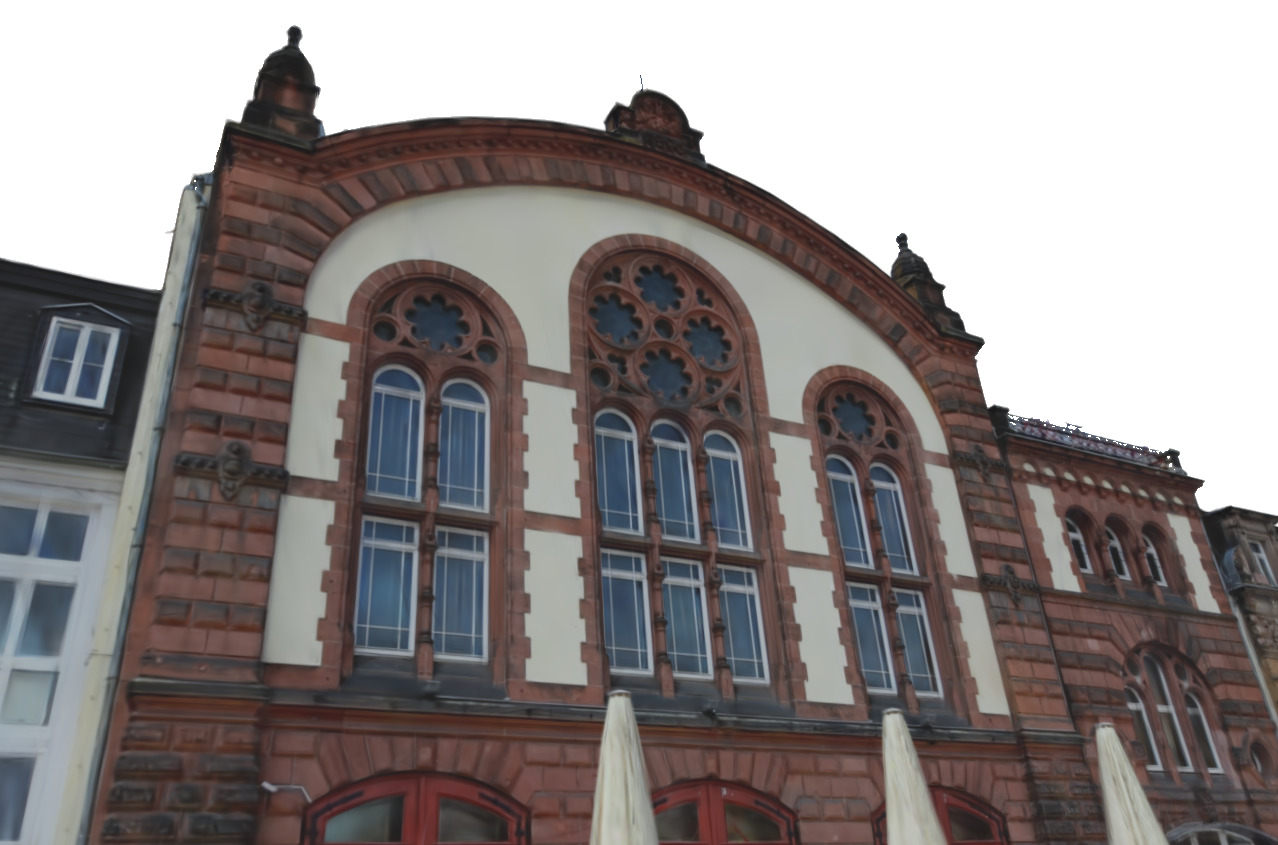} \&
        \includegraphics[width=0.3\textwidth]{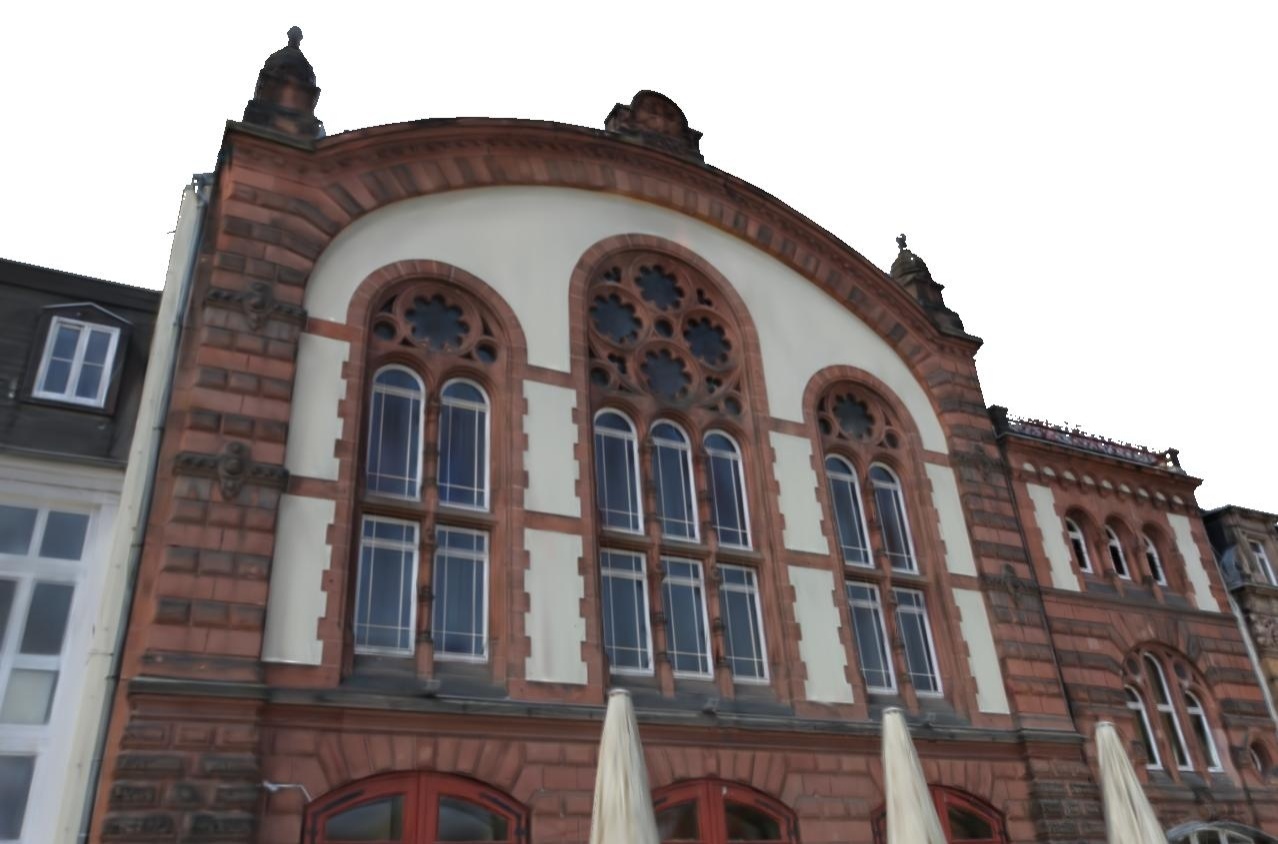}\&
        \includegraphics[width=0.3\textwidth]{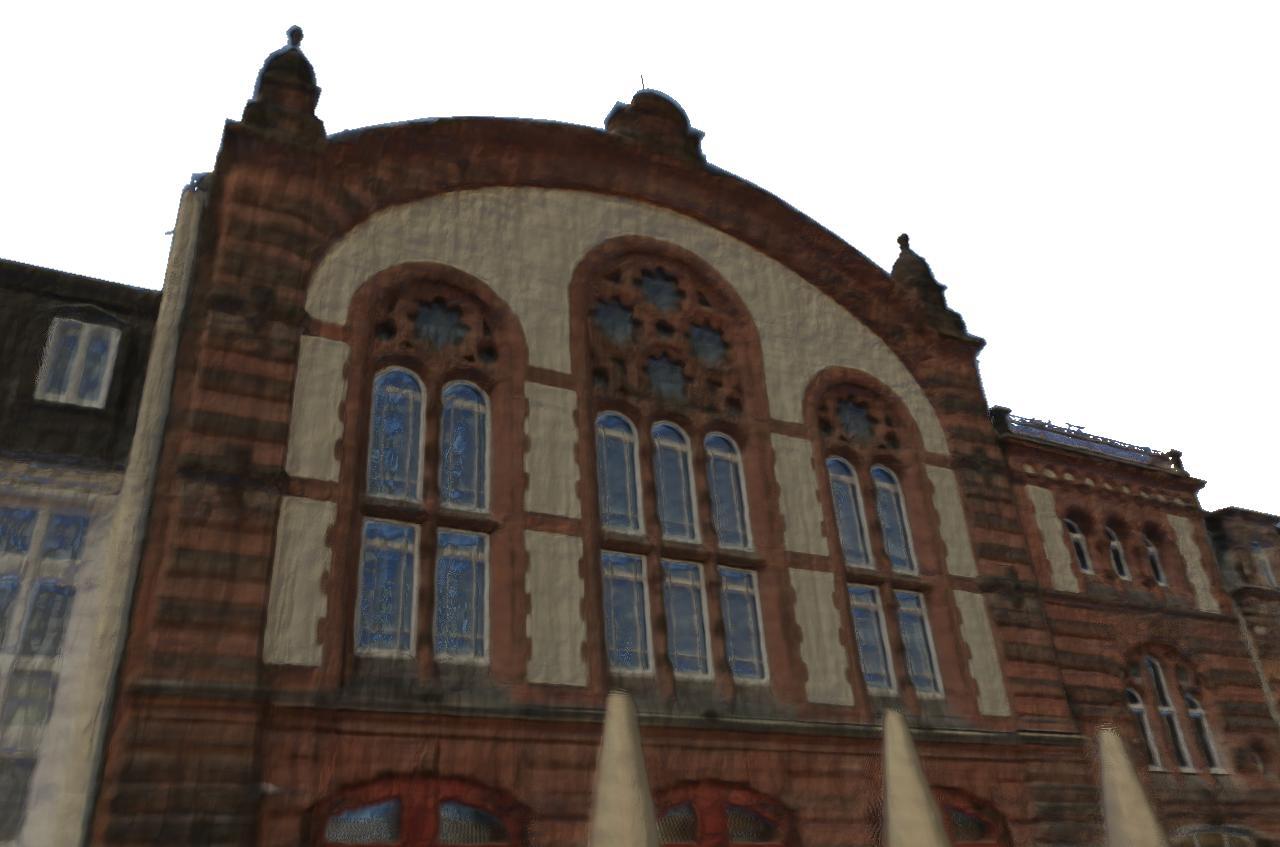} \\
        \includegraphics[width=0.3\textwidth]{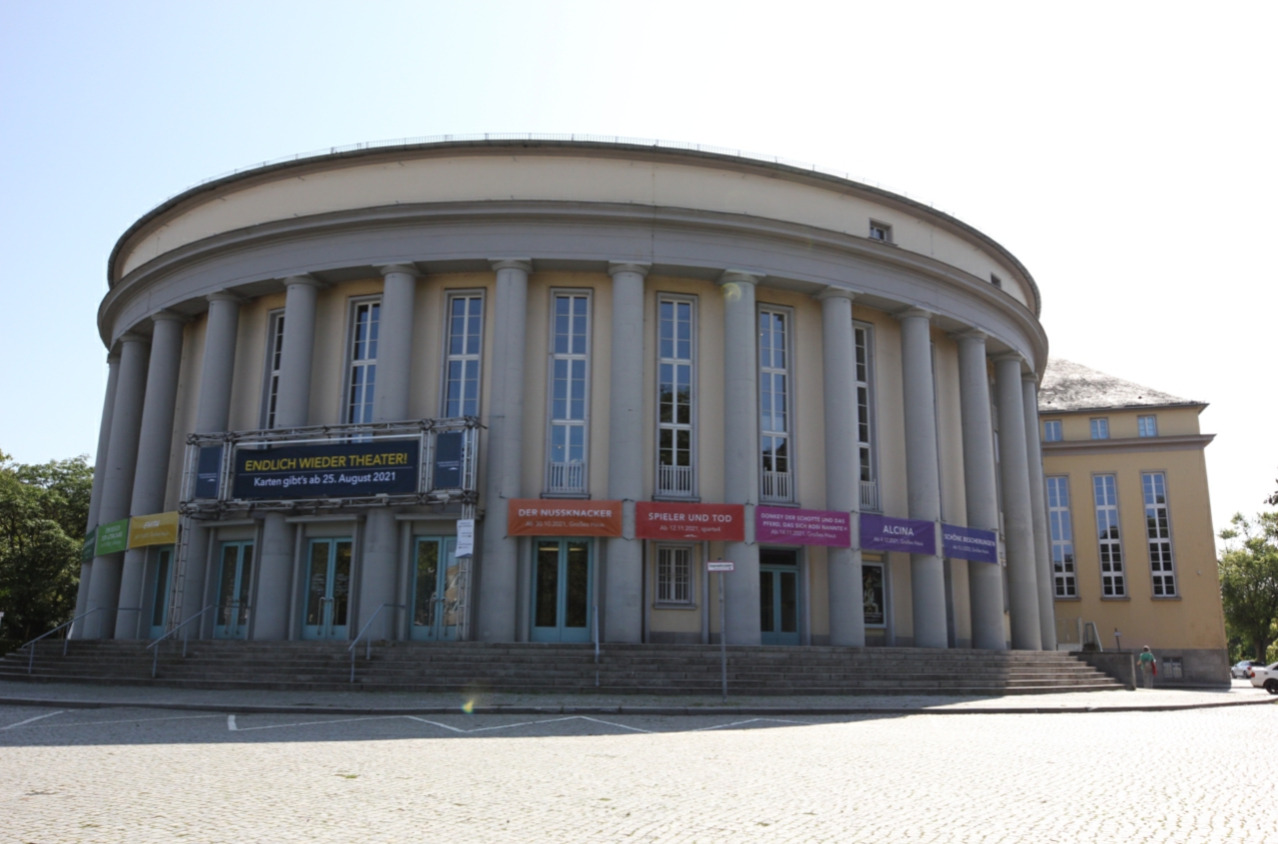}\&
        \includegraphics[width=0.3\textwidth]{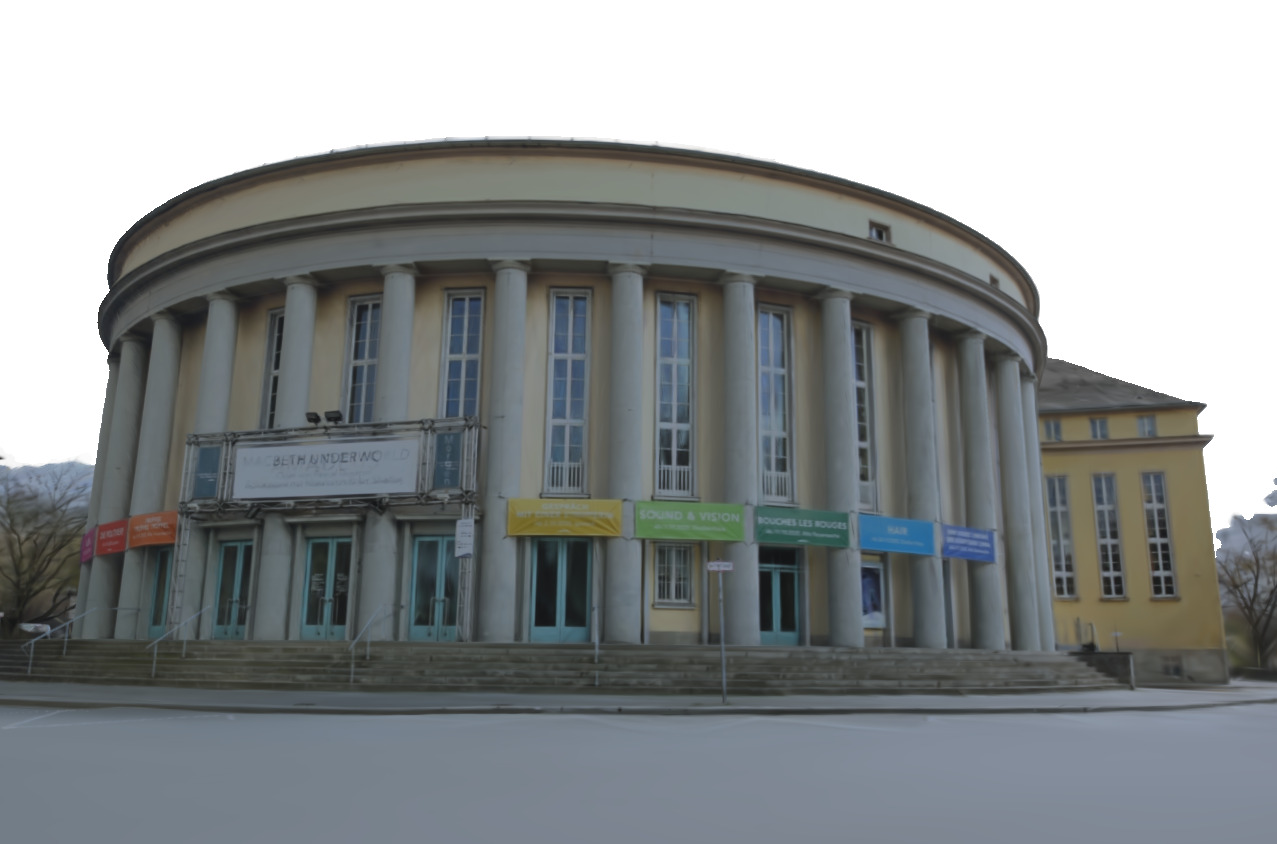} \&
        \includegraphics[width=0.3\textwidth]{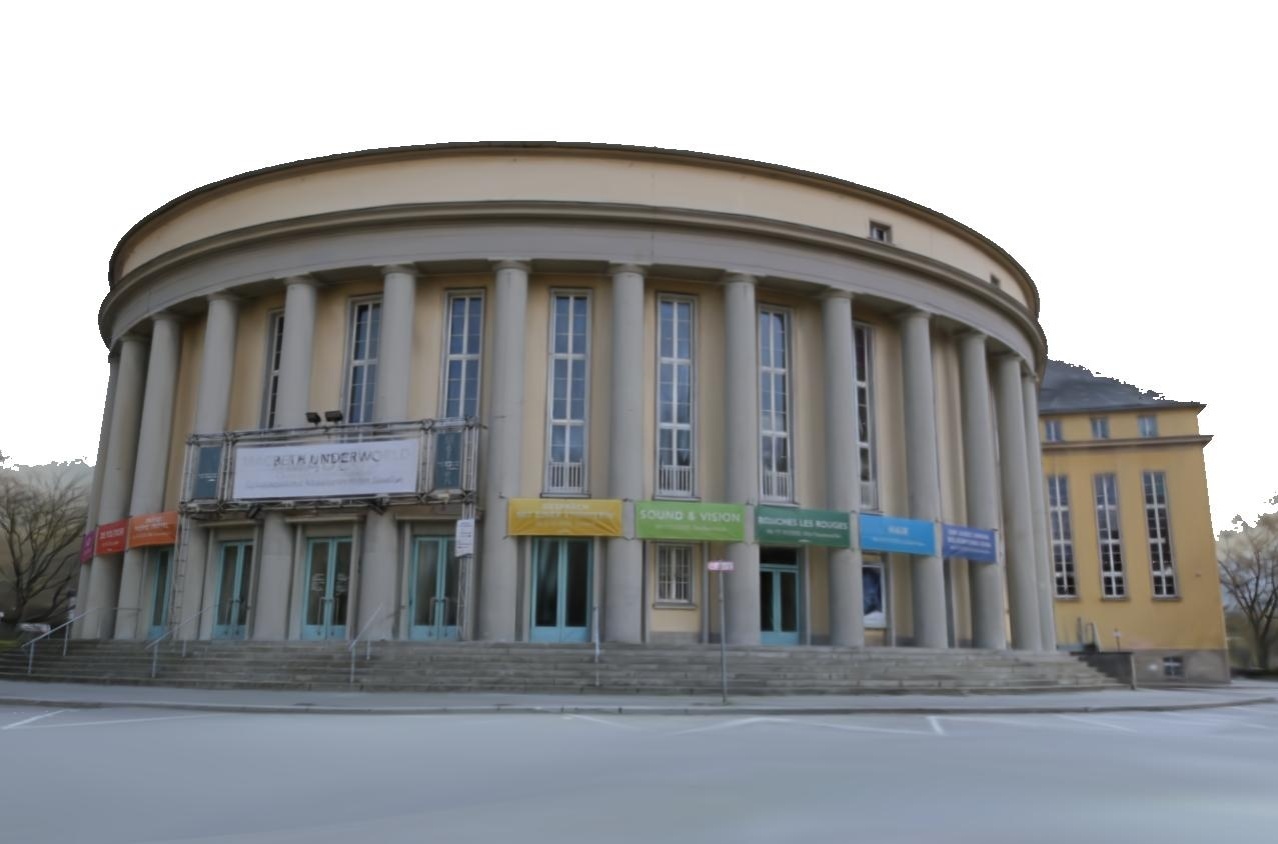} \&
        \includegraphics[width=0.3\textwidth]{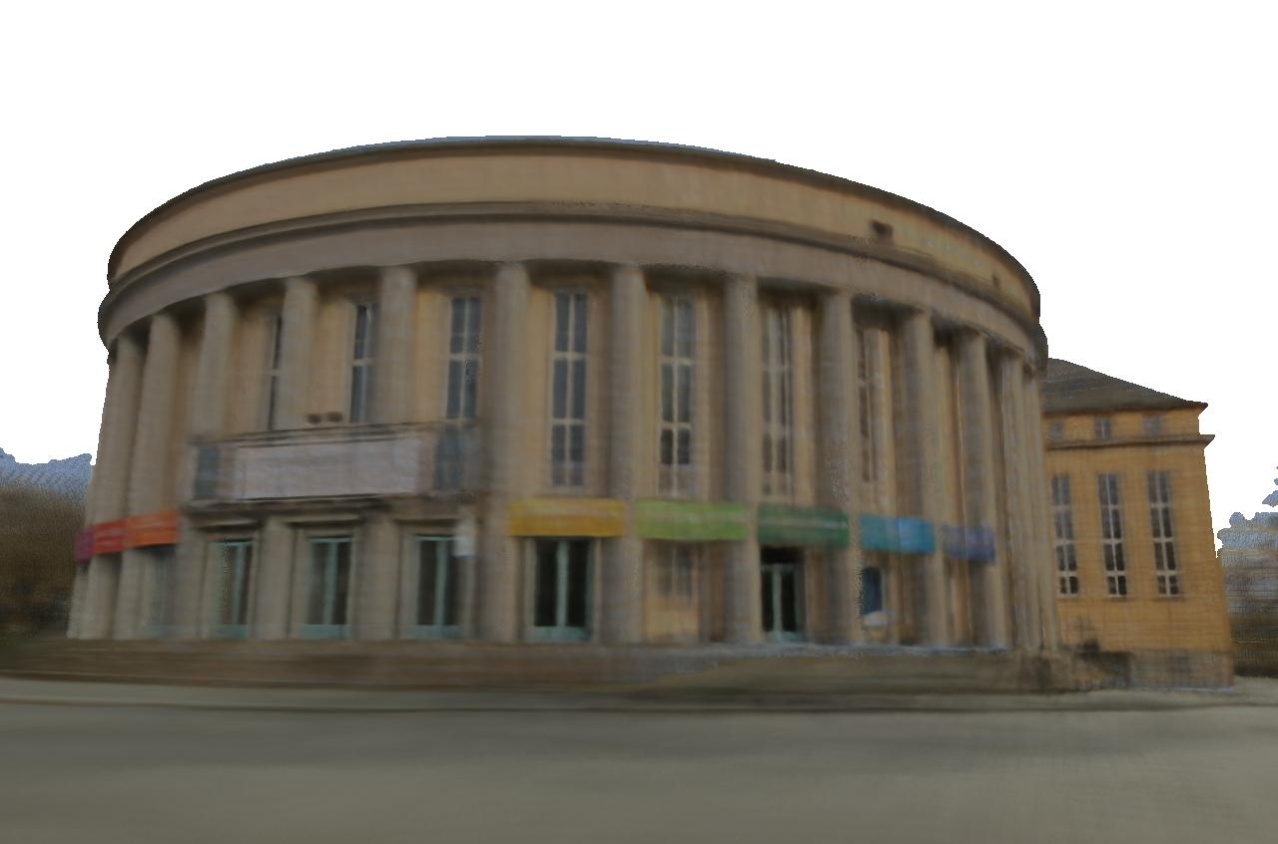} \\
    }; 

    \node[above=\distabove+0.1em of qualitative-1-1.north, title, font=\Large] {Ground Truth};
    \node[above=\distabove+0.1em of qualitative-1-2.north, title, font=\Large] {\textbf{Ours}};
    \node[above=\distabove+0.1em of qualitative-1-3.north, title, font=\Large] {\textbf{LumiGauss}};
    \node[above=\distabove+0.1em of qualitative-1-4.north, title, font=\Large] {\textbf{NeRF-OSR}};

    \foreach \n in {1,...,4}{
        \edef\temp{\noexpand\spy[anchor=center,magnification=3] on ([xshift=-2.0em,yshift=-0.5
        em] qualitative-1-\n.center) in node[below left=3.75px and 3.75px] at (qualitative-1-\n.north east);}
        \temp
    }

    \foreach \n in {1,...,4}{
        \edef\temp{\noexpand\spy[anchor=center,magnification=3] on ([xshift=0.6em,yshift=-0.7em] qualitative-2-\n.center) in node[below left=3.75px and 3.75px] at (qualitative-2-\n.north east);}
        \temp
    }
    \foreach \n in {1,...,4}{
        \edef\temp{\noexpand\spy[anchor=center,magnification=3] on ([xshift=-0.8em,yshift=1.0em] qualitative-3-\n.center) in node[below left=3.75px and 3.75px] at (qualitative-3-\n.north east);}
        \temp
    }

\end{tikzpicture}
}
}

\begin{figure*}[bt]
    \centering
    \begin{adjustbox}{center}
    \qualitativecompplus
    \end{adjustbox}
    \caption{Rendered novel views and illuminations of the NeRF-OSR test scenes (from top to bottom: \emph{lk2}, \emph{lwp}, \emph{st}) synthesized by our method, LumiGauss \cite{kaleta2025lumigauss} (shadowed version), and NeRF-OSR \cite{rudnev2022nerf}. We show novel views employed for the evaluation, rendered using ground truth environment maps. The sky is masked out in all the renderings.}
    \label{fig:qualitativecomp_renders_additional}
\end{figure*}

The implementation of our model builds upon the codebase of 3DGS \cite{kerbl3Dgaussians}.
\def\height{64px}
\def\spysize{42px}
\def\offset{36px}
\def\scale{0.7}
\def\scaleleft{0.7}
\def\distabove{0.3em}
\def\distleft{0.3em}

\newcommand{\qualitativecompalbedonormalplus}{
\centering
\resizebox{\linewidth}{!}{
    \begin{tikzpicture}[
        >=stealth',
        title/.style={
            anchor=base,
            align=center,
            scale=\scale,
        },
        spy using outlines={rectangle, red, magnification=3, connect spies, size=\spysize}
    ]
    \matrix[matrix of nodes, column sep=0pt, row sep=0pt, ampersand replacement=\&, inner sep=0, outer sep=0, font=\Large] (qualitative) {
        \includegraphics[width=.8\linewidth]{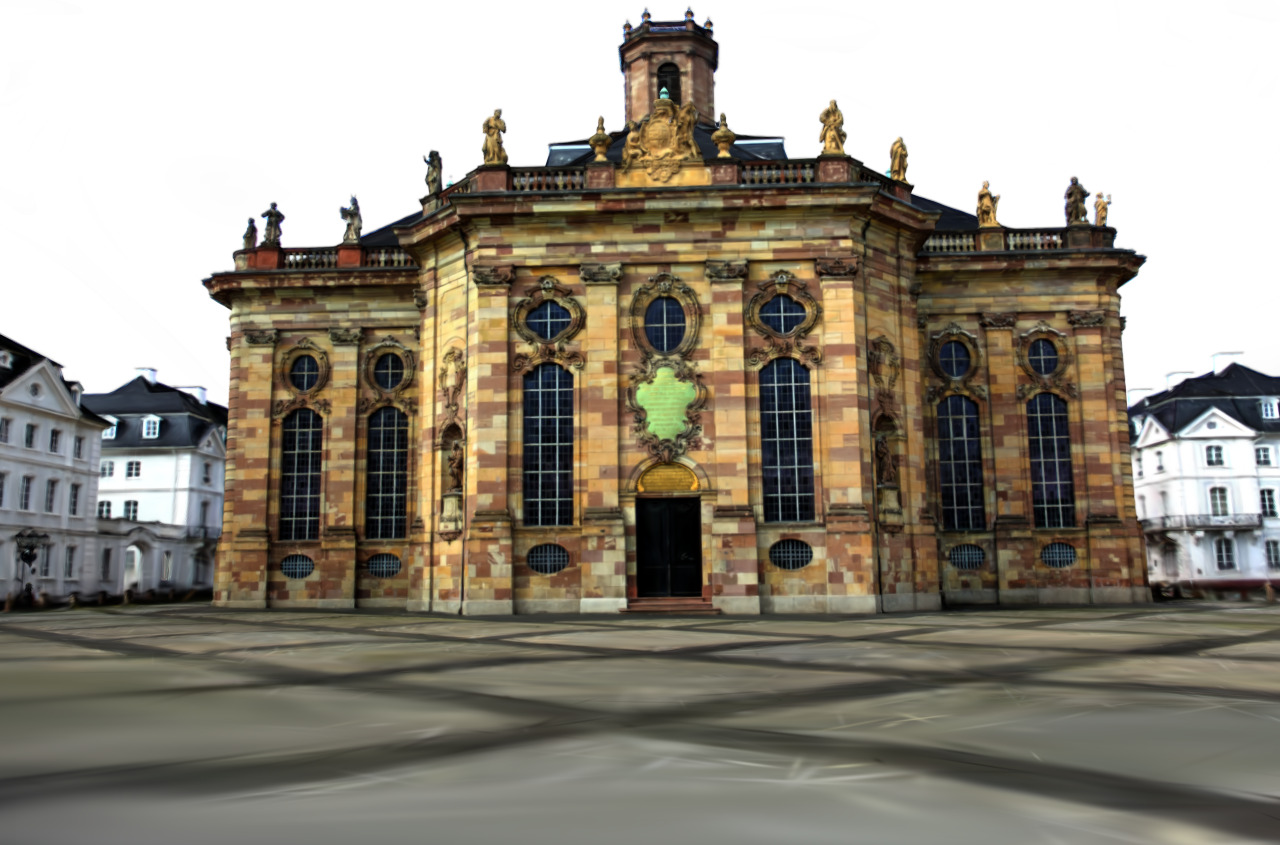} \&
        \includegraphics[width=.8\linewidth]{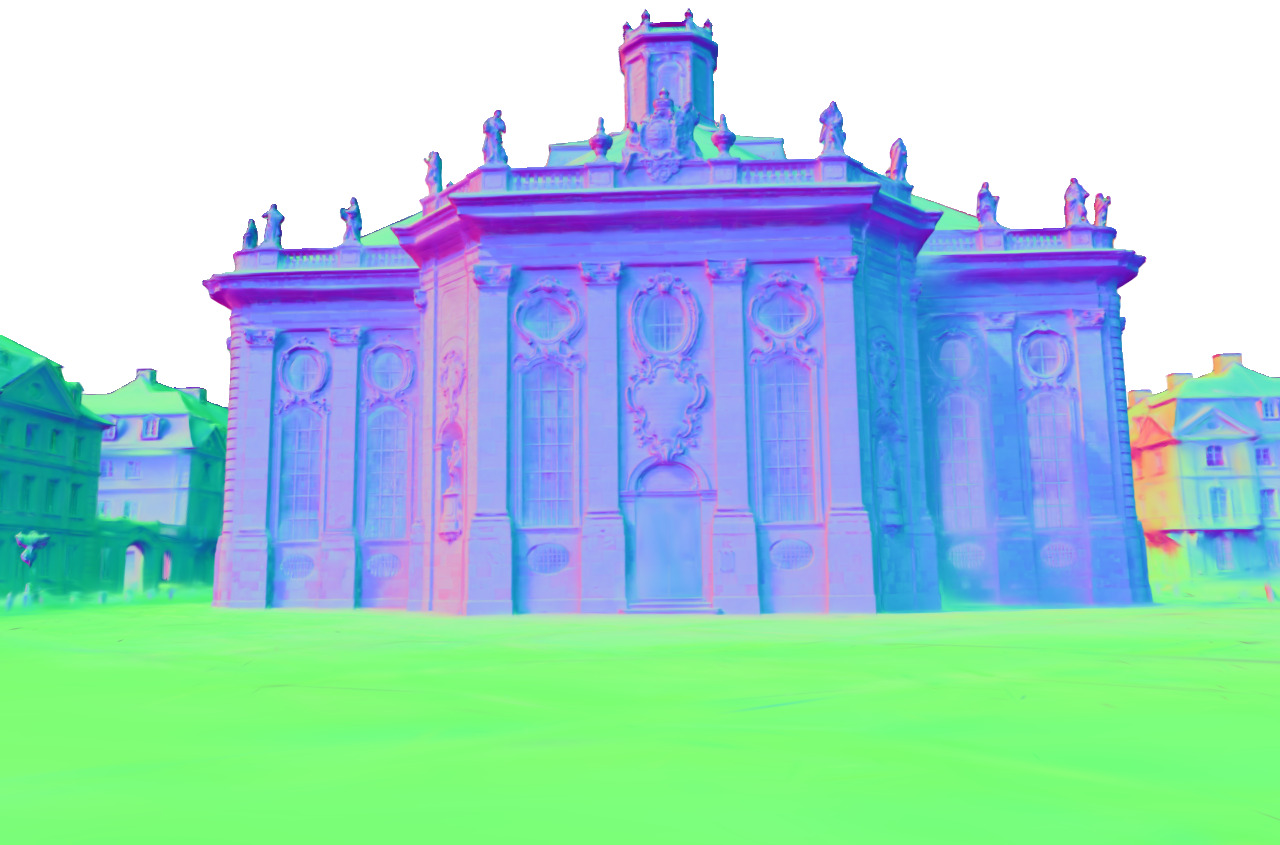} \\
        \includegraphics[width=.8\linewidth]{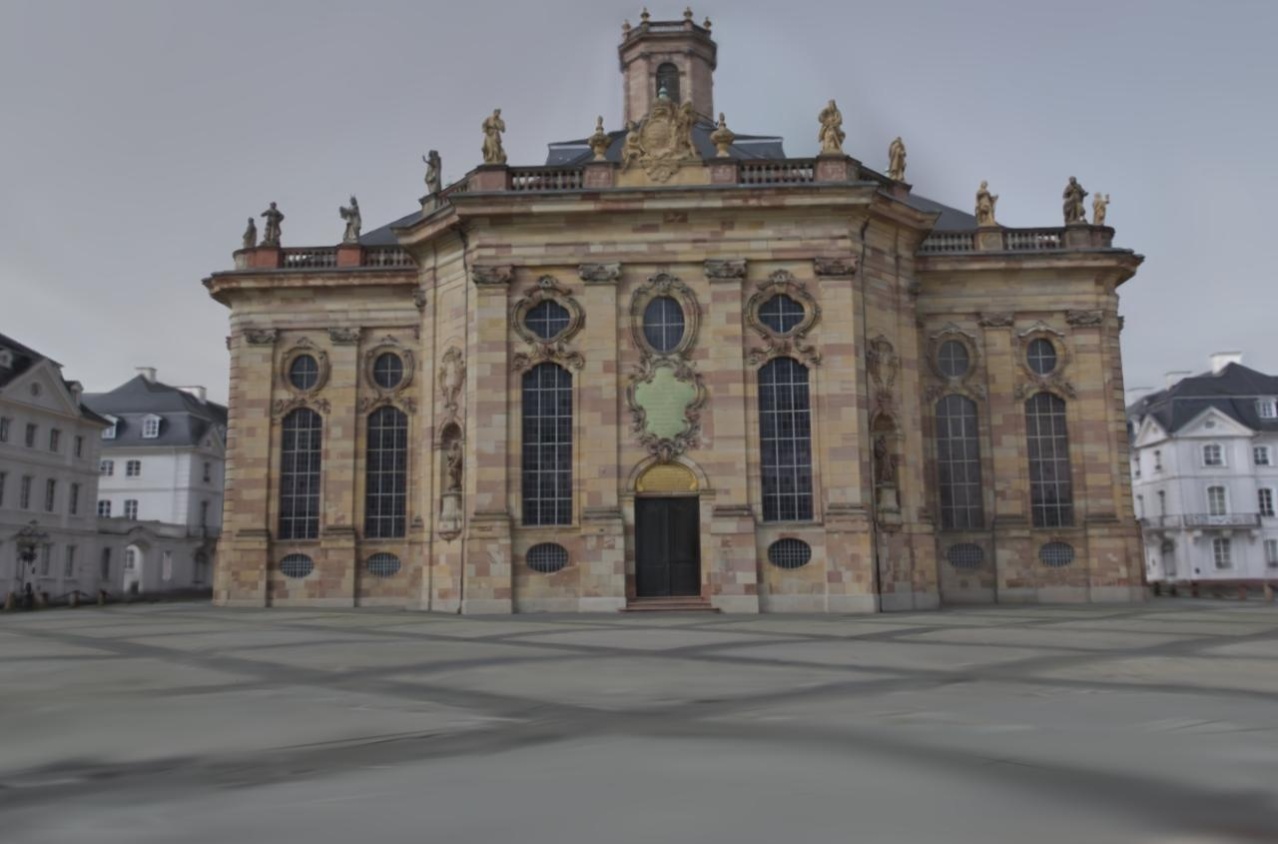} \&
        \includegraphics[width=.8\linewidth]{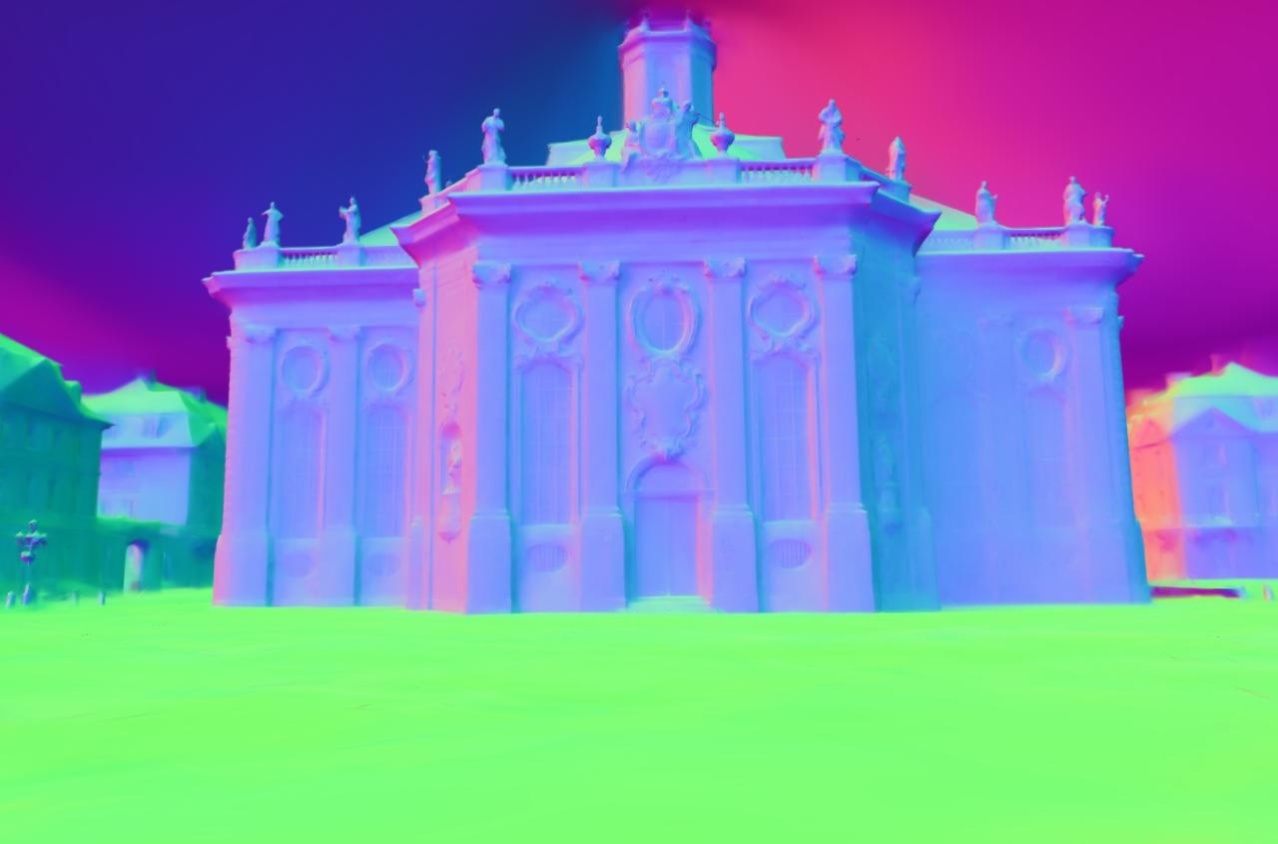} \\
        \includegraphics[width=.8\linewidth]{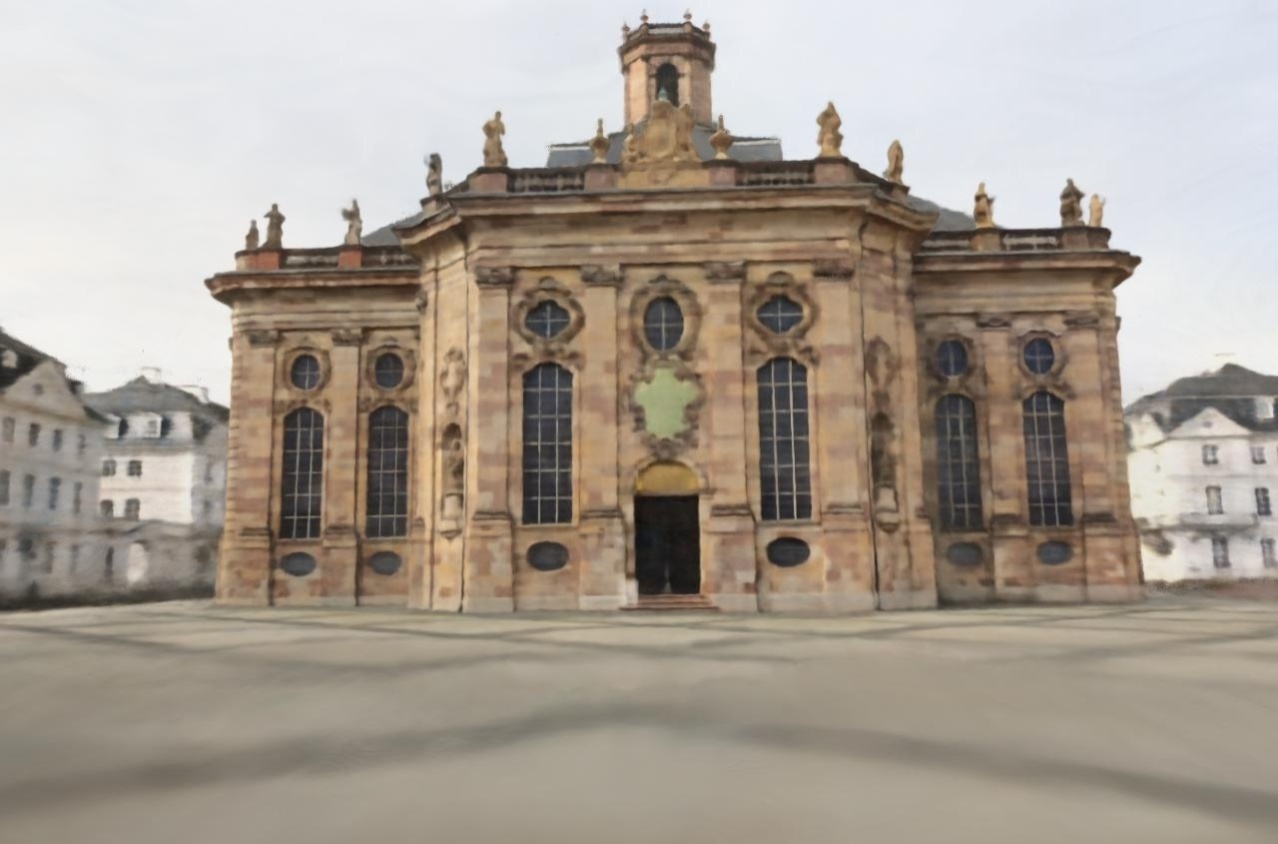} \&
        \includegraphics[width=.8\linewidth]{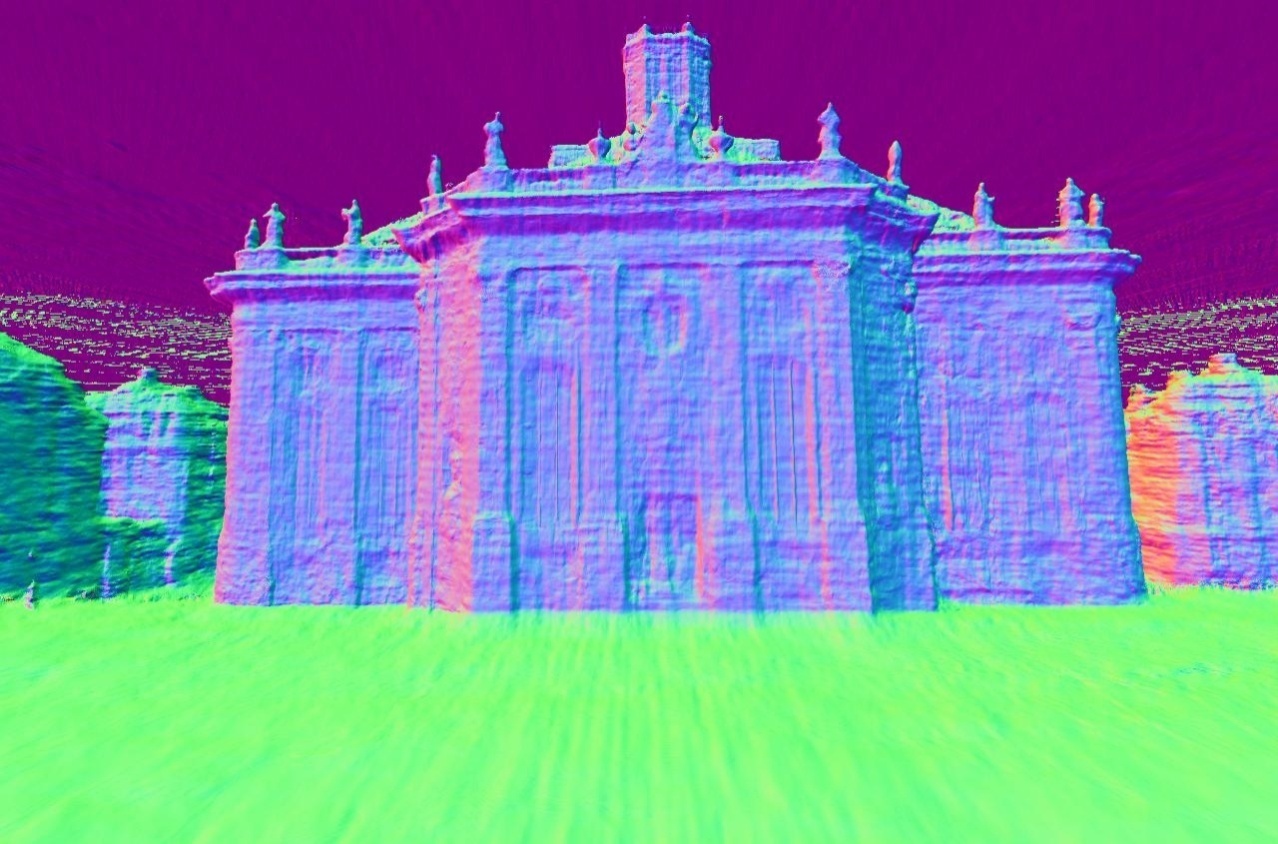}\\
    }; 

    \node[above=\distabove of qualitative-1-1.north, title, font=\Huge] {Albedo};
    \node[above=\distabove of qualitative-1-2.north, title, font=\Huge] {Surface Normals};

    \node[left=\distleft of qualitative-1-1.west, align=center,scale=\scaleleft, font=\Huge]{\rotatebox{90}{\textbf{Ours}}};
    \node[left=\distleft of qualitative-2-1.west, align=center,scale=\scaleleft, font=\Huge]{\rotatebox{90}{\textbf{LumiGauss}}};
    \node[left=\distleft of qualitative-3-1.west, align=center,scale=\scaleleft, font=\Huge]{\rotatebox{90}{\textbf{NeRF-OSR}}};

\end{tikzpicture}
}
}

\begin{figure}[!htb]
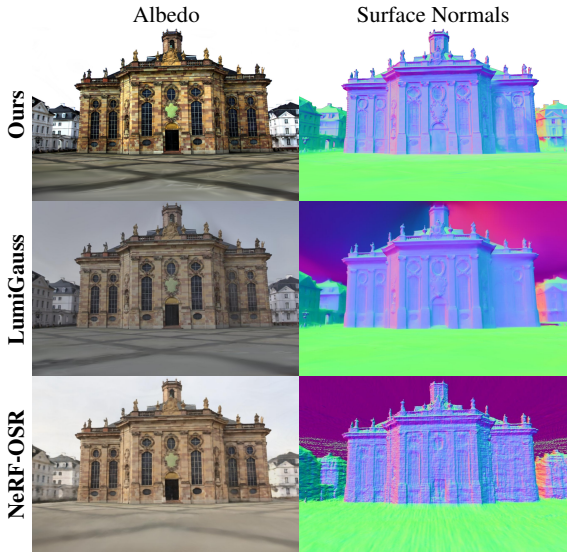

    \centering
    \begin{adjustbox}{center}
    \qualitativecompalbedonormalplus
    \end{adjustbox}
    \caption{Additional qualitative comparison of the albedo map and surface normals predicted by our method, NeRF-OSR \cite{rudnev2022nerf}, and LumiGauss \cite{kaleta2025lumigauss} for the NeRF-OSR scene \emph{lk2}. The ground truth novel view is visualized in Fig.~\ref{fig:qualitativecomp_renders_additional}.}
    \label{fig:qualitativecomp_albedo_normal_maps_additional}
\end{figure}

\paragraph*{MLP Architecture and Appearance Embeddings.}
The MLP first processes the input image embedding with 3 fully connected layers, each with 256 units. The output feature vector is then fed into a dense layer that returns the SH coefficients of the sky color and into an additional dense layer of size 128. The latter is followed by a fully connected layer that predicts the SH coefficients of the environment light. The image appearance embeddings have dimension $128$.

\paragraph*{Training Details.}
We optimized the parameters of our model using Adam optimizer~\cite{kingma2014adam}. We set the learning rate for the appearance embeddings, MLP weights, and foreground Gaussians' roughness to $0.0002$, to $0.002$ for the foreground Gaussians' albedo, and to $0.0001$ for the radius of the sky Gaussians. The sky Gaussians' angle attributes are optimized using the same learning rate as for the positions of the foreground Gaussians. The training starts with 500 warm-up iterations, where only $\mathcal{L_{\text{fg-sky}}}$ is active. Afterwards, all the losses are applied, except for $\mathcal{L}_{\text{normal}}$ and $\mathcal{L}_{\text{scale}}$, which are started after \num{2000} iterations. We set $\gamma$ in $\mathcal{L}_{\text{sky-depth}}$ to $0.02$. Occluders (e.g.\ people, cars), which characterize a scene captured in the wild in addition to its varying lighting, are masked out using binary segmentation masks in all the losses.

During training, both foreground and sky Gaussians are adaptively controlled as in 3DGS \cite{kerbl3Dgaussians} (see Fig.~\ref{fig:skyGauss} for a visualization of the initial sets of sky and foreground Gaussians). We modify the original densification scheme by adopting an exponential ascending gradient threshold to improve stability during optimization \cite{grubert2025improving}. When a sky Gaussian is split, the sampled positions of the two children are first projected onto the sky hemisphere. The Cartesian coordinates are then converted into spherical coordinates to obtain the new polar and azimuthal angle attributes.
\paragraph*{Segmentation Masks.}
Our method takes as input occluders and sky masks, which are required for the loss computation. We used the strategy of \cite{gardner2024sky} and extracted them from the segmentation masks generated using \cite{chen2022vision}.

\section*{Additional Qualitative Results}
Fig.~\ref{fig:qualitativecomp_renders_additional} and Fig.~\ref{fig:qualitativecomp_albedo_normal_maps_additional} provide an additional qualitative comparison of our method with LumiGauss \cite{kaleta2025lumigauss} and NeRF-OSR \cite{rudnev2022nerf}. Fig.~\ref{fig:viewdep_effects} shows how the appearance of the scene changes with viewpoint under fixed lighting.
\def\height{64px}
\def\spysize{42px}
\def\offset{36px}
\def\scale{0.7}
\def\scaleleft{0.7}
\def\distabove{0.3em}

\newcommand{\viewdepeffects}{
\centering
\resizebox{\linewidth}{!}{
    \begin{tikzpicture}[
        >=stealth',
        title/.style={
            anchor=base,
            align=center,
            scale=\scale,
        },
        spy using outlines={rectangle, red, magnification=3, connect spies, size=\spysize}
    ]
    \matrix[matrix of nodes, column sep=0pt, row sep=0pt, ampersand replacement=\&, inner sep=-0.1, outer sep=-0.1, font=\Large] (qualitative) {
        \includegraphics[width=0.3\textwidth]{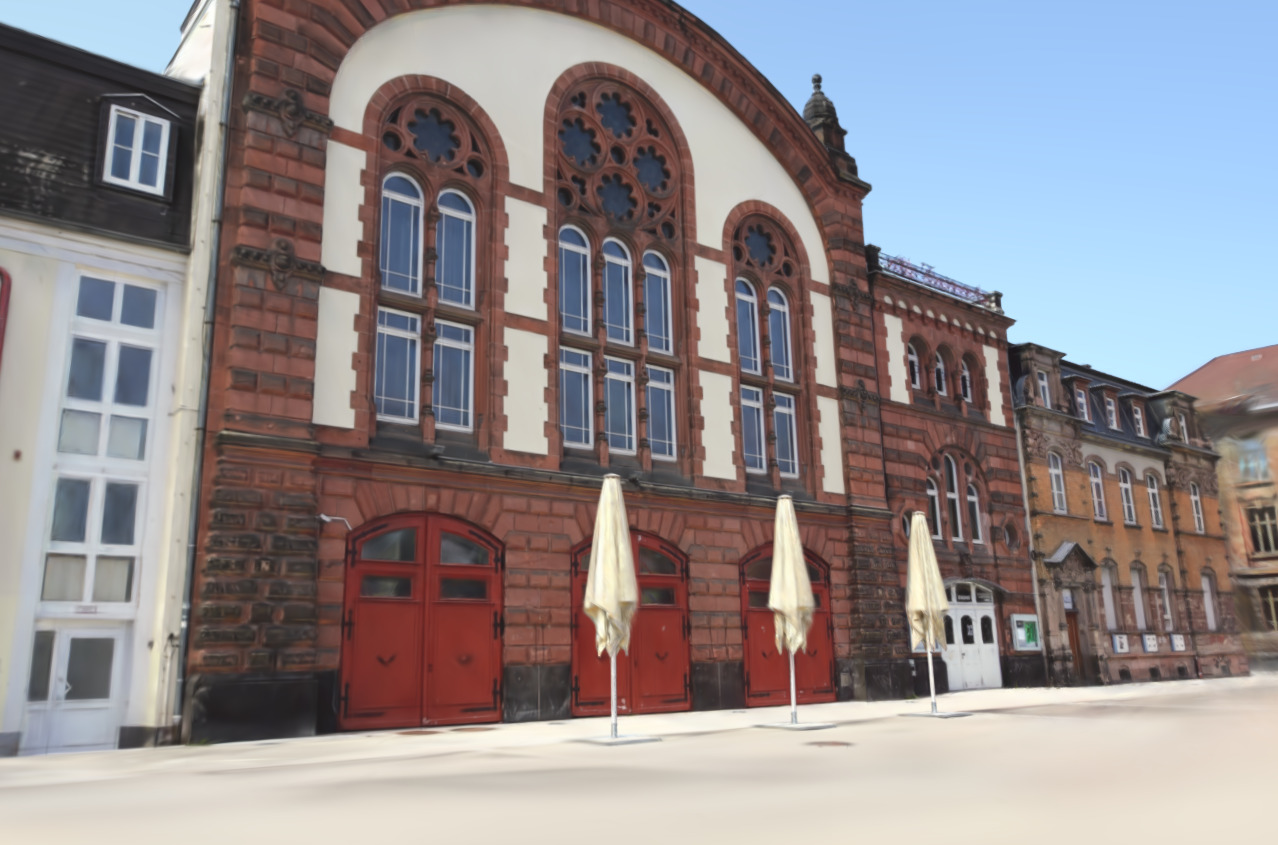} \&
        \includegraphics[width=0.3\textwidth]{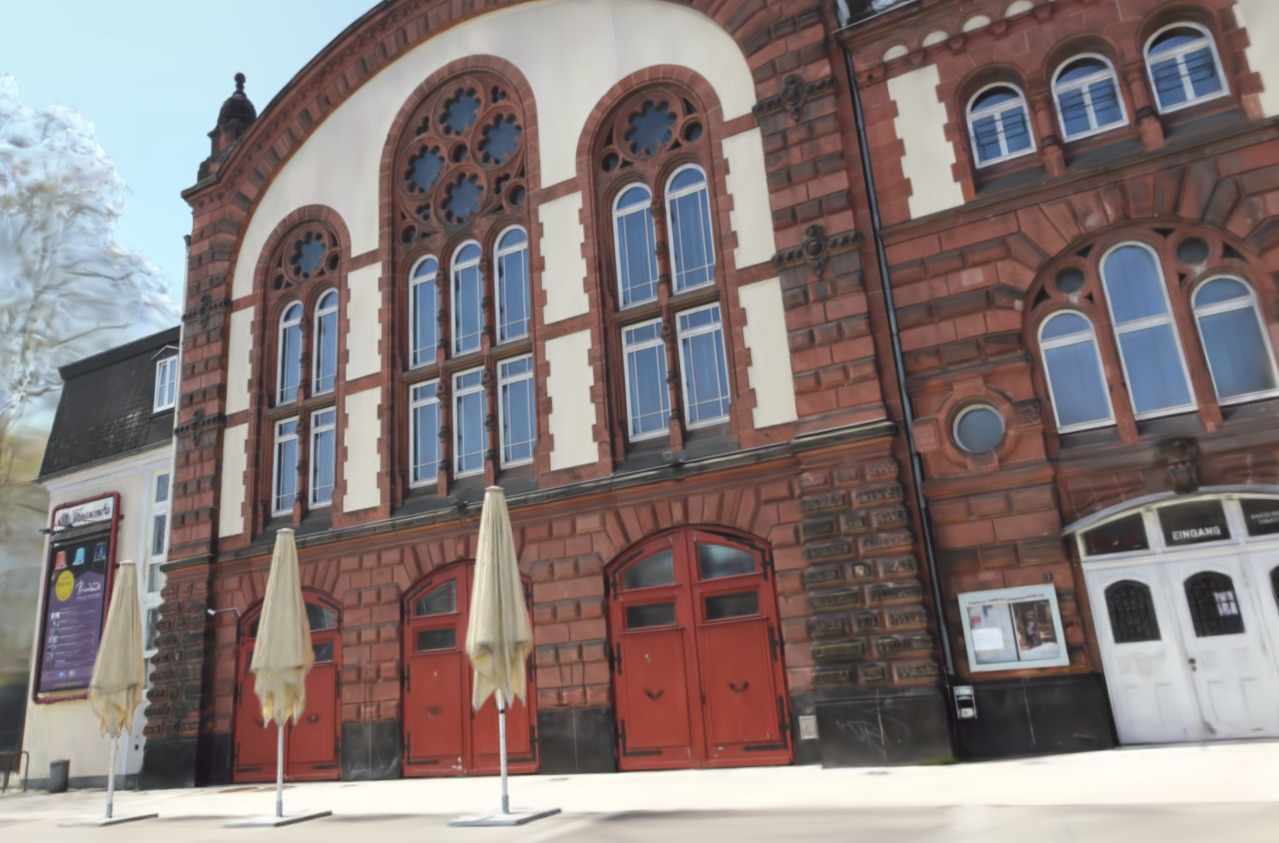} \\
    }; 
        
    \edef\temp{\noexpand\spy[anchor=center,magnification=3] on ([xshift=0.0em,yshift=1.4em] qualitative-1-1.center) in node[below left=3.75px and 3.75px] at (qualitative-1-1.north east);}
    \temp

    \edef\temp{\noexpand\spy[anchor=center,magnification=3] on ([xshift=-1.6em,yshift=1.4em] qualitative-1-2.center) in node[below left=3.75px and 3.75px] at (qualitative-1-2.north east);}
    \temp

\end{tikzpicture}
}
}

\begin{figure}[!b]
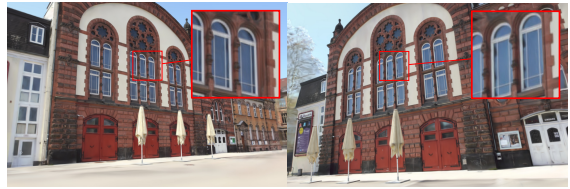

    \centering
    \viewdepeffects
    \caption{Comparison of two novel views of the NeRF-OSR scene \emph{lwp} rendered under the same trained lighting and sky color. Insets highlight that the appearance of reflective surfaces, such as windows, varies with viewing direction.}
    \label{fig:viewdep_effects}
\end{figure}

\end{document}